\documentclass[twoside,11pt]{article}

\def\twocolumns{0}
\def\viewchanges{1}
\def\addToC{0}
\def\jmlrstyle{1}
\def\preprint{1}
\usepackage[utf8]{inputenc} %I allow utf-8 input
\usepackage[T1]{fontenc}    % use 8-bit T1 fonts
\usepackage[round]{natbib}
\usepackage{url}            % simple URL typesetting
\usepackage{booktabs}       % professional-quality tables
\usepackage{amsfonts}       % blackboard math symbols
\usepackage{amssymb} 
\usepackage{mathtools}
\usepackage{nicefrac}       % compact symbols for 1/2, etc.
\usepackage{microtype}      % microtypography
\usepackage{lipsum}
\usepackage{graphicx}
\usepackage{amsmath}
\usepackage{amsthm}
\usepackage{caption}
\usepackage{subcaption}
\usepackage{enumerate}
\graphicspath{ {./images/} }
\usepackage{algorithmic}
\usepackage{algorithm}
\usepackage{multirow}
\usepackage{array}
\usepackage{multicol}
\usepackage{wrapfig}
\usepackage{float}
\usepackage{enumitem}
\usepackage{lscape}
\usepackage{csvsimple}
\newcolumntype{P}[1]{>{\centering\arraybackslash }p{#1}}
\renewcommand{\algorithmiccomment}[1]{\bgroup\hfill ~#1\egroup}
\usepackage{xr}
\if\jmlrstyle1
    \if\preprint1
     \usepackage[preprint]{jmlr2e}
    \else
     \usepackage{jmlr2e}
    \fi
\else
  \usepackage{arxiv}
  \usepackage[colorlinks, citecolor=blue]{hyperref}
  \newtheorem{theorem}{Theorem}[section]
  \newtheorem{lemma}[theorem]{Lemma}

\fi
\usepackage[dvipsnames]{xcolor}
\usepackage{colortbl}
\definecolor{Gray}{gray}{0.85}

\let\del\undefined
\let\com\undefined

\if\viewchanges1
	
	\newcommand{\del}[1]{{\color{red} #1}}
	\newcommand{\com}[1]{{\color{orange} #1}}
	
\else
	
	\newcommand{\del}[1]{}
	\newcommand{\com}[1]{}
	
\fi
\if\jmlrstyle1
\if\preprint0
\jmlrheading{}{2021}{1-30}{4/21}{00/00}{paper id}{Calypso Herrera, Florian Krach, Pierre Ruyssen and Josef Teichmann}

\ShortHeadings{Optimal Stopping via Randomized Neural Networks}{Herrera, Krach, Ruyssen and Teichmann}
\firstpageno{1}
\fi
\fi
\title{Optimal Stopping via Randomized Neural Networks}

\if\jmlrstyle1
\author{\name Calypso Herrera \email calypso.herrera@math.ethz.ch \\
        \addr Department of Mathematics, ETH Zurich, Rämistrasse 101, 8092 Zurich, Switzerland\\
       \name Florian Krach \email florian.krach@math.ethz.ch\\
       \addr Department of Mathematics, ETH Zurich, Rämistrasse 101, 8092 Zurich, Switzerland\\
       \name Pierre Ruyssen \email pierrot@google.com \\
       \addr Google Brain, Google Zurich, Brandschenkestrasse 110, 8002 Zurich, Switzerland \\
       \name Josef Teichmann \email Josef.Teichmann@math.ethz.ch \\
       \addr Department of Mathematics, ETH Zurich, Rämistrasse 101, 8092 Zurich, Switzerland\\
       }

\editor{TBA}
\else
\author{%
	  Calypso Herrera \quad Florian Krach \quad Pierre Ruyssen* \quad Josef Teichmann
\\
	  Department of Mathematics, ETH Zurich, Switzerland	  
\\ *Google Brain, Zurich, Switzerland}
\fi
\begin{document}
\maketitle
\if\addToC1
	\tableofcontents
\fi
\if\twocolumns1
  \begin{multicols}{1}
\fi

\begin{abstract}%
This paper presents the benefits of using randomized neural networks instead of standard basis functions or deep neural networks to approximate the solutions of optimal stopping problems. The key idea is to use neural networks, where the parameters of the hidden layers are generated randomly and only the last layer is trained, in order to approximate the continuation value. Our approaches are applicable to high dimensional problems where the existing approaches become increasingly impractical. In addition, since our approaches can be optimized using simple linear regression, they are easy to implement and theoretical guarantees can be provided. We test our approaches for American option pricing on Black--Scholes, Heston and rough Heston models and for optimally stopping a fractional Brownian motion. In all cases, our algorithms outperform the state-of-the-art and other relevant machine learning approaches in terms of computation time while achieving comparable results. Moreover, we show that they can also be used to efficiently compute Greeks of American options.
\end{abstract}

% \setcounter{footnote}{1}
% \footnotetext{Corresponding author: Josef Teichmann.}
% \setcounter{footnote}{2}
% \footnotetext{\add{MSC2020 classification: 60G40.}}

%% use \begin{keywords} .. \end{keywords} for keywordss.
%\newenvironment{keywords}
%{\bgroup\leftskip 20pt\rightskip 20pt \small\noindent{\bf Keywords:} }%
%{\par\egroup\vskip 0.25ex}

\if\jmlrstyle1
\else
\fi

\section{Introduction}

The optimal stopping problem consists in finding the optimal time to stop in order to maximize an expected reward. This problem is found in the areas of statistics, economics, and financial mathematics. Despite significant advances, it remains one of the most challenging problems in optimization, in particular when more than one factor affects the expected reward. A common provable and widely used approach is based on Monte Carlo simulations, where the stopping decision is estimated via backward induction \citep{Tsitsiklis2001Regression, Longstaff2001},  which is an (approximate) dynamic programming approach. Another provable approach is based on reinforcement learning (RL) \citep{ Tsitsiklis97optimalstopping, Tsitsiklis2001Regression, Bertsekas2007QlearningOptStop, Li2009Learning,Chen2020ZapQlearningOptStop}.
Both approaches are based on the ordinary least squares approximation which involves choosing basis functions. 
There are many different sets of basis functions available that are commonly used, however, it can be difficult to choose a good set for the considered problem.
Moreover, the number of basis functions often increases polynomially or even exponentially \citep[Section 2.2]{Longstaff2001} in the dimension of the underlying process, making those algorithms impractical for high dimensions.
 
A relatively new approach consists in replacing the basis functions by a neural network and performing gradient descent instead of ordinary least squares  \citep{Kohler2010Pricing, lapeyre2019neural, becker2019deep,becker2019pricing}. The big advantage is that the basis functions do not need to be chosen but are learned instead. 
Compared to using a  polynomial basis, neural networks have the advantage to be dense in any space $L^p(\mu)$, for $1 \leq p < \infty$ and finite measure $\mu$ \citep{hornik1991approximation}, while for polynomials this is only true under certain additional conditions on the measure \citep{bakan2008representation}.
Moreover, in many cases the neural network overcomes the curse of dimensionality, which means that it can easily scale to high dimensions. However, as the neural network is a non convex function with respect to its parameters, the gradient descent does not necessarily converge to the global minimum, while this is the case for the ordinary least squares minimization. Hence, the main disadvantage of those methods is that there are no convergence guarantees without strong and unrealistic assumptions.

In this paper, we propose two neural network based algorithms to solve the optimal stopping problem for Markovian settings: a backward induction and a reinforcement learning approach.
The idea is inspired by randomized neural networks \citep{CAO2018278, huang2006universal}. Instead of learning the parameters of all layers of the neural network, those of the hidden layers are randomly chosen and fixed and only the parameters of the last layer are learned. Hence, the non convex optimization problem is reduced to a convex problem that can be solved with linear regression. 
The hidden layers form random feature maps, which can be interpreted as random basis functions.
In particular, in this paper we show that there is actually no need for complicated or a large number of basis functions.
Our algorithms are based on the methods proposed by  \cite{Longstaff2001}  (backward-induction approach) and \cite{Tsitsiklis2001Regression} (reinforcement learning approach). The difference is that we use a randomized neural network instead of a linear combination of basis functions. However, a randomized neural network can also be interpreted as a linear combination of \emph{random} basis functions. 
On the other hand, our algorithms can also be interpreted as the neural network extensions of these methods, where not the entire neural network but only the last layer is trained.

In addition, we provide a randomized recurrent neural network approach for non-Markovian settings.
We compare our algorithms to the most relevant baselines in terms of accuracy and computational speed in different option pricing problems.
With only a fraction of trainable parameters compared to existing methods, we achieve high quality results considerably faster.

In this work we mainly focus on the well known and important application of optimal stopping strategies to compute lower bounds for American option prices\footnote{
Since American option prices can be formulated as a supremum over stopping times (cf. Section~\ref{sec:Option Price and Optimal Stopping}), this naturally leads to lower bounds of their prices when approximating an optimal stopping strategy (and to the correct price when finding the optimal stopping strategy).
Importantly, computing the lower bound of American option prices in this way, by actually solving the optimal stopping problem (approximately), provides a control algorithm that allows its user to make the decision whether to stop or not, to achieve (approximately) optimal outcomes. 
On the contrary, the dual formulation of American option prices as an infimum over martingales \citep{Rogers2002}, which naturally leads to upper bounds for those prices, is not an optimal stopping problem and therefore their solution does not provide a strategy how to achieve (nearly) optimal outcomes.} 
(as was done e.g.\ by \citet{Tsitsiklis2001Regression, Longstaff2001, Clement2001AnAO, Zanger2009convergence, Zanger2013QuantitativeError, Zanger2018convergence, Zanger2020, lapeyre2019neural}). 
However, we also show in Section~\ref{sec:Upper Bounds for the Optimal Stopping Problem} how our algorithms can be used to compute upper bounds of the American option prices without  additional costs via the dual approach introduced in \cite{Rogers2002} and refined by \cite{becker2019deep}.
Moreover, we show in Section~\ref{sec:Computation of Greeks} that our algorithms can be used to efficiently compute the Greeks of American options.

Finally, we note that our approach is very generic in the sense that it can be applied to any possible type of optimal stopping problem as long as one has access to a sampling method for paths of the underlying process (which should be stopped optimally). In particular, in the case of American option pricing this means that our approach can be applied to any type of underlying stock model, with or without (discrete or continuous) dividends, with positive, negative or stochastic interest rates, and with any type of payoff no matter its complexity.

\section{Optimal stopping via randomized neural networks}
\label{Reservoir optimal stopping via dynamic programming}

One of the most popular and most studied applications of optimal stopping is the pricing of American options.  Hence, we explain our approach in this context.

\subsection{American and Bermudan options}
 An American option gives the holder the right but not the obligation to exercise the option associated with a non-negative payoff function $g$ at any time up to the maturity. 
An American option can be approximated by a Bermudan option, which can be exercised only at some specific dates $t_0 < t_1 < t_2 < \dotsb < t_N$, transforming the continuous-time problem to a discrete one. 
If the time grid is chosen small enough, the American option is well approximated by the Bermudan option. In the case of a Rough Heston model, the convergence rate of the Bermudan option price to the American option price was shown in \cite[Theorem 4.2]{chevalier2021american}.
For equidistant  dates  we simply write $0, 1, 2, \dotsc, N$ instead of $t_0 < t_1 < t_2 < \dotsb < t_N$.

\subsection{Option price and optimal stopping}\label{sec:Option Price and Optimal Stopping}
For $d \in \mathbb{N}$, we assume to have a  $d$-dimensional Markovian stochastic process $(X_t)_{t\geq 0}$ describing the stock prices. 
With respect to a fixed (pricing) probability measure $\mathbb{Q}$, the (superhedging seller's) price of the discretized American option can be expressed through the Snell envelope  described by
\begin{equation}\label{equ:snell envelope 1}
\begin{split}
U_N &\coloneqq g(X_N), \\
U_n &\coloneqq \max\left(g(X_n),  \mathbb{E}[\alpha \, U_{n+1} \, | \, X_n] \right), \quad 0 \leq n < N , 
\end{split}
\end{equation}
where $\alpha$ is the step-wise discounting factor and $g(X_n)$ is assumed to be square integrable for all $n$. Then the (superhedging seller's) price of the option at time $n$ is given by $U_n$ and can equivalently be expressed as the optimal stopping problem
\begin{equation}\label{equ:snell envelope 2}
U_n = \operatorname{sup}_{\tau \in \mathcal{T}_n} \mathbb{E}[\alpha^{\tau - n} g(X_{\tau})  \, | \, X_n],
\end{equation}
where $\mathcal{T}_n$ is the set of all stopping times $\tau \geq n$.
The smallest optimal stopping time is given by 
\begin{equation}\label{equ: optimal stopping time}
\begin{split}
\tau_N &\coloneqq N, \\
\tau_n &\coloneqq \begin{cases}
n, & \text{if } g(X_n) \geq \mathbb{E}[\alpha \, U_{n+1}  \, | \, X_n],\\
\tau_{n+1}, & \text{otherwise}.
\end{cases} 
\end{split}
\end{equation}
In particular, at maturity $N$, the holder receives the final payoff,  and the value of the option $U_N$ is equal to the payoff $g(X_N)$. 
At each time prior to the maturity, the holder decides whether to exercise or not, depending on whether the current payoff $g(X_n)$ is greater than the continuation value $c_n (X_n) \coloneqq 
%\sup_{\tau \in \mathcal{T}_{n+1}} \mathbb{E}[\alpha^{\tau - n} g(X_{\tau})  \, | \, X_n] = 
 \mathbb{E}[\alpha \, U_{n+1}  \, | \, X_n]$. 
Combining expression \eqref{equ:snell envelope 1}, \eqref{equ:snell envelope 2} and \eqref{equ: optimal stopping time}, we can write the price at initial time as
\begin{equation*}
U_0 = \max \left( g(X_0), \mathbb{E}[ \alpha^{\tau_1}g(X_{\tau_1})]  \right).
\end{equation*}
In the following we approximate the price $U_0$ and continuation values $c_n(X_n)$ which are defined theoretically but cannot be computed directly.

\subsection{Monte Carlo simulation and backward recursion}
%\textbf{Monte Carlo simulation and backward recursion.}
We assume to have access to a procedure to sample discrete paths of $X$ under $\mathbb{Q}$. A standard example is that $X$ follows a certain stochastic differential equation (SDE) with known parameters. 
Therefore, we can sample $m$  realizations of the stock price paths, where the $i$-th realization is denoted by $x_{0}, x_{1}^i, x_{2}^i, \dots, x_{N}^i$, with the fixed initial value $x_{0}$. 
For each realization, the cash flow realized by the holder when following the stopping  strategy \eqref{equ: optimal stopping time} is given by the backward recursion 
\begin{align*}
p_N^i &\coloneqq g(x_N^i), \\
p_n^i &\coloneqq \begin{cases}
g(x_n^i), & \text{if } g(x_n^i) \geq c_n(x_n^i),\\
\alpha p_{n+1}^i, & \text{otherwise}.
\end{cases}
\end{align*} 
As $p_1^i$ are samples of $\alpha^{\tau_1 -1}g(X_{\tau_1}) $, we have by the strong law of large numbers that almost surely
\begin{equation}\label{equ:U0}
U_0 = \max \left( g(X_0), \lim_{m \to \infty} \frac{1}{m} \sum_{i=1}^m \alpha p_1^i  \right).
\end{equation}

\subsection{Randomized neural network approximation of the continuation value}
%\textbf{Randomized neural network approximation of the continuation value.} 
For each path $i$ in $\{1,2,\dots, m\}$ and each date $n$ in $\{1,2,\dots, N-1\}$, the continuation value is $c_n(x^{i}_n) = \mathbb{E}[\alpha U_{n+1}|X_n= x^{i}_n]\,,$  where $c_n:\mathbb{R}^d \to \mathbb{R}$ describes the expected value of the discounted price $\alpha U_{n+1}$ if we keep the option until next exercising date $n+1$, knowing the current values of the stocks $X_n$. 
We approximate this continuation value function by a neural network, where only the parameters of the last layer are learned. We refer to such a network as a randomized neural network. Even though the architecture of the neural network can be general, we present our algorithm with a simple dense shallow neural network, where the extension to deep networks is immediate. 
We call $\boldsymbol\sigma:\mathbb{R} \to \mathbb{R}$ the activation function. A common choice is $\boldsymbol\sigma(x) = \tanh(x)$, however, there are many other suitable alternatives.  
For $K \in \mathbb{N}$, we define $\sigma: \mathbb{R}^{K-1} \to \mathbb{R}^{K-1} $, $ \sigma(x) = ( \boldsymbol\sigma(x_1), \dots, \boldsymbol \sigma(x_{K-1}))^{\top}$ for $x \in \mathbb{R}^{K-1}$. 
Let  $\vartheta \coloneqq (A, b) \in \mathbb{R}^{(K-1) \times d}\times \mathbb{R}^{K-1} $ be the parameters of the hidden layer which are randomly and identically sampled and not optimized. 
In general, $A$ and $b$ can be sampled from different distributions that are continuous and have support $\mathbb{R}$.
The distributions and their parameters are hyperparameters of the randomized  neural network that can be tuned.
For simplicity we use a standard Gaussian distribution. 
Let us define 
\begin{equation*}
\phi: \mathbb{R}^d \to \mathbb{R}^{K}, x \mapsto \phi(x) = ( \sigma(A x + b)^\top, 1)^\top\,.
\end{equation*}
and let $\theta_{{n}} \coloneqq ((A_{n})^\top, b_{n})^\top \in \mathbb{R}^{K-1}\times \mathbb{R}$ be the parameters that are optimized. 
Then for each $n$ the continuation value is approximated by
\begin{equation*}
c_{\theta_{{n}}}(x)  \coloneqq  \theta_n^\top \phi(x)  = A_n^\top  \sigma(A x + b) + b_n\,.
\end{equation*}

\subsection{Least squares optimization of last layer's parameters \texorpdfstring{$\boldsymbol\theta_n$}{theta}}
While the parameters $\vartheta$ of the hidden layer are set randomly, the parameters $\theta_n$ of the last layer are found by minimizing the squared error of the difference between conditional expectation of the discounted future price and the approximation function. This is equivalent to finding $\theta_n$ which minimizes $\mathbb{E}[(c_{\theta_n}(x^{i}_{n}, n) - \alpha U_{n+1})^2|X_{n} = x^{i}_{n}]$ for each time $n$ in $\{1,2,\dots,  N-1\}$. The backward recursive Monte Carlo approximation of this expectation at time $n$ yields the loss function
\begin{equation}
\label{loss_function}
\psi_{n}(\theta_{n}) \coloneqq \sum_{i=1}^{m} \left( c_{\theta_{n}}(x^{i}_{n}) - \alpha{p}^{i}_{n+1}\right)^2\,.
\end{equation}
As the approximation function $c_{\theta_n}$ is linear in the parameters $\theta_n$,  the minimizer can be found by ordinary least squares. 
It is given by the following closed form expression, which is well defined under the standard assumptions (see Theorems \ref{thm: conv LS 1} and \ref{thm: conv LS 2})
\begin{equation*}
\theta_{n} = \alpha \left(\sum_{i=1}^{m} \phi(x^{i}_{n})\phi^{\top}(x^{i}_{n})\right)^{-1} \cdot \left(\sum_{i=1}^{m} \phi(x^{i}_{n}) p_{n+1}^i\right) .
\end{equation*}

\subsection{Splitting the data set into training and evaluation set}\label{sec:Splitting the Data Set into Training and Evaluation Set}
The parameters $\theta_n$ are determined using $50\%$ of the sampled paths (training data). Given $\theta_n$, the remaining $50\%$ of the sampled paths (evaluation data) are used to compute the option price. 
By definition, the continuation value $c_n$ is a conditional expectation, which is not allowed to depend on the future values $X_{n+k}$ for $0< k \leq N-n$. 
On the training set, this might not be satisfied, since the loss function \eqref{loss_function} uses the future values $X_{n+k}$. 
In particular, the neural network can suffer from overfitting to the training data, by memorizing the paths, instead of learning the continuation value. This is related to the maximization bias discussed in \cite[Section 6.7]{Sutton2018RL}.
By splitting the data into an independent training and evaluation set, we can however ensure that $c_n$ evaluated on the evaluation set is independent of future values $X_{n+k}$ of the evaluation set.

\subsection{Algorithm}\label{sec:algo 1}
%\textbf{Summary.}
We first sample $2m$ paths and then proceed backwards as follows. At maturity, the pathwise option price approximation is equal to the payoff, which means that $p_N^{i} = g(x_N^{i})$. For each time $n$ in $\{N-1, N-2, \dots, 0\}$, 
we first determine $\theta_n$ as described before using the paths $\{1,2,\dots, m\}$. For all paths $i \in \{1,2, \dots, 2m\}$ we then compare the exercise value $g(x^{i}_{n})$ to the continuation value $c_{\theta_n}(x^{i}_{n})$ and determine the path-wise option price approximation 
at time $n$ as
\begin{equation}
\label{backwardLSM}
p_{n}^{i} =  \underbrace{g(x_{n}^{i})}_{\hbox{payoff}}\underbrace{\mathbf{1}_{\{g(x_{n}^{i})\geq c_\theta(x_{n}^{i})\}}}_{\hbox{exercise}}  + \underbrace{\alpha{p}_{n+1}^{i}}_{\hbox{discounted future price}} \underbrace{\mathbf{1}_{\{g(x_{n}^{i})<c_\theta(x_{n}^{i})\}}}_{\hbox{continue}}\,.\nonumber
\end{equation}
Finally, the second half of the paths $\{m+1,\dots, 2m\}$ is used to compute the option price approximation $ p_0=\max(g(x_0), \frac{1}{m}\sum_{i=m+1}^{2m}  \alpha p_1^i)$. We call this algorithm, which is presented in  Algorithm~\ref{algo:1}, randomized least squares Monte Carlo  (RLSM).

%\begin{wrapfigure}{r}{0.53\textwidth}
%\begin{minipage}{0.53\textwidth}
%\vspace{-0.8cm}
\begin{algorithm}
   \caption{Optimal stopping via  randomized least squares Monte Carlo (RLSM) }
   \label{algo:1}
\begin{algorithmic}
   \STATE {\bfseries Input:}  discount factor $\alpha$,  initial value $x_0$
   \STATE {\bfseries Output:} price $p_0$
  \STATE {\bfseries 1:} sample a random matrix $A \in{\mathbb{R}^{ (K-1) \times d}}$ and a random vector $b \in \mathbb{R}^{K-1}$
  \STATE {\bfseries 2:} simulate $2m$ paths of the underlying process $(x_1^i, \dots, x_N^i)$ for $i \in \{1, \dots, 2m\}$
 \STATE {\bfseries 3:} for each path $i \in \{1, \dots, 2m\}$, set $p_N^i=g(x_N^i)$
\STATE {\bfseries 4:} for each time $n \in \{N-1, \dots, 1\}$
   \STATE \quad {\bfseries a:} 
   for each path $i \in \{1, \dots, 2m\}$, set $\phi(x_n^i) = (\sigma(A x_n^i +b)^\top,1)^\top \in \mathbb{R}^K$
   \STATE \quad  {\bfseries b:} set $ \theta_{n} =   \alpha \left(\sum_{i=1}^{m} \phi(x^{i}_{n})\phi^{\top}(x^{i}_{n})\right)^{-1}  \left(\sum_{i=1}^{m} \phi(x^{i}_{n}) p_{n+1}^i\right)$
   \STATE \quad   {\bfseries c:} for each path $i \in \{1, \dots, 2m\}$
   \STATE \qquad \quad set $ p_n^i=g(x_n^i)\textbf{1}_{g(x_n^i) \geq \theta_n^\top \phi(x_n^i) } + \alpha p_{n+1}^i \textbf{1}_{g(x_n^i) < \theta_n^\top \phi(x_n^i)} $
\STATE {\bfseries 5:} set $p_0=\max(g(x_0),\frac{1}{m}\sum_{i=m+1}^{2m} \alpha p_1^i )$
\end{algorithmic}
\end{algorithm}
%\vspace{-25pt}
%\end{minipage}
%\end{wrapfigure}

\subsection{Guarantees of convergence} 
%\textbf{Guarantees of convergence.} 
We present results that guarantee convergence of the price computed with our algorithm to the correct price of the discretized American option. The formal results with precise definitions and proofs are given in Appendix~\ref{sec:Convergence of the Longstaff-Schwartz version}. In contrast to comparable results for neural networks \citep{lapeyre2019neural, becker2019deep, becker2019pricing}, our results do not need the assumption that the optimal weights are found  by some optimization scheme like stochastic gradient descent. 
Instead, our algorithms imply that the optimal weights are found and used.

\begin{theorem}[informal]
\label{thm:informal}
As the number of sampled paths $m$ and the number of random basis functions $K$ go to $\infty$, the price $p_0$ computed with Algorithm \ref{algo:1} converges to the correct price of the Bermudan option.
\end{theorem}

\subsection{Possible extensions}\label{sec:Possible extensions}
%\textbf{Possible extensions.}
When the set of pricing measures $\mathcal{Q}$ has more than one element (in case of an incomplete market), the option price is given by $\sup_{\mathbb{Q} \in \mathcal{Q}} U_0^{\mathbb{Q}}$, where $U^{\mathbb{Q}}$ is defined as in \eqref{equ:snell envelope 1}.
%the Snell Envelope computed with respect to $\mathbb{Q} \in \mathcal{Q}$.
Assuming that we can sample from a finite subset $\mathcal{Q}_1 \subset \mathcal{Q}$, this price can be approximated by first computing the price for each measure in $\mathcal{Q}_1$ and then taking the maximum of them.

For simplicity we assume that the payoff function only takes the current price as input, however, all our methods and results stay valid if $g(X_n)$ is replaced by a square integrable $\mathcal{F}_n$-measurable random variable $Z_n$, where $\mathcal{F}_n$ denotes the information available up to time $n$.
In the case that $(Z_n)_{1 \leq n \leq N}$ is Markov,  Algorithm~\ref{algo:1} and Algorithm~\ref{algo:2} (Section~\ref{sec:optimal stopping via reinforcement learning}) can be used,  otherwise Algorithm~\ref{algo:RNN} (Section~\ref{sec:Optimal Stopping via Randomized Recurrent Neural Networks for Non-Markovian Processes}) has to be used, to deal with the path dependence.  In the following sections we stick to the notation $g(X_n)$ for the payoff,  keeping in mind that the extension to a general $Z_n$ is also valid there.

Similarly, a more general discounting can be incorporated, by assuming that $Z_n$ is given in discounted terms and setting $\alpha=1$.

\section{Optimal stopping via randomized reinforcement learning}
\label{sec:optimal stopping via reinforcement learning}
In order to avoid approximating the continuation value at each single date $n\in \{1, \dots, N-1 \}$ with a different function, as it is done in Section~\ref{Reservoir optimal stopping via dynamic programming}, we can directly learn the continuation  function which also takes the time as argument. Hence, instead of having a different function $c_{\theta_n} (x_n^i)$ for each date $n$, we learn one function which is used for all dates $n$. As previously, we define the parameters of the hidden layer $\vartheta \coloneqq (A, b) \in \mathbb{R}^{ (K-1) \times (d+2)}\times \mathbb{R}^{K-1} $, which are randomly chosen and not optimized,  and $\phi: \mathbb{R}^{d+2} \to \mathbb{R}^{K}$, $\phi(n, x) = ( \sigma(A \tilde{x}_n + b)^\top, 1)^\top$, where  $\tilde{x}_n = (  n, N-n, x_n^\top)^\top$. Let  $\theta \in  \mathbb{R}^K$ define the parameters that are optimized, then the continuation value is approximated by
$$c_{\theta}(n, x)  \coloneqq  \theta^\top \phi(n,x)\,.$$
Instead of having a loop backward in time with $N$ steps, we iteratively improve the approximation  $c_{\theta}$. More precisely, we start with some (random) initial weight $\theta_0$ and then iteratively improve it by minimizing the difference between the continuation function $c_{\theta_\ell}$ and the prices $p$ computed with the previous weight $\theta_{\ell-1}$. 
Moreover, differently than in Section \ref{Reservoir optimal stopping via dynamic programming}, we use the continuation value for the decision whether to continue \emph{and} for the approximation of the discounted future price, as in \citep{Tsitsiklis2001Regression}.
This second algorithm can be interpreted as a randomized fitted Q-iteration   (RFQI) and is presented in Algorithm \ref{algo:2}.
It is a very simple type of reinforcement learning, where the agent has only two possible actions and the agent's decision does not influence the transitions of the state. In particular  the agent's decision does not influence the evolution of the underlying stocks.
As a reinforcement learning method, 
it is based on the assumption that the optimization problem can be modeled by a Markov decision process. 
%\begin{wrapfigure}{r}{0.50\textwidth}
%\begin{minipage}{0.50\textwidth}
%\vspace{-0.8cm}
\begin{algorithm}
   \caption{Optimal stopping  via randomized fitted Q-Iteration (RFQI)}
   \label{algo:2}
\begin{algorithmic}
   \STATE {\bfseries Input:}  discount factor $\alpha$,  initial value $x_0$
   \STATE {\bfseries Output:} price $p_0$
  \STATE {\bfseries 1:} sample a random matrix $A \in{\mathbb{R}^{ (K-1)  \times (d+2)}}$ and a random vector $b \in \mathbb{R}^{K-1}$
  \STATE {\bfseries 2:} simulate $2m$ paths of the underlying process $(x_1^i, \dots, x_N^i)$ for $i \in \{1, \dots, 2 m\}$
  \STATE {\bfseries 3:} initialize weights $\theta_0 = 0 \in \mathbb{R}^K$ and set $\ell = 0$
 \STATE {\bfseries 4:} until convergence of $\theta_\ell$
\STATE \quad  {\bfseries a:}  for each path $i \in \{1, \dots, 2m\}$ 
%\STATE \qquad  {\bfseries i:}  simulate the $i$-th path of the underlying process \STATE \qquad \qquad$(x_0^i, \dots, x_N^i)$ 
\STATE \qquad  {\bfseries i:} set $p_N^i=g(x_N^i)$
   \STATE \qquad  {\bfseries ii:} 
   for each date $n \in \{1, \dots, N-1\}$ 
   \STATE \qquad \qquad set   $\phi(n, x_n^i) = (\sigma(A \tilde x_n^i +b),1) \in \mathbb{R}^K$ 
 \STATE \qquad \qquad  set $p_n^i= \max(g(x_n^i),\phi(n,x_n^i)^\top \theta_{\ell}) $

\STATE  \quad {\bfseries b:}
  set $$ \theta_{\ell+ 1} = \alpha \left(\sum_{n=1}^{N}\sum_{i=1}^{m} \phi(n,x^{i}_{n})\phi^{\top}(n,x^{i}_{n})\right)^{-1}  \left(\sum_{n=1}^{N}\sum_{i=1}^{m} \phi(n,x^{i}_{n}) p_{n+1}^i\right) \in \mathbb{R}^K$$
\STATE  \quad {\bfseries c:} set $\ell \leftarrow \ell + 1$
\STATE {\bfseries 5:} set $p_0=\max(g(x_0),\frac{1}{m}\sum_{i=m+1}^{2m}\alpha p_1^i)$ 
\end{algorithmic}
\end{algorithm}
%\vspace{-20pt}
%\end{minipage}
%\end{wrapfigure}

\subsection{Guarantees of convergence.}
%\textbf{Guarantees of convergence.} 
We present results that guarantee convergence of the price computed with our algorithm to the correct price of the discretized American option. The formal results with precise definitions and proofs are given in Appendix~\ref{sec:Convergence of reservoir optimal stopping via reinforcement learning}.

\begin{theorem}[informal]\label{thm2:informal}
As the number of iterations $L$, the number of sampled paths $m$ and the number of random basis functions $K$ go to $\infty$, the price $p_0$ computed with Algorithm \ref{algo:2} converges to the correct price of the Bermudan option.
\end{theorem}

\section{Optimal stopping via randomized recurrent neural networks for non-Markovian processes}
\label{sec:Optimal Stopping via Randomized Recurrent Neural Networks for Non-Markovian Processes}
For non-Markovian processes, for each date $n$,
 the continuation function is no longer a function of the last stock price, $c_n(X_n)$, but a function depending on the entire history $c_n(X_0, X_1, \dots, X_{n-1}, X_{n})$. 
More precisely, the continuation value is now defined by $c_n:= \mathbb{E}[\alpha g(X_{n+1})  \, | \, \mathcal{F}_n]$ where $\mathcal{F}_n$ denotes the information available up to time $n$.  Therefore, we replace the randomized feed-forward neural network by a randomized recurrent neural network (randomized RNN), which can utilize the entire information of the path up to the current time $(x_0, x_1, \dots, x_{n-1}, x_n )$. In particular, we define the parameters of the hidden layer $\vartheta \coloneqq (A_x,A_h, b) \in \mathbb{R}^{(K-1) \times d}\times \mathbb{R}^{(K-1)\times (K-1)} \times\mathbb{R}^{K-1} $, which are randomly sampled and not optimized. Their distributions and parameters, which don't have to be the same for $A_x$ and $A_h$, are hyperparameters that can be tuned. These tuning parameters are more important in this case, as they determine the interplay between past and new information. Moreover, we define 
\begin{equation*}
\begin{split}
\phi: &\mathbb{R}^d \times \mathbb{R}^K\to \mathbb{R}^{K+1}, \quad
(x, h) \mapsto \phi(x, h) = ( \sigma(A_x x+ A_h h + b)^\top, 1)^\top
\end{split}
\end{equation*}
and  $\theta_{{n}} \coloneqq ((A_{n})^\top, b_{n})^\top \in \mathbb{R}^{K-1}\times \mathbb{R}$, the parameters that are optimized. Then for each $n$, the continuation value is recursively approximated by
\begin{equation}\label{equ:Randomized recurrent NN}
\left\{
    \begin{array}{llll}
h_{n} &\coloneqq&       \sigma(A_x x_n +A_h h_{n-1} + b), \\
c_{\theta_{n}}(h_n) &\coloneqq&  A_n^\top  h_{n}  + b_n   = \theta_n^\top  \phi(x_n, h_{n-1}) ,
    \end{array}
\right.
\end{equation}
%\begin{equation*}
%\left\{
%    \begin{array}{lllll}
%h_{n} &=&       \sigma(A_x x_n +A_h h_{n-1} + b)\\
%c_{\theta_{n}}(h_n) &=&  A_n^\top  h_{n}  + b_n   &=& \theta_n^\top  \phi(x_n, h_{n-1}) 
%    \end{array}
%\right.
%\end{equation*}
with $h_{-1} \coloneqq 0$. We call this algorithm, which is presented in Algorithm~\ref{algo:RNN}, randomized recurrent least squares Monte Carlo (RRLSM).

%\begin{wrapfigure}{R}{0.50\textwidth}
%\begin{minipage}{0.50\textwidth}
%\vspace{-0.8cm}
\begin{algorithm}
   \caption{Optimal stopping via randomized recurrent neural network (RRLSM)
	}
   \label{algo:RNN}
\begin{algorithmic}
   \STATE {\bfseries Input:}  discount factor $\alpha$,  initial value $x_0$, initial latent variable $h_{-1}=0$
   \STATE {\bfseries Output:} price $p_0$
  \STATE {\bfseries 1:} sample random matrices $A_x \in{\mathbb{R}^{(K-1)\times d}}$,  $A_h \in{\mathbb{R}^{(K-1)\times (K-1)}}$ and a \STATE random vector $b \in \mathbb{R}^{K-1}$ 
 \STATE {\bfseries 2:} simulate $2m$ paths of the underlying process 
 $(x_1^i, \dots, x_N^i)$ for $i \in \{1, \dots, 2m\}$
 \STATE {\bfseries 3:} for each path $i \in \{1, \dots, 2m\}$, set $p_N^i=g(x_N^i)$
 \STATE {\bfseries 4:} for each date $n \in \{0, \dots, N-1\}$
   \STATE \quad {\bfseries a:} 
   for each path $i \in \{1, \dots, 2m\}$, set $h_{n}^i =       \sigma(A_x x_n^i +A_h h_{n-1}^i + b)$
\STATE {\bfseries 5:} for each date $n \in \{N-1, \dots, 1\}$
   \STATE \quad {\bfseries a:} 
   for each path $i \in \{1, \dots, 2m\}$, set $\phi_n^i = ((h_n^i)^\top, 1)^\top \in \mathbb{R}^K$
   \STATE \quad  {\bfseries b:}  set $ \theta_{n} =   \alpha \left(\sum_{i=1}^{m} \phi_n^i(\phi_n^i)^{\top}\right)^{-1}  \left(\sum_{i=1}^{m} \phi_n^i p_{n+1}^i\right)$
   \STATE \quad   {\bfseries c:} for each path $i \in \{1, \dots,2 m\}$  
   \STATE \qquad \quad  set $ p_n^i=g(x_n^i)\textbf{1}_{g(x_n^i) \geq \theta_n^\top \phi_n^i}+ \alpha p^i_{n+1} \textbf{1}_{g(x_n^i) < \theta_n^\top \phi_n^i} $
\STATE {\bfseries 6:} set $p_0=\max(g(x_0),\frac{1}{m}\sum_{i=m+1}^{2m} \alpha p_1^i )$
\end{algorithmic}
\end{algorithm}

\subsection{Guarantees of convergence} 
We present results that guarantee convergence of the price computed with our algorithm to the correct price of the discretized American option. The formal results with precise definitions and proofs are given in Appendix~\ref{sec:Convergence of the recurrent Longstaff-Schwartz version}.

\begin{theorem}[informal]
\label{thm3:informal}
As the number of sampled paths $m$ and the number of random basis functions $K$ go to $\infty$, the price $p_0$ computed with Algorithm \ref{algo:RNN} converges to the correct price of the Bermudan option.
\end{theorem}

\section{Upper bounds for American option prices}
\label{sec:Upper Bounds for the Optimal Stopping Problem}

So far we have approximated the value of an American option by computing the optimal stopping time \eqref{equ: optimal stopping time} with which the value of the Snell envelope \eqref{equ:snell envelope 2} can be determined. Since our algorithms only \emph{approximate} the optimal stopping time of this maximisation problem, the resulting price is a lower bound for the true value. 
The advantage of this method is that it not only provides an approximation for the price, but also a decision rule when to stop.

An upper bound of the true value is naturally implied by the dual method introduced by \cite{Rogers2002}. In particular, \cite[Theorem 2.1]{Rogers2002} yields that the starting value of the Snell envelope \eqref{equ:snell envelope 2}, which is the price of the American option, can equivalently be written as the minimisation problem 
\begin{equation}\label{equ:dual formulation Snell envelope}
    U_0 = \inf_{M \in \mathcal{M}_0} \mathbb{E}\left[ \sup_{0 \leq n \leq N} (Z_n - M_n) \right],
\end{equation}
where $Z_n$ denotes the \emph{discounted} payoff process (cf.\ Section~\ref{sec:Possible extensions}) and $\mathcal{M}_0$ is the set of $(\mathcal{F}_n)$-martingales starting at $0$.
As explained in \citep[Section 3.2]{becker2019deep}, the minimiser of \eqref{equ:dual formulation Snell envelope} is given by the martingale part of the Doob-Meyer decomposition of the Snell envelope $(\tilde{U}_n)_{0 \leq n \leq N}$ for the discounted payoff process. This martingale is defined through
\begin{equation}\label{equ:optimal martingale}
\begin{split}
	M_0^U &\coloneqq 0, \\
    M_n^U - M_{n-1}^U &\coloneqq \tilde{U}_n - \mathbb{E}[\tilde{U}_n | \mathcal{F}_{n-1}] = \max(Z_n, \mathbb{E}[\tilde{U}_{n+1} | \mathcal{F}_{n}]) - \mathbb{E}[\tilde{U}_n | \mathcal{F}_{n-1}],
\end{split}
\end{equation}
where we used (the discounted version of) \eqref{equ:snell envelope 1} for the last equality.
Since our algorithms compute the continuation values $c_n = \mathbb{E}[\tilde{U}_{n+1} | \mathcal{F}_{n}] $ (written in the most general form; cf. Section~\ref{sec:Optimal Stopping via Randomized Recurrent Neural Networks for Non-Markovian Processes}), we can compute an upper bound approximation of the option price together with the lower bound nearly without additional costs\footnote{We only need the additional computation and storage of the martingale differences $M_n^U - M_{n-1}^U$, the computation of $M^U$ as its cumulative sum and finally the computation of the maximum over $Z_n - M_n^U$. Since no additional loop or simulation is needed, the computational costs stay nearly the same. In particular, this may be considered as an advantage over the algorithm presented by \cite{becker2019deep}, where the computations of the lower bound as well as additional time consuming computations (approximately doubling the total computation time) are needed to get an approximation of the upper bound.} via \eqref{equ:dual formulation Snell envelope} and \eqref{equ:optimal martingale} as
\begin{equation*}
    U_0 = \mathbb{E}\left[ \sup_{0 \leq n \leq N} (Z_n - M_n^U) \right],
\end{equation*}
with 
\begin{equation*}
    M_0^U = 0, \quad  M_n^U - M_{n-1}^U =  \max(Z_n, c_n) - c_{n-1}.
\end{equation*}
Moreover, given our methods to compute the lower and upper bound of an American option price, confidence intervals can be computed exactly as described in \citep[Section~3.3]{becker2019deep}.

Algorithm~\ref{algo:4} is the extension of Algorithm~\ref{algo:1}, where the upper bound for the option price is computed additionally to the lower bound. The extensions of Algorithms~\ref{algo:2} and \ref{algo:RNN}  work similarly.

\begin{algorithm}
   \caption{Upper and lower bounds for Bermudan option price via randomized least squares Monte Carlo (RLSM) }
   \label{algo:4}
\begin{algorithmic}
   \STATE {\bfseries Input:}  discount factor $\alpha$,  initial value $x_0$
   \STATE {\bfseries Output:} lower price  bound $p_0^l$, upper price bound $p_0^u$
  \STATE {\bfseries 1:} sample a random matrix $A \in{\mathbb{R}^{ (K-1) \times d}}$ and a random vector $b \in \mathbb{R}^{K-1}$
  \STATE {\bfseries 2:} simulate $2m$ paths of the underlying process $(x_1^i, \dots, x_N^i)$ for $i \in \{1, \dots, 2m\}$ and compute the discounted payoffs $z_n^i = g(x_n^i)*\alpha^n$ for $1 \leq i \leq 2m$, $1 \leq n \leq N$
 \STATE {\bfseries 3:} for each path $i \in \{1, \dots, 2m\}$, set $p_N^i=z_N^i$ and $c_N^i = 0$
\STATE {\bfseries 4:} for each time $n \in \{N-1, \dots, 0\}$
   \STATE \quad {\bfseries a:} 
   for each path $i \in \{1, \dots, 2m\}$, set $\phi(x_n^i) = (\sigma(A x_n^i +b)^\top,1)^\top \in \mathbb{R}^K$
   \STATE \quad  {\bfseries b:} set $ \theta_{n} =    \left(\sum_{i=1}^{m} \phi(x^{i}_{n})\phi^{\top}(x^{i}_{n})\right)^{-1}  \left(\sum_{i=1}^{m} \phi(x^{i}_{n}) p_{n+1}^i\right)$
   \STATE \quad   {\bfseries c:} for each path $i \in \{1, \dots, 2m\}$
   \STATE \qquad {\bfseries i:} set $c_n^i = \theta_n^\top \phi(x_n^i)$
   \STATE \qquad {\bfseries ii:} set $ p_n^i=z_n^i \textbf{1}_{z_n^i \geq c_n^i } + p_{n+1}^i \textbf{1}_{z_n^i < c_n^i } $
   \STATE \qquad {\bfseries iii:} set $\Delta M_{n+1}^i = \max(z_{n+1}^i, c_{n+1}^i) - c_{n}^i $
\STATE {\bfseries 5:} set $p_0^l =\max(g(x_0),\frac{1}{m}\sum_{i=m+1}^{2m} p_1^i )$
\STATE {\bfseries 6:} set $p_0^u = \frac{1}{m}\sum_{i=m+1}^{2m}\left( \max_{0 \leq n \leq N}(z_n^i - \sum_{k=1}^n \Delta M_k^i ) \right) $
\end{algorithmic}
\end{algorithm}

\section{Related work}
We present the most relevant approaches for the optimal stopping problem: backward induction either with basis functions or with neural networks and reinforcement learning. Moreover, we explain the connection of our algorithms to randomized neural networks and reservoir computing techniques.

\subsection{Optimal stopping}
Numerous works studied the optimal stopping problem via different approaches. A common approach consists in using a  regression based method to estimate the continuation value   \citep{Tilley1993,Barraquand1995, Carriere1996, Tsitsiklis97optimalstopping,   Tsitsiklis2001Regression,Longstaff2001, Schweizer2002Bermudan, Boyle2003, Broadie2004, Kolodko2004iterativeconstruction, Egloff2007, Jain2015}, or the optimal stopping boundary \citep{Pham1997OptStop,Andersen1999, Garcia2003}. A different approach uses quantization \citep{Bally2003Discrete, Bally2005PricingHedging}. A dual approach was developed and extended in \citep{ Rogers2002, Haugh2004, Rogers2010}.
\cite{bank2019lenglarts} studied Lenglart's Theory of Meyer-sigma-fields and El Karoui's Theory of Optimal Stopping \citep{ElKaroui1981controle}. 
 An in depth review of the different methods is given in \citep{Bouchard2012, Pages2018Book}.

\subsubsection{Optimal stopping via backward induction}
One of the most popular approaches are the backward induction methods introduced by \cite{Tsitsiklis2001Regression} and \cite{Longstaff2001}.  
\cite{Tsitsiklis2001Regression} uses the approximated continuation value to estimate the current price, by using the backward recursion
\begin{equation}
\label{eq:tsitsiklis_nax}
{p}_{n}^i = \max (g(x_{n}^i), c_{\theta_n}(x_{n}^i))\,.
\end{equation}
Instead, \cite{Longstaff2001} uses the continuation value only for the decision to stop or to continue, yielding
\begin{equation}
\label{eq:LS}
 {p}_{n}^i = \left\{
    \begin{array}{ll}
        g(x_{n}^i), & \mbox{if } g(x_{n}^i) \geq c_{\theta_{n}}(x^i_{n})\\
        \alpha p_{n+1}^i,  & \mbox{otherwise.}
    \end{array}
\right. 
\end{equation} 
The second algorithm is more robust, as the approximation is only used for the decision and not for the estimation of the price. Hence, the method proposed by \cite{Longstaff2001} is the most used method in the financial industry and can be considered state-of-the-art. 
In both papers, the approximation $c_{\theta}(x^{i}_n)   =  \theta^\top \phi(x^i_n)$ is used, where $\phi = (\phi_1, \dotsc, \phi_K)$ is a set of $K$ basis functions and $\theta \in \mathbb{R}^K$ are the trainable weights. 
Possible choices for the basis functions proposed in \cite{Longstaff2001} are Laguerre, Hermite, Legendre, Chebyshev, Gegenbauer, and Jacobi polynomials.
While they have the advantage of having convergence guarantees, both algorithms do not easily scale to high dimensional problems since the number of basis functions usually grows polynomially or even exponentially \citep[Section 2.2]{Longstaff2001} in the number of stocks. One direction of research to overcome this problem is to apply dimension reduction techniques \citep{bayer2021pricing}.

\subsubsection{Optimal stopping via backward induction using neural networks.}
Another idea to overcome this issue was proposed by \cite{Kohler2010Pricing}, which consists in approximating the continuation value by a neural network
\begin{equation*}
f_\theta(x^i_{n})  \approx c_\theta(x^i_{n}).
\end{equation*} 
That way, the features are learned contrary to the basis functions which must be chosen.
While \cite{Kohler2010Pricing} use the backward recursion \eqref{eq:tsitsiklis_nax} introduced by \cite{Tsitsiklis2001Regression}, both \cite{lapeyre2019neural} and \cite{becker2019pricing} use the backward recursion \eqref{eq:LS} suggested by  \cite{Longstaff2001}. Instead of approximating the continuation value, \cite{becker2019deep} suggested to approximate the whole indicator function present in \eqref{eq:LS} by a neural network $ f_{\theta_{n}}(x_{n}^{i})\approx \mathbf{1}_{\{g(x_{n}^{i})\geq c(x_{n}^{i})\}} $. 
Then, the current price is estimated by
\begin{eqnarray}
\label{backwardLSM_Deep_optimal}
p_n^{i} &=& g(x_n^{i})\underbrace{f_{\theta_n}(x_n^{i})}_{\hbox{stop}}  + \alpha{p}_{n+1}^{i} \underbrace{\left(1-f_{\theta_k}(x_n^{i})\right)}_{\hbox{continue}}\,. \nonumber
\end{eqnarray}
Moreover, instead of minimizing the loss function \eqref{loss_function}  in order to find a good approximation of the continuation function, \cite{becker2019deep} optimize the parameters by directly maximizing the option price $
\psi_n(\theta_{n}) = \frac{1}{m} \sum_{i=1}^{m} \alpha p^{i}_{n}$. 

All those algorithms use stochastic gradient methods to determine the parameters of the neural networks. They have to find the parameters of $N-1$ neural networks (using a different neural network for each date). Since they use stochastic gradient methods with a non-convex  loss function they cannot provide theoretical convergence guarantees, without the strong assumption that they actually find the optimal parameters.

\subsubsection{Optimal stopping via reinforcement learning} 
By its nature, reinforcement learning is closely related to the dynamic programming principle as shown in \citep{Sutton2018RL, Bertsekas1996NeuroDynamic}. Moreover, the optimal stopping problem is well studied as an application of reinforcement learning  \citep{Tsitsiklis97optimalstopping, Tsitsiklis2001Regression, Bertsekas2007QlearningOptStop, Li2009Learning}. 
In all those methods, a linear approximator is used (linear combination of basis functions), similarly to the LSM method \citep{Longstaff2001}. If a standard set of basis functions  that grows polynomially in the dimension is used, then these methods suffer from the curse of dimensionality. In particular, they cannot practically be scaled to high dimensions as can be seen in our numerical results.
To the best of our knowledge,   our approach constitutes the first time that randomized neural networks are used to approximate the value function in  reinforcement learning.

\subsection{Randomized neural networks and reservoir computing}
In RLSM  and RFQI  we use a  neural network  with randomly sampled and fixed hidden layers, where only the last layer is reinitialized and trained at each time $n \in \{N-1, \dots, 1\}$. The architecture used at each time can be interpreted as a neural network with random weights (NNRW) studied and reviewed in \citep{CAO2018278}, where a universality result was provided in \citep{huang2006universal}. 
Randomized neural networks as approximation functions were also studied by \cite{Gorban2016ApproxRandBases}.

Randomized recurrent neural networks are an extension of randomized neural networks. 
A recurrent neural network (RNN) where the parameters are randomly generated and fixed and only the readout map is trained, is known as reservoir. 
Reservoir computing not only reduces the computation time, but also outperforms classical, fully trained RNNs in many tasks \citep{Schrauwen07ReservoirOverview, Verstraeten2007ReservoirUnification, Lukosevicius2009ReservoirRNN, Gallicchio2017DeepReservoirCritical}. 
Similarly as in  reservoir computing, in our randomized recurrent neural network algorithm RRLSM,  the parameters of the hidden layers are randomly sampled and fixed thereafter. However, while reservoir computing trains only one readout map which has the same parameters for all times, we train a different readout map for each single time $n \in \{N-1, \dots, 1\}$, similarly to RLSM. 

\subsection{Backward induction versus reinforcement learning}
Backward induction is an (approximate) dynamic programming (ADP) approach. While \cite{Sutton2018RL} regards ADP as a class of RL algorithms, we distinguish these two approaches in this work, because of their different algorithmic structure and their different ways of using the training data. 
In particular, backward recursion computes the approximation of the continuation value for each date sequentially. 
More precisely, it starts at the final date and goes backward in time.
For the approximation at each date, only the data of this date is used.
In contrast to this, RL starts with an initial approximation that is applied for all dates and iteratively improves this approximation.  This way, the data of all dates is used to improve the approximation of all dates. In comparison to backward recursion, this can be interpreted as a type of transfer learning between the dates.

\section{Experiments}
\label{Experiments}

There are numerous ways to empirically evaluate optimal stopping approaches. We choose the most studied settings that were considered in the American option pricing literature. In particular, we only consider synthetic data. Applications to real data involve model calibration, which is an independent problem and finally results in  applying the optimal stopping algorithm to synthetically generated data again.

Besides our algorithms, we also implemented the baselines and provided all of them at \url{https://github.com/HeKrRuTe/OptStopRandNN}.

\subsection{Experimental setup}
In our experiments, we mainly focus on the computation and comparison of lower bound approximations of American option prices. However, in Section~\ref{sec:Computation of Upper Bounds} we present experiments where also upper bound approximations are computed with RLSM\ based on the derivation in Section~\ref{sec:Upper Bounds for the Optimal Stopping Problem}.

The evaluation of all the algorithms was done on the same computer, a dedicated machine with $2\times$Intel Xeon CPU E5-2697 v2 (12 Cores) 2.70GHz and 256 GiB of RAM.

\subsubsection{Baselines (LSM, NLSM, DOS and FQI)}
We compare RLSM  and RFQI  to three backward induction algorithms and one reinforcement learning approach. First, the state-of-the-art least squares Monte Carlo (LSM) \citep{Longstaff2001}. Second, the algorithm proposed by \cite{lapeyre2019neural}, where the basis functions are replaced by a deep neural network (NLSM). Third, the deep optimal stopping (DOS) \citep{becker2019deep},  where instead of the continuation value the whole indicator function of the stopping decision is approximated by a neural network.
And finally, the fitted Q-iteration (FQI) presented  as the second algorithm in \citep{Tsitsiklis97optimalstopping}. \cite{Li2009Learning} studied and compared two reinforcement learning based methods (FQI and LSPI) to solve the optimal stopping problem. Since FQI always worked better in our experiments, we only show comparisons to this algorithm.
Our aim is to compare the main concepts of all the different algorithms in a fair way, hence we leave away certain (more sophisticated) particularities unique to each of them.

\subsubsection{Choice of basis functions for the baselines} 
There are many possible choices for the set of basis functions. \cite{Longstaff2001}   proposed to use the first three weighted Laguerre polynomials for LSM and  \cite{Li2009Learning} added three additional basis functions of the date for FQI. While the size of this set of basis functions scales linearly with the dimension,  it does not include any interaction terms. 
The classical polynomial basis functions up to the second order are the easiest way to include coupling terms in the basis. To deal with the time dependence of FQI, the relative date $t/T$ and $1-t/T$ are added as additional coordinates to the $d$-dimensional stock vector. 
The size of this basis grows quadratically in the dimension $d$, i.e. it has $1+2d+d(d-1)/2$ elements for LSM and for FQI $d$ is replaced by $d+2$. 
The results obtained with the classical polynomials up to degree two were better than with the weighted Laguerre polynomials for LSM and FQI, therefore we only present these results in our tables. For large $d$ the computations of LSM and FQI did not terminate within a reasonable amount of time (several hours) and therefore were aborted.

\subsubsection{No regularization for LSM and FQI}
While drastically increasing the number of hidden nodes without increasing the number of paths or applying penalization led to overfitting for RLSM  and RFQI, this was not observed for LSM and FQI. In particular, for LSM Ridge regression ($L^2$-penalisation) was tested without leading to better results than standard linear regression. Moreover, comparing the results of FQI, RFQI  and DOS for growing dimensions shows that overfitting does not become a problem when more basis functions are used. Therefore, also for FQI standard linear regression was used as suggested by \cite{Tsitsiklis97optimalstopping}.

\subsubsection{Architecture of neural networks} 
In order to have a fair comparison in terms of accuracy and in terms of computation time, we use the same number of hidden layers and nodes per layer for all the algorithms.
\begin{itemize}
\item We observed that one hidden layer was sufficient to have a good accuracy  (an increase of the number of the hidden layers did not lead to better accuracy). Therefore,  NLSM, DOS, and all algorithms that we proposed have only one hidden layer.
\item We use $20$ nodes for the hidden layer. 
Importantly, we do not claim that this choice is optimal for any of the methods.
For RFQI   the number of nodes is set to the minimum between $20$ and the number of stocks, for stability reasons. 
\item 
Leaky ReLU is used for RLSM  and RFQI  and tanh for the randomized recurrent neural network RRLSM. For NLSM and DOS, we use the suggested activation functions, Leaky ReLU for NLSM and ReLU and sigmoid for DOS.
\item The parameters $(A, b)$ of the random neural networks of RLSM  and RFQI   are sampled using a standard normal distribution with mean $0$ and standard deviation $1$. 
Different hyper-parameters were tested, but they didn't have a big influence on the results so we kept the standard choice.

\item For the randomized recurrent neural network of RRLSM, we use a standard deviation of $0.0001$ for $A_x$ and $0.3$ for $A_h$.
Also here different hyper-parameters were tested, and the best performing were chosen and used to present the results. The same holds for tested path-dependent versions of RFQI, however, none of the hyper-parameters performed very well as shown below.

\item Some of the reference methods suggest to use the payoff as additional input, while others do not or leave this open. Therefore, we tested  using the payoff as input and not using it for each method in each experiment.  We came to the conclusion that the backward induction algorithms  (LSM, DOS, NLSM, RLSM) usually work slightly better with, while the reinforcement learning algorithms (FQI, RFQI) usually  work slightly better without the payoff.
Hence, we  show  these results. 

\item As suggested by the authors, we used batch normalization for the implementation of DOS.  
\end{itemize}

\subsection{The Markovian case -- Bermudan option pricing}

First we evaluate RLSM  and RFQI  in the standard Markovian setting of Bermudan option pricing with different stock price models and payoff functions.

\subsubsection{Stock models (Black--Scholes and Heston)}

We test our algorithm on two multidimensional stochastic models, Black--Scholes and Heston with fixed parameters and standard discounting factor $\alpha = e^{-rT/N}$. For each model we sample $m=20'000$ paths on the time interval $[0,1]$, i.e.,  with maturity $T=1$, using the Euler-scheme with $N=10$ equidistant dates.
As explained in Section \ref{sec:Splitting the Data Set into Training and Evaluation Set}, we use half of the paths as training set and the second half to compute the approximated price using the trained continuation value respectively decision function.

The Black--Scholes model for a max call option is a widely used example in the literature \citep{Longstaff2001,lapeyre2019neural, becker2019deep}. The Stochastic Differential Equation (SDE) describing this model is  
$$d X_t = (r-\delta) X_t dt + \sigma X_t d W_t,$$ 
with $X_0 = x_0$, where $(W_t)_{t\geq 0}$ is a $d$-dimensional Brownian motion. 
If not stated differently, we choose the rate $r = 0 \%$, the dividend rate $\delta = 0\%$,  the volatility $\sigma = 20\%$ and the initial stock price $x_0 \in \{80, 100, 120\}$. 

To increase the complexity, we also compare the algorithms on the Heston model \citep{ heston1993closed},  which is also used in  \citep{lapeyre2019neural}. The SDE describing this model is 
\begin{equation}\label{equ:heston model}
\begin{split}
d X_t &= (r - \delta) X_t dt + \sqrt{v_t} X_t d W_t, \\
d v_t &= -\kappa (v_t - v_{\infty}) dt + \sigma \sqrt{v_t}d B_t, 
\end{split}
\end{equation}
 with $X_0 = x_0$ and $v_0 = \nu_0$, where  
$(W_t)_{t\geq 0}$ and $(B_t)_{t\geq 0}$ are two $d$-dimensional Brownian motions correlated with coefficient $\rho \in (-1,1)$. 
Here, $X$ is the stock price and $v$ the stochastic variance process.
If not stated differently, we choose the drift $r = 0 \%$,  the dividend rate $\delta = 0\%$,  the volatility of volatility $\sigma = 20\%$, the long term variance $v_\infty = 0.01$, the
 mean reversion speed $\kappa = 2$, the correlation $\rho = -30\%$, the initial stock price $x_0 = 100$ and the  initial variance $\nu_0 = 0.01$ (in particular, the Feller condition $2 \kappa v_{\infty} > \sigma^2$ is not satisfied, hence $v_t$ might touch the value $0$ but is reflected immediately).
Since the Heston model is Markovian only if the price and the variance $(X_t, v_t)$ are observed simultaneously, we give both values as  inputs to the algorithms here, and denote this below by ``Heston (with variance)''.

\subsubsection{Payoffs (max call, geometric put, basket call and min put)}

We test our algorithms on four different types of options.
First, we consider the max call option as it is a classical example used in optimal stopping \citep{lapeyre2019neural, becker2019deep}. The payoff of a max call option is defined by $g(x) = ( \max (x_1, x_2, \dots, x_d)-K)_{+}$ for any $x = (x_1, x_2, \dots, x_d)\in \mathbb{R}^d$. Moreover, we  consider the geometric put option, used in \citep{lapeyre2019neural}, with payoff
$g(x) =( K -  (\prod_{i=1}^d x_i)^{1/d} )_{+}$. We also test our approach on a basket call option  \citep{Hanbali2019Basket}, where the payoff is given by $g(x) = 
( \tfrac{1}{d}\sum_{i=1}^{d} x_i-K)_{+}$ and a min put option with payoff $g(x) = (K -  \min (x_1, x_2, \dots, x_d))_{+}$. 
For all these payoffs, the strike $K$ is set to $100$, unless stated differently.

\subsubsection{Reference prices}
In some cases reference prices can be computed as additional baselines.
All call options where the underlying stocks have a  rate $r \geq 0$ and dividend $\delta = 0$ are optimally executed at maturity. Therefore, the price of the American option and of the corresponding European option are the same under these constraints  \citep{FoellmerSchied2016}.
For all examples where this is the case, we compute the European option price (EOP) as an approximation of the correct American option price.

Moreover, as explained in \citep{lapeyre2019neural}, geometric put options on $d$-dimensional stocks following Black--Scholes are equivalent to one-dimensional put options on a $1$-dimen\-sional stock following Black--Scholes with adjusted parameters. The $1$-dimensional problem can be priced efficiently with the CRR binomial-tree method (B) \citep{cox1979option}. With the adjusted parameters $\hat \sigma = \frac{\sigma}{\sqrt{d}}$ and $\hat \delta = \delta + \frac{\sigma^2 - \hat \sigma ^2}{2} $ the binomial-tree model is defined with factors for the stock price going up and down $\gamma_{up} = \exp(\hat \sigma \sqrt{T/N})$, $\gamma_{down}=\frac{1}{\gamma_{up}}$, probabilities to go up and down $p = \frac{\exp((r - \hat \delta)  T/N) - \gamma_{down}}{\gamma_{up} - \gamma_{down}}$, $1-p$ and step-wise discounting factor $\exp(-r T/N)$.  The price computed with this method converges to the correct price under the Black--Scholes model as $N \to \infty$ \citep{cox1979option}. 
Hence, this method yields good approximations of the correct option price for large $N$. Whenever applied, we use $N=10'000$ for the binomial-tree method.
While in the first case of call options, the optimal stopping problem has an easy solution, i.e., to wait until maturity, this is not the case here, where the optimal stopping problem is more complex.

The remaining options, i.e., call options with $\delta > 0$, put options with $r > 0$ and geometric put option with underlying stocks following a Heston model, also constitute more complex stopping problems, where no efficient methods to compute the (approximately) correct price are available.
Therefore, we evaluate the performance of the algorithms by comparing the approximated prices directly. Since these prices are computed on unseen paths for all algorithms, where at each time, the algorithm can  only decide whether to exercise or not, higher prices imply better performance of the algorithms.

\subsubsection{Results and discussion}
All algorithms are run $10$ times in parallel and the mean and standard deviation (in parenthesis) of the prices respectively the median of the corresponding computation times  are reported.
In particular, the computation times do not include the time for generating the stock paths, since the main interest is in the actual time the algorithms need to compute prices and paths can be generated offline and stored.
In the following discussion, we always compare computation times for large $d$, since random machine influences have less impact there.

\begin{table*}[!h]
\center
\resizebox{\textwidth}{!}{
\begin{tabular}{|cc|r r r r r r r|r r r r r r r|}
\toprule
     & {} & \multicolumn{7}{c |}{price} & \multicolumn{7}{c |}{duration} \\
$d$ & $x_0$ & LSM & DOS & NLSM & RLSM & FQI & RFQI & EOP & LSM & DOS & NLSM & RLSM & FQI & RFQI & EOP\\
\midrule
\multirow{3}{*}{5} & 80  &   5.23 (0.07) &    5.12 (0.12) &    5.19 (0.09) &    5.28 (0.12) &   5.26 (0.10) &    5.20 (0.06) &    5.31 (0.05) &       11s &        9s &       0s &       0s &        2s &       0s &       0s \\
     & 100 &  24.95 (0.14) &   24.64 (0.21) &   24.72 (0.15) &   24.91 (0.16) &  24.96 (0.17) &   25.00 (0.19) &   24.97 (0.15) &       11s &        8s &       2s &       0s &        2s &       0s &       0s \\
     & 120 &  49.73 (0.21) &   49.45 (0.18) &   49.47 (0.22) &   49.62 (0.25) &  49.68 (0.22) &   49.75 (0.17) &   49.77 (0.15) &      11s &        7s &       2s &       0s &        2s &       0s &       0s \\
\cline{1-16}
\multirow{3}{*}{10} & 80  &   9.20 (0.07) &    9.19 (0.14) &    8.82 (0.15) &    9.24 (0.11) &   9.25 (0.12) &    9.25 (0.10) &    9.27 (0.09) &       28s &        7s &       1s &       0s &        6s &       0s &       0s \\
     & 100 &  34.33 (0.15) &   34.03 (0.17) &   33.69 (0.20) &   34.28 (0.11) &  34.25 (0.19) &   34.17 (0.11) &   34.26 (0.09) &      29s &        7s &       2s &       0s &        7s &       0s &       0s \\
     & 120 &  60.94 (0.24) &   60.90 (0.20) &   60.33 (0.25) &   61.08 (0.23) &  61.10 (0.19) &   61.07 (0.21) &   61.20 (0.13) &        29s &        7s &       2s &       0s &        6s &       0s &       0s \\
\cline{1-16}
\multirow{3}{*}{50} & 80  &  22.45 (0.11) &   23.17 (0.10) &   21.78 (0.34) &   22.03 (0.16) &  23.51 (0.13) &   23.42 (0.11) &   23.52 (0.09) &      8m39s &        8s &       2s &       0s &     6m28s &       1s &       0s \\
     & 100 &  53.49 (0.10) &   53.93 (0.12) &   52.15 (0.60) &   52.44 (0.21) &  54.24 (0.09) &   54.23 (0.08) &   54.37 (0.09) &      8m42s &        8s &       3s &       0s &     6m57s &       1s &       0s \\
     & 120 &  84.31 (0.12) &   84.72 (0.12) &   82.48 (0.79) &   82.98 (0.16) &  85.03 (0.18) &   85.00 (0.20) &   85.28 (0.07) &     8m46s &        9s &       3s &       0s &     7m 4s &       1s &       0s \\
\cline{1-16}
\multirow{3}{*}{100} & 80  &  24.02 (0.21) &   29.56 (0.13) &   27.08 (0.47) &   28.50 (0.06) &  29.59 (0.15) &   29.88 (0.08) &   29.95 (0.08) &   39m44s &       13s &       3s &       0s &  1h23m39s &       1s &       0s \\
     & 100 &  56.83 (0.18) &   61.84 (0.26) &   58.99 (0.62) &   60.58 (0.10) &  62.07 (0.15) &   62.32 (0.16) &   62.43 (0.08) &    40m42s &       13s &       4s &       0s &  1h23m28s &       1s &       0s \\
     & 120 &  88.05 (0.31) &   94.26 (0.16) &   90.48 (0.89) &   92.71 (0.08) &  94.41 (0.17) &   94.65 (0.14) &   94.99 (0.14) &   40m25s &       13s &       4s &       0s &  1h22m15s &       1s &       0s \\
\cline{1-16}
\multirow{3}{*}{500} & 80  &             - &   42.54 (0.16) &   39.45 (0.61) &   43.00 (0.07) &             - &   44.15 (0.09) &   44.34 (0.08) &         - &     53s &      11s &       1s &         - &       1s &       0s \\
     & 100 &             - &   78.27 (0.16) &   74.23 (1.03) &   78.80 (0.10) &             - &   80.21 (0.13) &   80.45 (0.08) &       - &       53s &      12s &       1s &         - &       1s &       0s \\
     & 120 &             - &  113.83 (0.18) &  108.60 (1.01) &  114.54 (0.09) &             - &  116.26 (0.19) &  116.48 (0.11) &      - &       53s &      12s &       1s &         - &       1s &       0s \\
\cline{1-16}
\multirow{3}{*}{1000} & 80  &             - &   47.91 (0.08) &   45.37 (0.91) &   49.13 (0.11) &             - &   50.14 (0.10) &   50.32 (0.07) &         - &     1m34s &      20s &       2s &         - &       1s &       0s \\
     & 100 &             - &   84.99 (0.19) &   81.06 (0.56) &   86.40 (0.08) &             - &   87.70 (0.09) &   87.93 (0.08) &       - &     1m35s &      20s &       3s &         - &       1s &       0s \\
     & 120 &             - &  121.98 (0.11) &  118.61 (1.31) &  123.68 (0.08) &             - &  125.19 (0.14) &  125.48 (0.09) &        - &     1m36s &      19s &       2s &         - &       1s &       0s \\
\cline{1-16}
\multirow{3}{*}{2000} & 80  &             - &   53.14 (0.13) &   51.45 (0.82) &   55.13 (0.09) &             - &   56.03 (0.04) &   56.27 (0.07) &         - &     2m57s &      34s &       5s &         - &       2s &       0s \\
     & 100 &             - &   91.43 (0.13) &   89.84 (0.67) &   93.87 (0.12) &             - &   95.00 (0.14) &   95.31 (0.06) &        - &     3m 2s &      39s &       5s &         - &       2s &       0s \\
     & 120 &             - &  129.77 (0.15) &  127.14 (1.30) &  132.69 (0.15) &             - &  134.09 (0.08) &  134.34 (0.10) &        - &     2m57s &      37s &       4s &         - &       2s &       0s \\
\bottomrule
\end{tabular}

}
\caption{Max call option on Black--Scholes for different number of stocks $d$ and varying initial stock price $x_0$.
}
\label{table_$x_0$s_Dim_BS_MaxCallr0_gt1}
\end{table*}

%\caption{Max call option on Black--Scholes for different number of stocks $d$ and varying initial stock price $x_0$. }

\begin{table*}[!ht]
\center
\resizebox{\textwidth}{!}{
\begin{tabular}{|c|r r r r r r r|r r r r r r r|}
\toprule
{} & \multicolumn{7}{c |}{price} & \multicolumn{7}{c |}{duration} \\
$d$ & LSM & DOS & NLSM & RLSM & FQI & RFQI & EOP & LSM & DOS & NLSM & RLSM & FQI & RFQI & EOP\\
\midrule
5         &   8.34 (0.08) &   8.36 (0.07) &   8.22 (0.09) &   8.37 (0.07) &   8.25 (0.03) &   8.33 (0.07) &   8.23 (0.04) &       31s &        6s &        3s &       0s &        8s &       0s &       0s \\
10        &  11.81 (0.06) &  11.83 (0.07) &  11.51 (0.12) &  11.83 (0.02) &  11.79 (0.06) &  11.83 (0.05) &  11.79 (0.07) &     1m30s &        6s &        3s &       0s &       28s &       0s &       0s \\
50        &  16.85 (0.07) &  20.01 (0.06) &  18.60 (0.32) &  19.31 (0.05) &  20.05 (0.06) &  20.09 (0.05) &  20.04 (0.04) &    39m37s &        8s &        4s &       0s &  1h22m45s &       1s &       0s \\
100       &             - &  23.49 (0.06) &  21.75 (0.41) &  22.90 (0.02) &             - &  23.69 (0.06) &  23.66 (0.04) &         - &       14s &        6s &       0s &         - &       1s &       0s \\
500       &             - &  31.31 (0.06) &  29.93 (0.32) &  31.35 (0.06) &             - &  32.14 (0.06) &  32.13 (0.07) &         - &     1m19s &       24s &       3s &         - &       2s &       0s \\
1000      &             - &  34.23 (0.08) &  33.79 (0.29) &  35.09 (0.06) &             - &  35.82 (0.06) &  35.86 (0.04) &         - &     2m59s &       41s &       6s &         - &       4s &       0s \\
2000      &             - &  35.18 (0.14) &  37.76 (0.23) &  38.84 (0.05) &             - &  39.63 (0.08) &  39.60 (0.05) &         - &    13m11s &     1m28s &      13s &         - &       7s &       0s \\
\bottomrule
\end{tabular}

}
\caption{Max call option on Heston (with variance) for different number of stocks $d$.
}
\label{table_Dim_HestonV_MaxCallr0_gt1}
\end{table*}

%\caption{Max call option on Heston (with variance) for different number of stocks $d$.}

\begin{table*}[!h]
\center
\resizebox{\textwidth}{!}{
\begin{tabular}{|c|r r r r r r r|r r r r r r r|}
\toprule
{} & \multicolumn{7}{c |}{price} & \multicolumn{7}{c |}{duration} \\
$d$ & LSM & DOS & NLSM & RLSM & FQI & RFQI & EOP & LSM & DOS & NLSM & RLSM & FQI & RFQI & EOP\\
\midrule
5         &  3.60 (0.05) &  3.57 (0.05) &  3.49 (0.06) &  3.58 (0.03) &  3.61 (0.03) &  3.62 (0.06) &  3.59 (0.02) &       13s &        6s &       2s &       0s &        2s &       0s &       0s \\
10        &  2.54 (0.04) &  2.52 (0.03) &  2.45 (0.06) &  2.54 (0.04) &  2.53 (0.03) &  2.53 (0.03) &  2.54 (0.01) &       30s &        6s &       1s &       0s &        7s &       0s &       0s \\
50        &  0.94 (0.01) &  1.12 (0.01) &  0.83 (0.03) &  1.06 (0.01) &  1.13 (0.01) &  1.15 (0.01) &  1.14 (0.01) &     8m51s &        8s &       1s &       0s &     7m 3s &       1s &       0s \\
100       &  0.51 (0.01) &  0.78 (0.01) &  0.55 (0.01) &  0.75 (0.01) &  0.80 (0.01) &  0.81 (0.01) &  0.81 (0.01) &    38m59s &       13s &       2s &       0s &  1h21m59s &       1s &       0s \\
500       &            - &  0.33 (0.01) &  0.24 (0.00) &  0.34 (0.00) &            - &  0.36 (0.00) &  0.36 (0.00) &         - &     1m 7s &       7s &       1s &         - &       1s &       0s \\
1000      &            - &  0.22 (0.00) &  0.17 (0.00) &  0.24 (0.00) &            - &  0.25 (0.00) &  0.26 (0.00) &         - &     2m24s &      14s &       2s &         - &       2s &       0s \\
2000      &            - &  0.13 (0.00) &  0.12 (0.01) &  0.17 (0.00) &            - &  0.18 (0.00) &  0.18 (0.00) &         - &     5m35s &      25s &       7s &         - &       3s &       0s \\
\bottomrule
\end{tabular}

}
\caption{Basket call  options on Black--Scholes for different number of stocks $d$.
}
\label{table_BasketCall_payoffsr0_gt1}
\end{table*}

%\caption{Basket call  options on Black--Scholes for different number of stocks $d$. }

\begin{table*}[!h]
\center
\resizebox{\textwidth}{!}{
\begin{tabular}{|cc|r r r r r r r|r r r r r r r|}
\toprule
       & {} & \multicolumn{7}{c |}{price} & \multicolumn{7}{c |}{duration} \\
model & $d$ & LSM & DOS & NLSM & RLSM & FQI & RFQI & B & LSM & DOS & NLSM & RLSM & FQI & RFQI & B\\
\midrule
\multirow{5}{*}{BlackScholes} & 5   &  3.34 (0.04) &  3.31 (0.03) &  3.29 (0.06) &  3.33 (0.04) &  3.31 (0.05) &  3.35 (0.04) &  3.35 (nan) &       11s &       6s &       1s &       0s &        2s &       0s &     3m12s \\
       & 10  &  2.37 (0.04) &  2.42 (0.02) &  2.33 (0.02) &  2.40 (0.04) &  2.39 (0.03) &  2.40 (0.03) &  2.40 (nan) &       28s &       6s &       1s &       0s &        7s &       0s &     3m12s \\
       & 20  &  1.65 (0.02) &  1.71 (0.04) &  1.57 (0.04) &  1.65 (0.03) &  1.73 (0.04) &  1.72 (0.02) &  1.71 (nan) &     1m31s &       6s &       1s &       0s &       32s &       1s &     3m12s \\
       & 50  &  0.91 (0.01) &  1.07 (0.02) &  0.80 (0.02) &  1.03 (0.01) &  1.09 (0.02) &  1.09 (0.02) &  1.09 (nan) &     8m26s &      10s &       2s &       0s &     7m24s &       1s &     3m31s \\
       & 100 &  0.50 (0.01) &  0.76 (0.01) &  0.54 (0.01) &  0.73 (0.01) &  0.77 (0.01) &  0.77 (0.01) &  0.78 (nan) &    38m37s &      16s &       2s &       0s &  1h23m50s &       1s &     3m31s \\
\cline{1-16}
\multirow{5}{*}{Heston} & 5   &  2.45 (0.03) &  2.44 (0.03) &  2.30 (0.06) &  2.44 (0.02) &  2.44 (0.04) &  2.43 (0.03) &           - &       11s &       6s &       1s &       0s &        2s &       0s &         - \\
       & 10  &  2.00 (0.02) &  2.00 (0.02) &  1.75 (0.04) &  2.00 (0.03) &  2.00 (0.02) &  2.01 (0.02) &           - &       29s &       6s &       2s &       0s &        7s &       0s &         - \\
       & 20  &  1.68 (0.02) &  1.69 (0.02) &  1.21 (0.05) &  1.62 (0.05) &  1.72 (0.02) &  1.71 (0.01) &           - &     1m31s &       7s &       2s &       0s &       32s &       1s &         - \\
       & 50  &  1.33 (0.02) &  1.47 (0.01) &  0.83 (0.03) &  1.24 (0.01) &  1.49 (0.01) &  1.48 (0.01) &           - &     8m31s &       7s &       3s &       0s &     7m13s &       1s &         - \\
       & 100 &  0.88 (0.01) &  1.39 (0.01) &  0.71 (0.02) &  1.18 (0.01) &  1.41 (0.01) &  1.40 (0.01) &           - &    41m34s &      15s &       4s &       0s &  1h24m11s &       1s &         - \\
\bottomrule
\end{tabular}

}
\caption{Geometric put options on Black--Scholes and Heston (with variance) for different number of stocks $d$. Here $r = 2\%$ is used as interest rate.
}
\label{table_GeoPut_payoffs_gt1}
\end{table*}

%\caption{Geometric put options on Black--Scholes and Heston (with variance) for different number of stocks $d$. Here $r = 2\%$ is used as interest rate.}

\begin{table*}[!h]
\center
\resizebox{\textwidth}{!}{
\begin{tabular}{|cc|r r r r r r|r r r r r r|}
\toprule
     & {} & \multicolumn{6}{c |}{price} & \multicolumn{6}{c |}{duration} \\
$d$ & $x_0$ & LSM & DOS & NLSM & RLSM & FQI & RFQI & LSM & DOS & NLSM & RLSM & FQI & RFQI\\
\midrule
\multirow{3}{*}{5} & 80  &  35.49 (0.07) &  35.48 (0.06) &  35.21 (0.12) &  35.46 (0.07) &  35.53 (0.08) &  35.54 (0.05) &       11s &       10s &       3s &       0s &        3s &       0s \\
     & 100 &  19.98 (0.09) &  19.96 (0.09) &  19.68 (0.07) &  19.96 (0.14) &  19.97 (0.10) &  19.95 (0.09) &       11s &        9s &       3s &       0s &        3s &       0s \\
     & 120 &   7.46 (0.10) &   7.36 (0.08) &   7.25 (0.07) &   7.39 (0.10) &   7.45 (0.11) &   7.38 (0.10) &       11s &        6s &       1s &       0s &        2s &       0s \\
\cline{1-14}
\multirow{3}{*}{10} & 80  &  40.22 (0.05) &  40.17 (0.05) &  39.91 (0.10) &  40.21 (0.07) &  40.31 (0.07) &  40.30 (0.04) &       28s &        6s &       2s &       0s &        9s &       0s \\
     & 100 &  25.74 (0.09) &  25.74 (0.10) &  25.36 (0.12) &  25.76 (0.09) &  25.79 (0.10) &  25.83 (0.13) &       28s &        6s &       3s &       0s &        6s &       0s \\
     & 120 &  11.98 (0.07) &  11.92 (0.09) &  11.62 (0.14) &  11.94 (0.13) &  11.96 (0.10) &  12.03 (0.07) &       28s &        5s &       1s &       0s &        6s &       0s \\
\cline{1-14}
\multirow{3}{*}{50} & 80  &  48.08 (0.05) &  48.27 (0.04) &  47.03 (0.19) &  47.72 (0.03) &  48.36 (0.05) &  48.34 (0.04) &     8m25s &        8s &       3s &       0s &     5m20s &       1s \\
     & 100 &  35.57 (0.07) &  35.80 (0.08) &  34.27 (0.40) &  35.11 (0.04) &  35.91 (0.07) &  35.87 (0.08) &     8m35s &        8s &       3s &       0s &     6m57s &       1s \\
     & 120 &  22.93 (0.08) &  23.33 (0.07) &  21.40 (0.41) &  22.50 (0.05) &  23.41 (0.06) &  23.42 (0.10) &     8m28s &        8s &       2s &       0s &     6m52s &       1s \\
\cline{1-14}
\multirow{3}{*}{100} & 80  &  49.71 (0.06) &  50.93 (0.04) &  48.78 (0.26) &  50.48 (0.04) &  50.93 (0.04) &  50.99 (0.03) &    39m57s &       13s &       3s &       0s &  1h22m58s &       1s \\
     & 100 &  37.63 (0.07) &  39.11 (0.05) &  36.42 (0.80) &  38.55 (0.05) &  39.11 (0.06) &  39.22 (0.04) &    40m12s &       12s &       3s &       0s &  1h23m26s &       1s \\
     & 120 &  25.52 (0.08) &  27.31 (0.05) &  24.13 (0.52) &  26.64 (0.03) &  27.28 (0.07) &  27.42 (0.05) &    40m40s &       12s &       3s &       0s &  1h22m53s &       1s \\
\cline{1-14}
\multirow{3}{*}{500} & 80  &             - &  55.71 (0.03) &  51.51 (0.46) &  55.66 (0.03) &             - &  56.04 (0.03) &         - &       54s &      13s &       1s &         - &       1s \\
     & 100 &             - &  45.14 (0.05) &  40.06 (1.04) &  45.05 (0.02) &             - &  45.51 (0.03) &         - &       53s &      13s &       1s &         - &       1s \\
     & 120 &             - &  34.53 (0.05) &  28.39 (0.75) &  34.45 (0.05) &             - &  34.99 (0.02) &         - &       54s &      12s &       1s &         - &       1s \\
\cline{1-14}
\multirow{3}{*}{1000} & 80  &             - &  57.40 (0.03) &  53.50 (0.64) &  57.52 (0.03) &             - &  57.84 (0.02) &         - &     1m36s &      21s &       3s &         - &       2s \\
     & 100 &             - &  47.24 (0.05) &  42.35 (0.63) &  47.40 (0.05) &             - &  47.76 (0.03) &         - &     1m37s &      21s &       3s &         - &       2s \\
     & 120 &             - &  37.04 (0.03) &  31.10 (0.87) &  37.25 (0.04) &             - &  37.68 (0.03) &         - &     1m34s &      20s &       3s &         - &       2s \\
\cline{1-14}
\multirow{3}{*}{2000} & 80  &             - &  58.59 (0.04) &  55.21 (0.67) &  59.21 (0.02) &             - &  59.50 (0.02) &         - &     3m 1s &      30s &       6s &         - &       3s \\
     & 100 &             - &  48.72 (0.04) &  44.37 (0.61) &  49.49 (0.04) &             - &  49.83 (0.03) &         - &     2m56s &      31s &       6s &         - &       3s \\
     & 120 &             - &  38.84 (0.06) &  33.31 (0.73) &  39.79 (0.04) &             - &  40.18 (0.04) &         - &     3m 0s &      31s &       6s &         - &       3s \\
\bottomrule
\end{tabular}

}
\caption{Min put option on Black--Scholes for different number of stocks $d$ and varying initial stock price $x_0$. Here $r=2\%$ is used as interest rate.
}
\label{table_$x_0$s_Dim_BS_MinPut_gt1}
\end{table*}

%\caption{Min put option on Black--Scholes for different number of stocks $d$ and varying initial stock price $x_0$. Here $r=2\%$ is used as interest rate. }

\begin{table*}[!h]
\center
\resizebox{\textwidth}{!}{
\begin{tabular}{|c|r r r r r r|r r r r r r|}
\toprule
{} & \multicolumn{6}{c |}{price} & \multicolumn{6}{c |}{duration} \\
$d$ & LSM & DOS & NLSM & RLSM & FQI & RFQI & LSM & DOS & NLSM & RLSM & FQI & RFQI\\
\midrule
5         &  18.83 (0.17) &  18.66 (0.11) &  18.62 (0.18) &  18.83 (0.12) &  18.43 (0.10) &  18.83 (0.16) &       22s &        7s &       3s &       0s &        2s &       0s \\
10        &  26.67 (0.14) &  26.72 (0.17) &  26.35 (0.14) &  26.60 (0.12) &  26.56 (0.13) &  26.77 (0.09) &       46s &        7s &       3s &       0s &        8s &       0s \\
50        &  43.86 (0.10) &  44.52 (0.13) &  43.27 (0.33) &  43.37 (0.11) &  44.66 (0.14) &  44.78 (0.12) &    10m 5s &       10s &       4s &       0s &     7m23s &       1s \\
100       &  46.62 (0.19) &  51.71 (0.09) &  49.31 (0.56) &  50.61 (0.10) &  51.79 (0.17) &  52.16 (0.09) &    49m27s &       15s &       5s &       0s &  1h21m26s &       1s \\
500       &             - &  67.12 (0.09) &  62.82 (0.72) &  67.02 (0.08) &             - &  68.48 (0.13) &         - &       59s &      14s &       2s &         - &       2s \\
1000      &             - &  73.37 (0.12) &  69.25 (0.83) &  73.85 (0.10) &             - &  75.31 (0.08) &         - &     1m52s &      26s &       4s &         - &       2s \\
2000      &             - &  78.17 (0.11) &  76.57 (0.83) &  80.54 (0.07) &             - &  81.96 (0.16) &         - &     5m26s &      47s &       8s &         - &       3s \\
\bottomrule
\end{tabular}

}
\caption{Max call  option on Black--Scholes for different number of stocks $d$. Here $r=5\%$ is used as interest rate and $\delta = 10\%$ as dividend rate.
}
\label{table_Dim_BS_MaxCall_div_gt1}
\end{table*}

%\caption{Max call  option on Black--Scholes for different number of stocks $d$. Here $r=5\%$ is used as interest rate and $\delta = 10\%$ as dividend rate. }

\begin{table*}[!h]
\center
\resizebox{\textwidth}{!}{
\begin{tabular}{|c|r r r r r r|r r r r r r|}
\toprule
{} & \multicolumn{6}{c |}{price} & \multicolumn{6}{c |}{duration} \\
$d$ & LSM & DOS & NLSM & RLSM & FQI & RFQI & LSM & DOS & NLSM & RLSM & FQI & RFQI\\
\midrule
5         &  12.34 (0.05) &  12.31 (0.05) &  12.16 (0.11) &  12.29 (0.06) &  12.35 (0.09) &  12.37 (0.09) &       30s &        6s &        3s &       0s &        8s &       0s \\
10        &  16.48 (0.07) &  16.52 (0.08) &  16.09 (0.13) &  16.55 (0.06) &  16.64 (0.07) &  16.61 (0.08) &     1m31s &        6s &        3s &       0s &       28s &       0s \\
50        &  22.86 (0.05) &  25.56 (0.04) &  24.03 (0.42) &  24.85 (0.08) &  25.72 (0.03) &  25.71 (0.07) &    39m57s &        9s &        4s &       0s &  1h21m59s &       1s \\
100       &             - &  29.13 (0.04) &  27.30 (0.46) &  28.50 (0.06) &             - &  29.33 (0.07) &         - &       16s &        6s &       0s &         - &       1s \\
500       &             - &  36.26 (0.05) &  34.74 (0.31) &  36.28 (0.04) &             - &  36.95 (0.05) &         - &     1m21s &       24s &       3s &         - &       2s \\
1000      &             - &  38.62 (0.08) &  38.19 (0.20) &  39.32 (0.03) &             - &  39.93 (0.05) &         - &     3m18s &       45s &       6s &         - &       4s \\
2000      &             - &  39.22 (0.13) &  41.05 (0.21) &  42.22 (0.04) &             - &  42.81 (0.04) &         - &    12m51s &     1m37s &      13s &         - &       8s \\
\bottomrule
\end{tabular}

}
\caption{Min put option on Heston (with variance) for different number of stocks $d$. Here $r=2\%$ is used as interest rate.
}
\label{table_$x_0$s_Dim_HestonV_MinPut_gt1}
\end{table*}

%\caption{Min put option on Heston (with variance) for different number of stocks $d$ and varying initial stock price $x_0$. Here $r=2\%$ is used as interest rate.}

\begin{table*}[!h]
\center
\resizebox{\textwidth}{!}{
\begin{tabular}{|c|r r r r r r|r r r r r r|}
\toprule
{} & \multicolumn{6}{c |}{price} & \multicolumn{6}{c |}{duration} \\
$d$ & LSM & DOS & NLSM & RLSM & FQI & RFQI & LSM & DOS & NLSM & RLSM & FQI & RFQI\\
\midrule
5         &   4.88 (0.03) &   4.89 (0.03) &   4.69 (0.05) &   4.83 (0.04) &   4.37 (0.06) &   4.59 (0.08) &       31s &        5s &        3s &       0s &        8s &       0s \\
10        &   7.19 (0.06) &   7.20 (0.04) &   6.90 (0.07) &   7.17 (0.04) &   6.63 (0.07) &   6.84 (0.06) &     1m33s &        5s &        2s &       0s &       27s &       0s \\
50        &  11.68 (0.05) &  13.99 (0.07) &  12.93 (0.28) &  13.70 (0.05) &  13.72 (0.09) &  13.71 (0.04) &    41m 6s &        8s &        3s &       0s &  1h22m14s &       1s \\
100       &             - &  17.04 (0.07) &  15.94 (0.29) &  16.80 (0.03) &             - &  16.97 (0.05) &         - &       11s &        5s &       0s &         - &       1s \\
500       &             - &  24.05 (0.05) &  22.95 (0.40) &  24.35 (0.05) &             - &  24.70 (0.05) &         - &     1m19s &       23s &       3s &         - &       2s \\
1000      &             - &  26.86 (0.05) &  26.47 (0.39) &  27.71 (0.04) &             - &  28.08 (0.05) &         - &     2m48s &       41s &       6s &         - &       4s \\
2000      &             - &  28.01 (0.11) &  30.12 (0.18) &  31.13 (0.05) &             - &  31.55 (0.07) &         - &    12m56s &     1m30s &      14s &         - &       7s \\
\bottomrule
\end{tabular}

}
\caption{Max call  option on Heston (with variance) for different number of stocks $d$. Here $r=5\%$ is used as interest rate and $\delta = 10\%$ as dividend rate.
}
\label{table_Dim_HestonV_MaxCall_div_gt1}
\end{table*}

%\caption{Max call  option on Heston (with variance) for different number of stocks $d$. Here $r=5\%$ is used as interest rate and $\delta = 10\%$ as dividend rate.}

\begin{table*}[!h]
\center
\resizebox{\textwidth}{!}{
\begin{tabular}{|cc|r r r r r r r|r r r r r r r|}
\toprule
    & {} & \multicolumn{7}{c |}{price} & \multicolumn{7}{c |}{duration} \\
$d$ & $N$ & LSM & DOS & NLSM & RLSM & FQI & RFQI & EOP & LSM & DOS & NLSM & RLSM & FQI & RFQI & EOP\\
\midrule
\multirow{3}{*}{10} & 10  &  34.33 (0.15) &  34.03 (0.17) &  33.69 (0.20) &  34.28 (0.11) &  34.25 (0.19) &  34.17 (0.11) &  34.26 (0.09) &      29s &        7s &        2s &       0s &        7s &       0s &       0s \\
    & 50  &  34.13 (0.12) &  34.14 (0.20) &  33.96 (0.11) &  33.98 (0.08) &  34.25 (0.21) &  34.15 (0.13) &  34.23 (0.11) &     2m44s &       32s &       20s &       0s &       46s &       6s &       0s \\
    & 100 &  34.15 (0.14) &  34.14 (0.26) &  33.98 (0.24) &  34.05 (0.15) &  34.29 (0.16) &  33.97 (0.11) &  34.28 (0.10) &     5m18s &     1m 6s &       32s &       1s &     1m27s &       7s &       0s \\
\cline{1-16}
\multirow{3}{*}{50} & 10  &  53.49 (0.10) &  53.93 (0.12) &  52.15 (0.60) &  52.44 (0.21) &  54.24 (0.09) &  54.23 (0.08) &  54.37 (0.09) &     8m42s &        8s &        3s &       0s &     6m57s &       1s &       0s \\
    & 50  &  52.82 (0.13) &  53.94 (0.18) &  53.24 (0.26) &  50.85 (0.18) &  54.31 (0.14) &  53.74 (0.08) &  54.46 (0.11) &    48m36s &       41s &       18s &       1s &    21m15s &       7s &       0s \\
    & 100 &  52.74 (0.11) &  54.09 (0.15) &  53.61 (0.18) &  50.42 (0.17) &  54.15 (0.10) &  53.77 (0.12) &  54.33 (0.14) &  1h37m48s &     1m36s &       37s &       2s &    41m 8s &      16s &       0s \\
\cline{1-16}
\multirow{3}{*}{100} & 10  &  56.83 (0.18) &  61.84 (0.26) &  58.99 (0.62) &  60.58 (0.10) &  62.07 (0.15) &  62.32 (0.16) &  62.43 (0.08) &    40m42s &       13s &        4s &       0s &  1h23m28s &       1s &       0s \\
    & 50  &             - &  61.88 (0.06) &  60.72 (0.24) &  58.68 (0.13) &             - &  61.66 (0.14) &  62.48 (0.07) &         - &     1m15s &       22s &       1s &         - &       8s &       0s \\
    & 100 &             - &  62.07 (0.11) &  61.19 (0.15) &  58.26 (0.16) &             - &  61.79 (0.11) &  62.46 (0.04) &         - &     2m23s &       44s &       3s &         - &      15s &       0s \\
\cline{1-16}
\multirow{3}{*}{500} & 10  &             - &  78.27 (0.16) &  74.23 (1.03) &  78.80 (0.10) &             - &  80.21 (0.13) &  80.45 (0.08) &        - &       53s &       12s &       1s &         - &       1s &       0s \\
    & 50  &             - &  79.14 (0.08) &  75.63 (1.07) &  76.68 (0.05) &             - &  79.23 (0.07) &  80.44 (0.09) &         - &     4m59s &     1m 4s &       8s &         - &       9s &       0s \\
    & 100 &             - &  79.44 (0.09) &  76.46 (0.41) &  76.33 (0.05) &             - &  79.34 (0.08) &  80.47 (0.10) &         - &    10m12s &     2m13s &      18s &         - &      19s &       0s \\
\bottomrule
\end{tabular}

}
\caption{Max call option on Black--Scholes for different number of stocks $d$ and higher number of exercise dates $N$.
}
\label{table_manyDates_BS_MaxCallr0_gt1}
\end{table*}

%\caption{Max call option on Black--Scholes for different number of stocks $d$ and higher number of exercise dates $N$.
%}

\begin{table*}[!h]
\center
\resizebox{\textwidth}{!}{
\begin{tabular}{|cc|r r r r r r|r r r r r r|}
\toprule
    & {} & \multicolumn{6}{c |}{price} & \multicolumn{6}{c |}{duration} \\
$d$ & $N$ & LSM & DOS & NLSM & RLSM & FQI & RFQI & LSM & DOS & NLSM & RLSM & FQI\footnotemark & RFQI\\
\midrule
\multirow{3}{*}{10} & 10  &  26.67 (0.14) &  26.72 (0.17) &  26.35 (0.14) &  26.60 (0.12) &  26.56 (0.13) &  26.77 (0.09) &       46s &        7s &        3s &       0s &        8s &       0s \\
    & 50  &  26.65 (0.12) &  26.56 (0.20) &  26.42 (0.17) &  26.61 (0.13) &  26.51 (0.07) &  26.69 (0.18) &     2m21s &       44s &       53s &       2s &       44s &       4s \\
    & 100 &  26.68 (0.19) &  26.44 (0.14) &  26.42 (0.19) &  26.59 (0.13) &  26.55 (0.14) &  26.65 (0.16) &     4m46s &     2m46s &     1m45s &       1s &     1m25s &       8s \\
\cline{1-14}
\multirow{3}{*}{50} & 10  &  43.86 (0.10) &  44.52 (0.13) &  43.27 (0.33) &  43.37 (0.11) &  44.66 (0.14) &  44.78 (0.12) &    10m 5s &       10s &        4s &       0s &     7m23s &       1s \\
    & 50  &  43.42 (0.09) &  44.50 (0.08) &  44.13 (0.19) &  42.26 (0.10) &  44.68 (0.15) &  44.72 (0.12) &    43m 5s &     1m14s &       52s &       3s &    16m46s &       9s \\
    & 100 &  43.21 (0.14) &  44.45 (0.15) &  44.30 (0.13) &  41.89 (0.31) &  44.64 (0.17) &  44.60 (0.14) &  1h25m 4s &     2m 0s &     1m45s &       2s &    36m10s &      17s \\
\cline{1-14}
\multirow{3}{*}{100} & 10  &  46.62 (0.19) &  51.71 (0.09) &  49.31 (0.56) &  50.61 (0.10) &  51.79 (0.17) &  52.16 (0.09) &    49m27s &       15s &        5s &       0s &  1h21m26s &       1s \\
    & 50  &             - &  51.73 (0.10) &  50.89 (0.21) &  49.36 (0.09) &             - &  51.91 (0.12) &         - &       52s &       33s &       1s &         - &       8s \\
    & 100 &             - &  51.72 (0.15) &  51.27 (0.12) &  48.90 (0.08) &             - &  51.84 (0.11) &         - &     1m48s &       46s &       3s &         - &      19s \\
\cline{1-14}
\multirow{3}{*}{500} & 10  &             - &  67.12 (0.09) &  62.82 (0.72) &  67.02 (0.08) &             - &  68.48 (0.13) &         - &       59s &       14s &       2s &         - &       2s \\
    & 50  &             - &  67.48 (0.11) &  64.75 (0.50) &  65.70 (0.07) &             - &  68.03 (0.12) &         - &     2m56s &     1m 4s &       7s &         - &      10s \\
    & 100 &             - &  67.50 (0.15) &  65.55 (0.34) &  65.22 (0.07) &             - &  67.91 (0.10) &         - &     5m48s &     1m52s &      16s &         - &      20s \\
\bottomrule
\end{tabular}

}
\caption{
Max call option on Black--Scholes for different number of stocks $d$ and higher number of exercise dates $N$.  Here $r=5\%$ is used as interest rate and $\delta = 10\%$ as dividend rate.
}
\label{table_manyDates_BS_MaxCall_div_gt1}
\end{table*}

%\caption{Max call option on Black--Scholes for different number of stocks $d$ and higher number of exercise dates $N$.  Here $r=5\%$ is used as interest rate and $\delta = 10\%$ as dividend rate.
%}

\begin{figure}[h!]
\center
%\begin{figure}[htp!]% [H] is so declass\'e!
\includegraphics[width=0.8\textwidth]{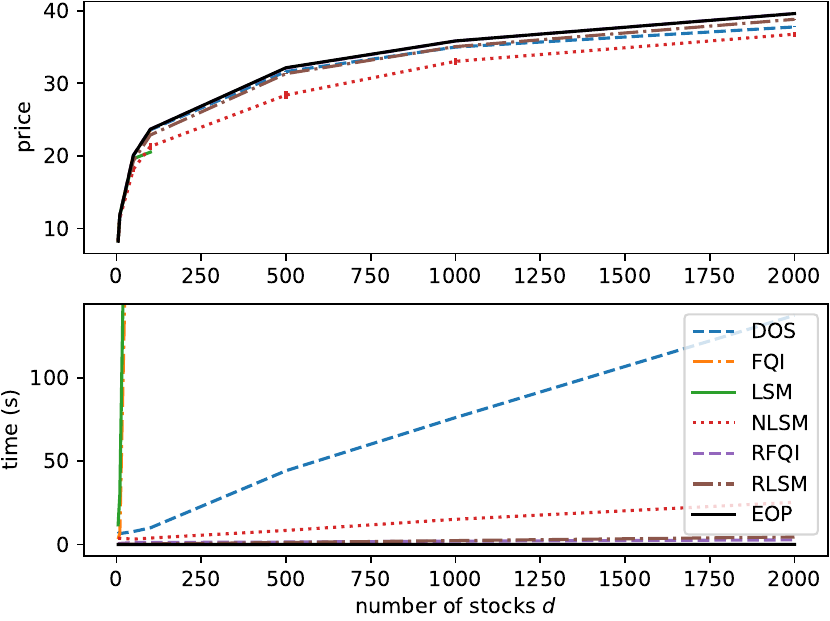}
\caption{At-the-money max call option without dividend on Heston.
}
\label{fig:comparison1 BS}
%\end{wrapfigure}
\end{figure}

\begin{figure}[h!]
\center
%\begin{figure}[htp!]% [H] is so declass\'e!
\includegraphics[width=0.8\textwidth]{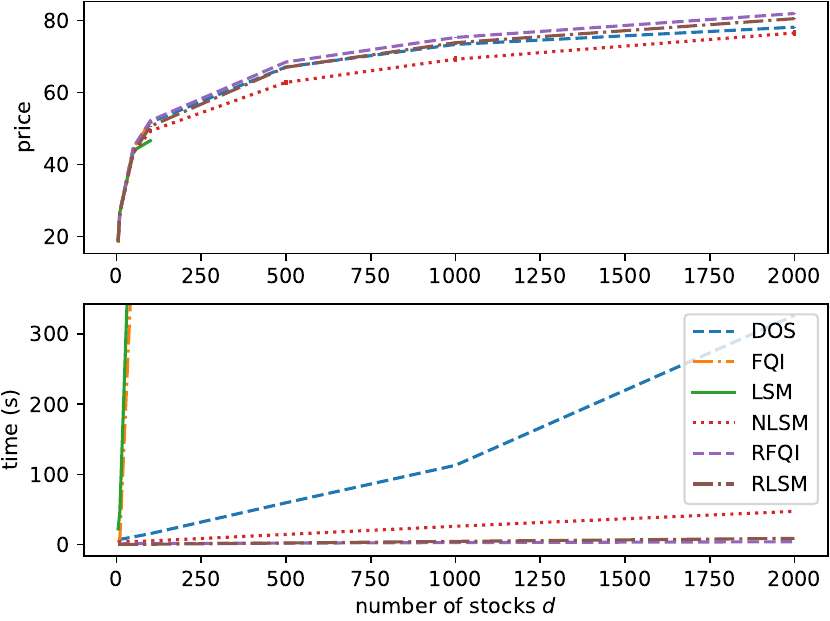}
\caption{At-the-money max call option with dividend on Black--Scholes.
}
\label{fig:comparison2 BS}
%\end{wrapfigure}
\end{figure}

In all cases RLSM  and RFQI  are the fastest algorithms while achieving at least similar prices to the best performing baselines. 
Their biggest strength are high dimensional problems ($d \geq 500$), where this speed-up becomes substantial.

In these high dimensional problems,  RLSM  performs as good as or outperforms the baselines in terms of prices, even tough RLSM  has much less trainable parameters than DOS and NLSM. Moreover, RFQI  achieves the highest prices there, and therefore works best, while having considerably less trainable parameters, since only one neural network (with a random hidden layer) of the respective size is used for all exercise dates. 
In particular, RFQI  has only $21$ trainable parameters, compared to more than $20 d N$ for DOS and NLSM.

Comparing the achieved prices of LSM and FQI, we can confirm the claim of \cite{Li2009Learning}, that reinforcement learning techniques usually outperform the backward induction in the Markovian setting.
RFQI, achieving similar prices as FQI, therefore naturally outperforms RLSM  which achieves similar prices as LSM. A possible explanation for the outperformance of the reinforcement learning algorithm is the following.
The backward induction algorithms have approximately $N$ times the number of trainable parameters used in the reinforcement learning algorithms, since a different network is trained for each discretisation date. Moreover, for the backward induction algorithms, a different continuation value function is approximated for each date, hence, only the data of this date is used to learn the parameters. In contrast, the reinforcement learning methods train their parameters using the data of all dates. Hence, the reinforcement learning methods use $N$ times the number of data to train $1/N$ times the number of parameters, which seems to lead to better approximations.

\footnotetext{Due to memory overflow issues, FQI could only be run with 5 instead of 10 parallel runs for larger $N$, hence the computation times are smaller then they would otherwise be, due to more CPU power per run.}

We first give a detailed discussion of results for the easy optimal stopping problems, where it is optimal to exercise the option at maturity. 
Although these optimal stopping problems are less complex, they are still interesting, because a minimal requirement for the algorithms should be that they perform well in these basic examples. Moreover, a comparison to the reference price is possible.
In Table~\ref{table_$x_0$s_Dim_BS_MaxCallr0_gt1}  we show results of a max call option on Black--Scholes. For high dimensions ($d\ge 500$), RLSM  is about $8$ times faster than the fastest baseline NLSM and about $30$ times faster than DOS. Moreover, RFQI  is  about twice as fast as  RLSM. For $d=100$ we also see the large difference in computation time between LSM  (respectively FQI), where the number of basis functions grows quadratically in $d$, and RLSM  (respectively RFQI), where the number of basis functions does not grow in $d$. 
The computed prices of RLSM   are at most $2\%$  smaller than those of LSM and the prices of RFQI  are at most  $1.2\%$ smaller than those of FQI.
For $d\leq 100$ the maximal relative errors compared to the reference prices are $19.7\%$ for LSM, $3.5\%$ for DOS, $9.5\%$ for NLSM, $6.4\%$ for RLSM, $1.2\%$ for FQI and $2.2\%$ for RFQI.
For $d\geq 500$ these errors are $5.6\%$  for DOS, $11\%$ for NLSM, $3\%$ for RLSM  and $0.4\%$ for RFQI.
The results of Table~\ref{table_Dim_HestonV_MaxCallr0_gt1} (max call on Heston with variance) and Table~\ref{table_BasketCall_payoffsr0_gt1} (basket call on Black--Scholes) are similar, except that relative errors become larger in Table \ref{table_BasketCall_payoffsr0_gt1} for growing $d$, since the prices become very small.
In Figure~\ref{fig:comparison1 BS} we plot the price and computation time for at-the-money max call options without dividend on the Heston model when increasing the number of stocks. It is well visible that the computation time of RLSM  and RFQI   hardly increases, while the prices are similar to the other algorithms.

In the remaining examples, it is in general not optimal to exercise the options at maturity, making the stopping decisions harder and therefore more challenging for the algorithms. 

For the geometric put options (Table \ref{table_GeoPut_payoffs_gt1}), we do not present dimensions larger than $100$, because prices cannot be computed numerically any more.
In the Black--Scholes case, the maximal relative errors compared to the reference price are $35.7\%$ for LSM, $2.1\%$ for DOS, $29.9\%$ for NLSM, $6\%$ for RLSM, $1.1\%$ for FQI and  $0.5\%$ for RFQI. 
Again, the prices computed with RLSM  (respectively RFQI)
 are never much smaller than those of LSM (FQI); $0.3\%$ ($0.1\%$) for Black--Scholes and $6\%$ ($0.6\%$)  for Heston (with variance).
 On the Heston model, RFQI, FQI and DOS achieve  the highest prices that never deviate more than $1.5\%$ from each other.

For the min put option on Black--Scholes (Table~\ref{table_$x_0$s_Dim_BS_MinPut_gt1}) RLSM  is about $7$ times faster than NLSM and more than $30$ times faster than DOS  for high dimensions. Furthermore, RFQI  is again about twice as fast as RLSM.
For $d \leq 50$ all algorithms yield very similar prices and for larger $d$ the highest prices are always achieved by RFQI, whereby the prices computed with RFQI  never deviate more than $1\%$ from those computed with FQI. Moreover, the prices computed with RLSM  are never more than $1.9\%$ smaller than those computed with LSM.
In addition, RLSM  achieves the second highest prices for high dimensions.
For the max call option with dividends on Black--Scholes (Table~\ref{table_Dim_BS_MaxCall_div_gt1} and Figure~\ref{fig:comparison2 BS}), the situation is similar. However, the highest prices are always achieved by RFQI  and the prices computed with RLSM  are at most $1.1\%$ smaller than those of LSM.
For the min put option on Heston (with variance) (Table~\ref{table_$x_0$s_Dim_HestonV_MinPut_gt1}) we have similar results as on Black--Scholes, but the prices computed with RLSM  (RFQI) are at most $0.5\%$ ($0.2\%$) smaller than those computed with LSM (FQI).

For the max call option with dividend on Heston (with variance) (Table~\ref{table_Dim_HestonV_MaxCall_div_gt1}), RLSM  is about $7$ times faster than NLSM and more than $26$ times faster then DOS  for high dimensions.
 RFQI  is again about twice as fast as RLSM.
 For $d \in \{ 5, 10 \}$ DOS yields the highest prices, RLSM  deviates at most  $1.2\%$ from them and RFQI  at most $6\%$. FQI yields lower prices than RFQI.
 For $d \in \{  50, 100\}$, DOS, RLSM and RFQI yield very similar prices deviating at most $2\%$ from each other.
 For higher dimensions of $d \geq 500$,  RFQI  yields the highest  and RLSM  the second highest prices.

When increasing the number of exercise dates for the max call option on Black--Scholes from $N=10$ to $N \in \{50, 100\}$ (Table~\ref{table_manyDates_BS_MaxCallr0_gt1}) the Bermudan option price should become closer to the American option price. 
The highest prices are achieved either by RFQI, FQI or DOS, with a maximum deviation of less than $1.4\%$ between their results and a maximum deviation from the reference prices of $2.7\%$ for DOS and $1.5\%$ for RFQI. RFQI  is more than $30$ times faster than DOS for high dimensions. Increasing the number of dates further, the computation time can become a limiting factor for DOS, while this is not the case for RFQI. 
We see similar results for the more complex max call option on Black--Scholes with dividends (Table~\ref{table_manyDates_BS_MaxCall_div_gt1}), where RFQI  always achieves the highest price.

\subsubsection{Empirical convergence study}
We confirm the theoretical results of Theorem \ref{thm:informal} (Figure \ref{fig:convergenceStudy} left) and Theorem \ref{thm2:informal} (Figure \ref{fig:convergenceStudy} right) by an empirical convergence study for a growing number of paths $m$. For RLSM  we also increase the number of hidden nodes $K$, while they are fixed for RFQI  since $d=5$ is used. 
For each combination of the number of paths $m$ and the hidden size $K$, the algorithms are run $20$ times and their mean prices with  standard deviations are shown. 
For small $m$, we see that smaller hidden sizes achieve better prices. This is due to overfitting to the training paths when using larger networks. Regularization techniques  like $L^1$- or $L^2$-penalization could be used to reduce overfitting for larger networks. However, our results suggest that restricting the hidden size is actually the simplest and best regularization technique, since it additionally leads to lower training times.
  \begin{figure}[!h]
        \centering
        \begin{subfigure}[b]{0.45\textwidth}
            \centering
            \includegraphics[width=\textwidth]{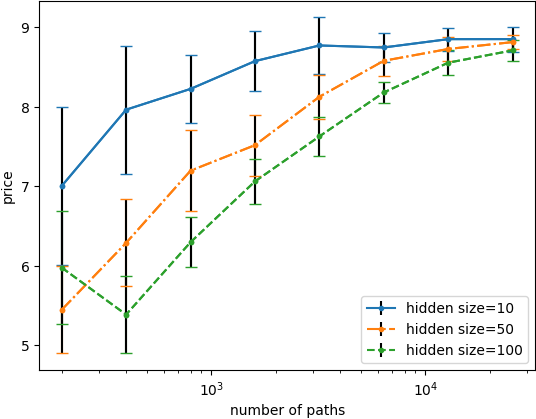}
%            \caption[Network2]%
%            {{\small Network 1}}    
            \label{fig:mean and std of net14}
        \end{subfigure}
        \hfill
        \vspace{-0.8cm}
        \begin{subfigure}[b]{0.475\textwidth}  
            \centering 
            \includegraphics[width=\textwidth]{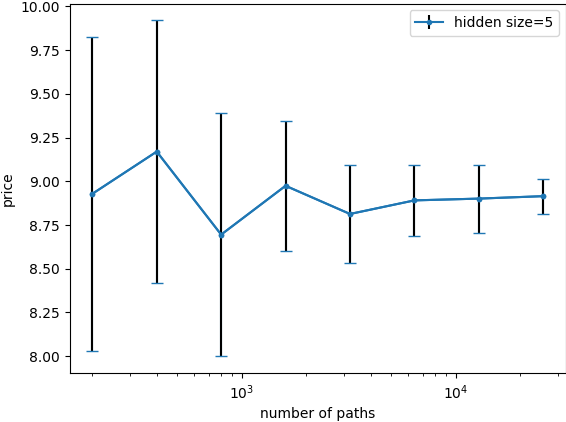}
%            \caption[]%
%            {{\small Network 2}}    
            \label{fig:mean and std of net24}
        \end{subfigure}
        \caption{ Mean $\pm$ standard deviation (bars) of the price for a max call on 5 stocks following the Black--Scholes model for RLSM  (left) and RFQI  (right) for varying the number of paths $m$ and varying for RLSM  the number of neurons in the hidden layer $K$.}
        \label{fig:convergenceStudy}
    \end{figure}

\subsection{The non-Markovian case -- optimally stopping fractional Brownian motions}\label{sec:The Non-Markovian Case -- Optimally Stopping Fractional Brownian Motions}

In order to compare our algorithms on a problem where the underlying process is non-Markovian, we take the example of the fractional Brownian motion $(W_t^H)_{t\geq 0}$ as in \citep{becker2019deep}. 
Unlike classical Brownian motion, the increments of fractional Brownian motion need not be independent.
Fractional Brownian motion is a continuous centered Gaussian process with covariation function
$E\left(W_t^{H} \ W_s^{H}\right) = \frac{1}{2}\left(|t|^{2H}+|s|^{2H} - |t-s|^{2H} \right)$
where $H \in (0,1]$ is called the Hurst parameter. 
When the Hurst parameter $H=0.5$, then $W^H$ is a standard Brownian motion; when $H\neq0.5$, the increments of $(W_t^H)_{t\geq 0}$  are correlated
 (positively if $H> 0.5$ and negatively if $H<0.5$) 
which means that for $H\neq0.5$, $(W_t^H)_{t\geq 0}$ is not Markovian \citep{bayer2016pricing, livieri2018rough, Gatheral2018,el2018microstructural, abi2019multifactor}.

\subsubsection{Stock model, payoffs and baselines}
In this section we use a $d$-dimensional fractional Brownian motion, with independent coordinates all starting at $X_0=0$, as the underlying process $X_t = W_t^H$.  In contrast to the price processes we used before, this process can become negative.
In the one-dimensional case, we use the identity as ``payoff'' function $g = \operatorname{id}$ as in \citep{becker2019deep}, which can lead to negative ``payoff'' values. 
Moreover, we use the maximum $g(x) =  \max (x_1, x_2, \dots, x_d)$ for any $x = (x_1, x_2, \dots, x_d)\in \mathbb{R}^d$ and the mean $g(x) = 1/d \sum_{i=1}^d x_i$  as ``payoffs'' for higher dimensions, which can also yield negative values.
In particular, this setting leads to an optimal stopping problem outside of the standard discretized American option pricing setting.
We compare RLSM  and RRLSM  to 
DOS and the path-version of DOS (denoted pathDOS for our implementation of it and pathDOS-paper for results reported from \citep{becker2019deep}), where the entire path until the current date is used as input \citep{becker2019deep}. Moreover, we test RFQI  and its recurrent and path-version in this setting.

For two values of the Hurst parameter the optimal value can be computed explicitly. In particular, for $H=0.5$ we have a Brownian motion and therefore the optimal value is $0$ and for $H=1$ we have a fully correlated process (i.e., all information is known after the first step), where the optimal value is approximately $0.39495$ \citep{becker2019deep}.

\subsubsection{Results and discussion}
For $d=1$, we clearly see the outperformance of the algorithms processing information of the path compared to the ones using only the current value as input (Figure~\ref{fig:HurstPlot} top left). 
Moreover, this application highlights the limitation of reinforcement learning techniques when applied in non-Markovian settings as discussed in \citep{kaelbling1996reinforcement}. In particular, RFQI, the randomized RNN version of it (RRFQI) and its path-version do not work well in this example (Figure~\ref{fig:HurstPlot} top right).
This poor performance was consistent under varying hyper-parameters.

%\begin{wrapfigure}{R}{0.50\textwidth}
\begin{figure}
\center
%\begin{figure}[htp!]% [H] is so declass\'e!
\includegraphics[width=0.45\textwidth]{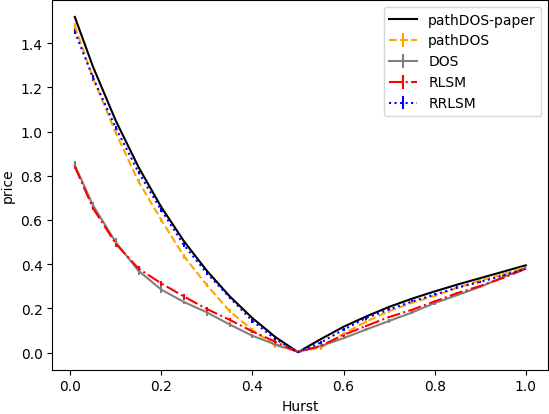}
\includegraphics[width=0.45\textwidth]{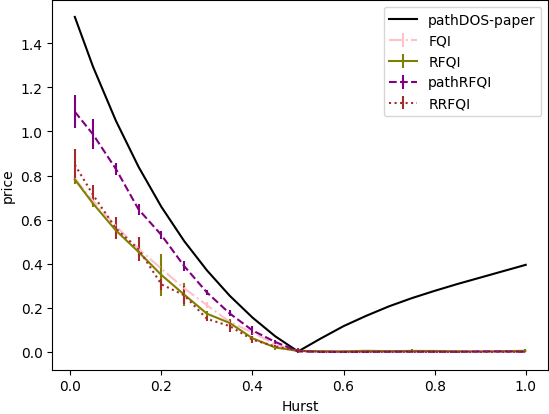}
\includegraphics[width=0.45\textwidth]{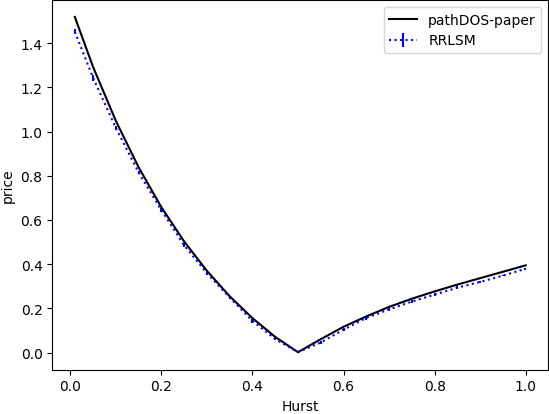}
%\vspace{-0.6cm}
%\end{figure}
\caption{Top left: algorithms processing  path information outperform. Top right: reinforcement learning algorithms do not work well in non-Markovian cases. Bottom: RRLSM  achieves similar results as reported in \citep{becker2019deep}, while using only 20K paths instead of 4M for training which took only $1s$ instead of the reported $430s$.}
\label{fig:HurstPlot}
%\end{wrapfigure}
\end{figure}

RRLSM  achieves very similar results to those reported for pathDOS in \citep{becker2019deep} with an MSE of $0.0005$ between their reported values and ours, while using only 20K instead of 4M paths (Figure~\ref{fig:HurstPlot} bottom). RRLSM  needs only $1s$ to be trained  in contrast to $430s$  reported in \citep{becker2019deep}. 
The longer training times can partly be explained by the larger amount of paths used.
However, our implementation of pathDOS using the same number of 20 hidden nodes as RRLSM  and also being trained on 20K  paths (hence completely comparable to the training of RRLSM) takes approximately $175s$ and achieves slightly worse results than RRLSM  (Figure~\ref{fig:HurstPlot} top left) with an MSE of $0.0018$. 
%In a different setup, similar to the evaluation of RRLSM, our implementation of pathDOS using the same number of 20 hidden nodes as RRLSM  is trained on 20K  paths. Then pathDOS takes $154s$ and achieves worse results than RRLSM  (Figure~\ref{fig:HurstPlot} left). 
The exact prices displayed in Figure \ref{fig:HurstPlot} are provided in Appendix \ref{sec:Stopping of Fractional Brownian Motion Table}.

For higher dimensions, we use the small Hurst parameter $H=0.05$ for which a big difference between the standard and the path dependent algorithms was visible in the one-dimensional case. 
RLSM  yields very  similar prices as DOS and  RRLSM  yields very similar prices as pathDOS.
However, RLSM  and RRLSM  are considerably faster than DOS and pathDOS  (Table~\ref{table_highdim_hurst_gt}).

\begin{table*}[!h]
\center
\resizebox{\textwidth}{!}{
\begin{tabular}{|cc|r r r r|r r r r|}
\toprule
     & {} & \multicolumn{4}{c |}{price} & \multicolumn{4}{c |}{duration} \\
payoff & $d$ & DOS & pathDOS & RLSM & RRLSM & DOS & pathDOS & RLSM & RRLSM\\
\midrule
Identity & 1  &  0.67 (0.02) &   1.24 (0.01) &  0.65 (0.01) &  1.24 (0.01) &     1m15s &   3m 1s &       0s &       1s \\
\cline{1-10}
\multirow{2}{*}{Max} & 5  &  1.96 (0.01) &  2.15 (0.01) &  2.00 (0.01) &  2.16 (0.01) &     3m 8s &    21m46s &       4s &       1s \\
     & 10 &  2.34 (0.01) &  2.43 (0.01) &  2.40 (0.01) &  2.43 (0.02) &     3m49s &    37m46s &       4s &       2s \\
\cline{1-10}
\multirow{2}{*}{Mean} & 5  &  0.29 (0.01) &  0.53 (0.00) &  0.28 (0.01) &  0.52 (0.01) &     3m40s &    21m 8s &       3s &       1s \\
     & 10 &  0.20 (0.01) &  0.36 (0.00) &  0.21 (0.01) &  0.33 (0.01) &     3m39s &    36m 1s &       5s &       1s \\
\bottomrule
\end{tabular}

}
\caption{Identity, maximum and mean on the fractional Brownian motion with $H=0.05$ and different number of stocks $d$.
}
\label{table_highdim_hurst_gt}
\end{table*}

%\caption{Identity, maximum and mean on the fractional Brownian motion with $H=0.05$ and different number of stocks $d$.}

\subsection{Non-Markovian Stock Models}
In  Section~\ref{sec:The Non-Markovian Case -- Optimally Stopping Fractional Brownian Motions} we saw that the RL based algorithms do not perform well on problems which are highly path dependent. 
In this section, we consider ``intermediate'' problems of typical non-Markovian stock models, where a path dependence exists, but where this path dependence is not very strong.

\subsubsection{Heston without variance as input}
%\input{tables_draft/table_Dim_Heston_MaxCallr0_gt.tex}
%%\caption{Max call option on Heston for different numbers of stocks $d$. }
%\input{tables_draft/table_spots_Dim_Heston_MinPut_gt.tex}
%%\caption{Min put option on Heston  for different numbers of stocks $d$ and varying initial stock price $x_0$. Here $r=2\%$ is used as interest rate. }
%\input{tables_draft/table_Dim_Heston_MaxCall_div_gt.tex}
%\caption{Max call  option on Heston  for different numbers of stocks $d$. Here $r=5\%$ is used as interest rate and $\delta = 10\%$ as dividend rate.}
First, we revisit the Heston model  \eqref{equ:heston model}, but this time without feeding the algorithms the variance, which makes it a non-Markovian problem.
For the max call (Table~\ref{table_Dim_Heston_MaxCallr0_gt1}),  min put (Table~\ref{table_$x_0$s_Dim_Heston_MinPut_gt1}) and max call with dividend (Table~\ref{table_Dim_Heston_MaxCall_div_gt1}) options on Heston without variance, all the algorithms yield very similar prices as on Heston with variance (Tables~\ref{table_Dim_HestonV_MaxCallr0_gt1}, \ref{table_$x_0$s_Dim_HestonV_MinPut_gt1} and \ref{table_Dim_HestonV_MaxCall_div_gt1}), therefore we do only show the tables in  Appendix~\ref{sec:Non-Markovian stock models -- Additional Tables}. In particular, this suggests that even though the Heston model is not Markovian without providing the current variance, this doesn't make a difference  for option pricing.

\subsubsection{Rough Heston}
Moreover,  we test on the rough Heston model, where the variance itself is path-dependent. This model  recently became a very popular choice for modelling financial markets \citep{el2018perfect, ElEuch2019RougheningH, gatheral2020quadratic}.
The rough Heston model \citep{el2018perfect} is defined as
\begin{align*}
dX_t &=  (r - \delta) X_t dt +  \sqrt{v_t} X_t dW_t, \\
v_t &= v_0 + \int_0^t \frac{(t-s)^{H-1/2} }{\Gamma(H + 1/2)} \kappa (v_\infty - v_s)ds + \int_0^t \frac{(t-s)^{H-1/2} }{\Gamma(H + 1/2)} \sigma \sqrt{v_s}dB_s,
\end{align*}
where  $X_0 = x_0$, the Hurst parameter $H \in (0, 1/2)$ and  $(W_t)_{t\geq 0}$ and $(B_t)_{t\geq 0}$ are two $d$-dimensional Brownian motions correlated with coefficient $\rho \in (-1,1)$.
We choose the drift $r = 5 \%$,  the dividend rate $\delta = 10\%$, the volatility of volatility $\sigma = 20\%$, the long term variance $v_\infty = 0.01$, the
 mean reversion speed $\kappa = 2$, the correlation $\rho = -30\%$, the initial stock price $x_0 = 100$ and the  initial variance $v_0 = 0.01$ and consider a max call option on the stock price $X$.

As for the Heston model, also for the rough Heston model there is no significant difference between the computed prices with and without providing the current variance, therefore we only show prices where the current variance was also fed to the algorithms, which is still a non-Markovian setting.
For the max call option on the rough Heston model (with variance) (Table~\ref{table_Dim_RoughHestonV_MaxCall_gt1}), we see  that the reinforcement learning based algorithms FQI and RFQI  do not work well for $d \in \{5, 10\}$ but perform better for $d \in \{ 50, 100\}$.
Overall, DOS, pathDOS, RLSM  and RRLSM  achieve very similar prices, never deviating more than $2.2\%$ from each other. In particular, we do not see a better performance of the path dependent algorithms pathDOS and RRLSM  compared to DOS and RLSM.

\begin{table*}[!h]
\center
\resizebox{\textwidth}{!}{
\begin{tabular}{|c|r r r r r r r r|r r r r r r r r|}
\toprule
{} & \multicolumn{8}{c |}{price} & \multicolumn{8}{c |}{duration} \\
$d$ & LSM & DOS & pathDOS & NLSM & RLSM & RRLSM & FQI & RFQI & LSM & DOS & pathDOS & NLSM & RLSM & RRLSM & FQI & RFQI\\
\midrule
5         &   6.58 (0.05) &   6.56 (0.05) &   6.46 (0.06) &   6.39 (0.06) &   6.50 (0.04) &   6.46 (0.04) &   6.10 (0.08) &   6.33 (0.16) &        0s &       7s &       11s &       3s &       0s &       0s &       15s &       0s \\
10        &   9.41 (0.04) &   9.46 (0.04) &   9.28 (0.05) &   9.27 (0.11) &   9.48 (0.04) &   9.37 (0.05) &   9.19 (0.09) &   9.02 (1.18) &        1s &       7s &       13s &       3s &       0s &       0s &       37s &       0s \\
50        &  13.90 (0.07) &  16.69 (0.06) &  16.47 (0.06) &  15.68 (0.32) &  16.35 (0.04) &  16.37 (0.03) &  16.72 (0.07) &  16.75 (0.04) &    18m51s &       9s &       39s &       4s &       0s &       0s &  1h24m22s &       1s \\
100       &             - &  19.79 (0.05) &  19.51 (0.05) &  18.39 (0.35) &  19.50 (0.04) &  19.49 (0.04) &             - &  19.99 (0.05) &         - &      13s &     1m16s &       6s &       0s &       0s &         - &       1s \\
\bottomrule
\end{tabular}

}
\caption{Max call option on Rough--Heston for different number of stocks $d$. The interest rate is $r=5\%$ and the dividend rate is $\delta = 10\%$.
}
\label{table_Dim_RoughHestonV_MaxCall_gt1}
\end{table*}

%\caption{Max call option on Rough--Heston for different number of stocks $d$. The interest rate is $r=5\%$ and the dividend rate is $\delta = 10\%$.}

%\begin{figure}
%\center
%%\begin{figure}[htp!]% [H] is so declass\'e!
%\includegraphics[width=0.8\textwidth]{plots/RoughHeston-20000-MaxCall-10-100}
%%\vspace{-0.6cm}
%%\end{figure}
%\caption{Results for the rough Heston model.}
%\label{fig:roughHeston}
%%\end{wrapfigure}
%\end{figure}

\subsection{Computation of upper bounds}
\label{sec:Computation of Upper Bounds}

While this work's focus lies on the lower bound approximations, we conduct a small experiment to show that also the upper bound computation works efficiently with our method.
In Table~\ref{table: upper bounds} we show mean and standard deviation (over 10 independent runs) of the upper bound approximations for the price of an American option computed with RLSM. Additionally, we show the lower bound and the midpoint (computed as the average of the lower and upper bound). 
As expected, the upper bound approximations are a bit larger than the lower bound approximations. The same method can also be used to computed upper bound approximations for RFQI, however, their quality is relatively sensitive to the number of training iterations and other hyper-parameters, hence, they are not shown here.

\begin{table*}
\center
\scalebox{0.7}{
\begin{tabular}{|r r||c| c |  c |}
\toprule
d & $x_0$ &  price lower &  price midpoint & price upper \\
\midrule
\begin{filecontents*}{plots/upper_bound_comparison_with_dos2-1.csv}
nbstocks,spot,pricew,midpricew,priceupperboundw,timew,doslowerw,dosupperw,dosmidw,dostimetotw
2,90,7.9772 (0.0512),8.0077 (0.0362),8.0382 (0.0619),11s,8.072,8.075,8.074,54s
2,100,13.7902 (0.0664),13.8627 (0.0552),13.9353 (0.0566),11s,13.895,13.903,13.899,54s
2,110,21.2070 (0.0757),21.2735 (0.0420),21.3399 (0.1001),10s,21.353,21.346,21.349,54s
\end{filecontents*}
\csvreader[head to column names, late after line=\\]{plots/upper_bound_comparison_with_dos2-1.csv}{}%
{ \nbstocks & \spot & \pricew & \midpricew &  \priceupperboundw  }%
\midrule
\begin{filecontents*}{plots/upper_bound_comparison_with_dos2-2.csv}
nbstocks,spot,pricew,midpricew,priceupperboundw,timew,doslowerw,dosupperw,dosmidw,dostimetotw
3,90,11.1869 (0.0373),11.1937 (0.0466),11.2005 (0.0603),10s,11.290,11.283,11.287,55s
3,100,18.5698 (0.0803),18.6061 (0.0401),18.6425 (0.0515),10s,18.690,18.691,18.690,55s
3,110,27.3981 (0.1359),27.4284 (0.0757),27.4588 (0.0706),9s,27.564,27.581,27.573,54s
\end{filecontents*}
\csvreader[head to column names, late after line=\\]{plots/upper_bound_comparison_with_dos2-2.csv}{}%
{ \nbstocks & \spot & \pricew & \midpricew &  \priceupperboundw  }
\midrule
\begin{filecontents*}{plots/upper_bound_comparison_with_dos2-3.csv}
nbstocks,spot,pricew,midpricew,priceupperboundw,timew,doslowerw,dosupperw,dosmidw,dostimetotw
5,90,16.4518 (0.0645),16.4995 (0.0410),16.5473 (0.0479),10s,16.648,16.640,16.644,56s
5,100,25.9604 (0.0856),25.9868 (0.0569),26.0133 (0.0893),9s,26.156,26.162,26.159,56s
5,110,36.5271 (0.1066),36.6024 (0.0610),36.6777 (0.1023),9s,36.766,36.777,36.772,56s
\end{filecontents*}
\csvreader[head to column names, late after line=\\]{plots/upper_bound_comparison_with_dos2-3.csv}{}%
{ \nbstocks & \spot & \pricew & \midpricew &  \priceupperboundw  }
\midrule
\begin{filecontents*}{plots/upper_bound_comparison_with_dos2-4.csv}
nbstocks,spot,pricew,midpricew,priceupperboundw,timew,doslowerw,dosupperw,dosmidw,dostimetotw
10,90,25.9808 (0.0923),26.0235 (0.0568),26.0662 (0.0805),8s,26.208,26.272,26.240,64s
10,100,38.0031 (0.0759),38.0750 (0.0592),38.1469 (0.0952),9s,38.321,38.353,38.337,64s
10,110,50.5117 (0.0689),50.5668 (0.0567),50.6219 (0.0962),8s,50.857,50.914,50.886,65s
\end{filecontents*}
\csvreader[head to column names, late after line=\\]{plots/upper_bound_comparison_with_dos2-4.csv}{}%
{ \nbstocks & \spot & \pricew & \midpricew &  \priceupperboundw  }
\midrule
\begin{filecontents*}{plots/upper_bound_comparison_with_dos2-5.csv}
nbstocks,spot,pricew,midpricew,priceupperboundw,timew,doslowerw,dosupperw,dosmidw,dostimetotw
20,90,37.4659 (0.0944),37.5135 (0.0737),37.5611 (0.1440),8s,37.701,37.903,37.802,82s
20,100,51.3532 (0.1073),51.3900 (0.0929),51.4269 (0.1043),9s,51.571,51.765,51.668,82s
20,110,65.2114 (0.0820),65.2774 (0.0451),65.3434 (0.0680),8s,65.494,65.762,65.628,82s
\end{filecontents*}
\csvreader[head to column names, late after line=\\]{plots/upper_bound_comparison_with_dos2-5.csv}{}%
{ \nbstocks & \spot & \pricew & \midpricew &  \priceupperboundw  }
\bottomrule
\end{tabular}
}
\caption{
Lower, midpoint and upper approximations with RLSM  of the price of a max call option on Black--Scholes for different number of stocks $d$ and varying initial stock price $x_0$. The parameters for the stock model are $r=5\%$, $\delta = 10\%$, $N=9$, $T=3$ and $K=100$. We use $m=100'000$ paths and $100$ neurons for the hidden layer.
}
\label{table: upper bounds}
\end{table*}

\subsection{Computation of Greeks}\label{sec:Computation of Greeks}
The Greeks are the sensitivities of the option price to a small change in a given underlying parameter. More precisely, they are partial derivatives of the option prices with respect to different parameters, such as the spot price, time, rate and volatility.
We provide experiments (and the code), where we  compute the most popular Greeks: delta ($\frac{\partial p_0}{\partial x_0} $), gamma ($\frac{\partial^2 p_0}{\partial x_0^2} $), theta ($ \frac{\partial p_0}{\partial t} $), rho ($\frac{\partial p_0}{\partial r} $) and vega ($\frac{\partial p_0}{\partial \sigma} $). 
The straight forward method to compute them is via the finite difference (FD) method. 
For theta, rho and vega, the standard forward finite difference method can be used with our algorithms, however, they turn out to be unstable for NLSM and DOS. 
Therefore, we use the central finite difference method, where the exercise boundary is frozen to be the one of the central point and report results only with this method. 
For computing delta we use the same method, since the others are unstable for all algorithms.
Moreover, the computation of gamma, as a second derivative, turns out to be unstable when computed with the second order finite difference method, even when using the same technique as for delta. Therefore, we use two alternative ways to circumvent this instability. 
The first one (PDE method) is specific to the case of an underlying Black--Scholes model, where the Black--Scholes PDE 
\begin{equation*}
\frac{\partial p_0}{\partial t} + \frac{1}{2} \sigma^2 x_0^2  \frac{\partial^2 p_0}{\partial x_0^2} + r x_0 \frac{\partial p_0}{\partial x_0} - r  p_0 = 0
\end{equation*}
can be used to express gamma in terms of the price, delta and theta.
The second one (regression method) is the ``naive method'' suggested in \citep[Section 3.1]{letourneau2019simulated}. It fits a polynomial regression to option prices achieved when distorting the initial price $x_0$ by a noise term $\xi \sim N(0, \epsilon^2)$. Then the price, delta and gamma can easily be computed by evaluating the fitted regression and its first and second derivative (which are easily computed, since polynomial regression is used) at the initial price $x_0$. The parameter $\epsilon$ controls the variance-bias trade-off and has to be chosen by hand. However, the authors also suggested a 2-step method that reduces variance and bias, where this parameter is chosen automatically.

\begin{table*}
\center
\resizebox{\textwidth}{!}{
\begin{tabular}{|cc||c c| c c|  c c|c |c| c |}
\toprule
 &  &  \multicolumn{2}{c |}{price} & \multicolumn{2}{c |}{delta} & \multicolumn{2}{c |}{gamma} & theta & rho & vega \\
K & algo &  \multicolumn{1}{c |}{FD} & regr. &  \multicolumn{1}{c |}{FD} & regr. &  \multicolumn{1}{c |}{PDE} & regr. & & & \\
\midrule
\begin{filecontents*}{plots/greeks_comparison1.csv}
algo,strike,pricew,deltaw,gammaw,thetaw,rhow,vegaw,pricewreg,deltawreg,gammawreg
B,36,0.9192 (--),-0.1982 (--),0.0389 (--),-0.7152 (--),-6.7085 (--),10.9100 (--),,,
LSM,36,0.9024 (0.0086),-0.1919 (0.0019),0.0368 (0.0004),-0.6615 (0.0068),-6.8267 (0.0497),10.7107 (0.0918),0.9006 (0.0094),-0.1888 (0.0027),0.0381 (0.0009)
RLSM,36,0.9020 (0.0069),-0.1940 (0.0019),0.0371 (0.0003),-0.6665 (0.0063),-6.8241 (0.0639),10.7590 (0.0629),0.9073 (0.0110),-0.1947 (0.0026),0.0379 (0.0010)
FQI,36,0.8504 (0.0067),-0.1777 (0.0016),0.0329 (0.0003),-0.5753 (0.0047),-7.7168 (0.0676),10.3836 (0.0841),0.8642 (0.0116),-0.1770 (0.0015),0.0329 (0.0011)
RFQI,36,0.9005 (0.0087),-0.1924 (0.0029),0.0368 (0.0005),-0.6612 (0.0105),-6.8282 (0.0891),10.7093 (0.0880),0.8766 (0.0138),-0.1847 (0.0041),0.0369 (0.0009)
NLSM,36,0.8948 (0.0238),-0.2010 (0.0174),0.0368 (0.0013),-0.6427 (0.0124),-6.9383 (0.0625),10.6464 (0.1207),0.8817 (0.0151),-0.1905 (0.0038),0.0380 (0.0006)
DOS,36,0.9068 (0.0093),-0.1957 (0.0024),0.0359 (0.0013),-0.6251 (0.0393),-7.0119 (0.3001),10.7249 (0.1419),0.9081 (0.0106),-0.1940 (0.0033),0.0378 (0.0012)
\end{filecontents*}
\csvreader[head to column names, late after line=\\]{plots/greeks_comparison1.csv}{}%
{ \strike & \algo & \pricew & \pricewreg & \deltaw & \deltawreg & \gammaw & \gammawreg & \thetaw & \rhow & \vegaw}%
\midrule
\begin{filecontents*}{plots/greeks_comparison2.csv}
algo,strike,pricew,deltaw,gammaw,thetaw,rhow,vegaw,pricewreg,deltawreg,gammawreg
B,40,2.3196 (--),-0.4047 (--),0.0611 (--),-0.8446 (--),-11.2405 (--),14.7517 (--),,,
LSM,40,2.2916 (0.0107),-0.3930 (0.0022),0.0579 (0.0003),-0.7711 (0.0051),-11.6289 (0.0573),14.6886 (0.0548),2.2715 (0.0116),-0.4049 (0.0053),0.0637 (0.0009)
RLSM,40,2.2897 (0.0095),-0.3984 (0.0052),0.0583 (0.0006),-0.7721 (0.0067),-11.6357 (0.1092),14.7026 (0.0467),2.2970 (0.0135),-0.4076 (0.0039),0.0610 (0.0012)
FQI,40,2.2593 (0.0121),-0.3661 (0.0017),0.0541 (0.0003),-0.7179 (0.0057),-12.6145 (0.0856),14.7479 (0.0670),2.2078 (0.0161),-0.3934 (0.0038),0.0655 (0.0013)
RFQI,40,2.2239 (0.0407),-0.3623 (0.0140),0.0529 (0.0024),-0.6897 (0.0415),-12.9713 (0.6541),14.6794 (0.0831),2.1252 (0.0374),-0.3654 (0.0123),0.0570 (0.0024)
NLSM,40,2.2586 (0.0149),-0.3830 (0.0136),0.0565 (0.0013),-0.7529 (0.0166),-11.7895 (0.2469),14.6545 (0.0970),2.2599 (0.0236),-0.4034 (0.0035),0.0620 (0.0012)
DOS,40,2.2884 (0.0102),-0.4031 (0.0037),0.0587 (0.0004),-0.7727 (0.0055),-11.5870 (0.1041),14.7045 (0.0579),2.2963 (0.0099),-0.4071 (0.0039),0.0611 (0.0006)
\end{filecontents*}
\csvreader[head to column names, late after line=\\]{plots/greeks_comparison2.csv}{}%
{ \strike & \algo & \pricew & \pricewreg & \deltaw & \deltawreg & \gammaw & \gammawreg & \thetaw & \rhow & \vegaw}%
\midrule
\begin{filecontents*}{plots/greeks_comparison3.csv}
algo,strike,pricew,deltaw,gammaw,thetaw,rhow,vegaw,pricewreg,deltawreg,gammawreg
B,44,4.6629 (--),-0.6654 (--),0.0779 (--),-0.6169 (--),-11.8974 (--),12.8541 (--),,,
LSM,44,4.6141 (0.0175),-0.6476 (0.0037),0.0743 (0.0003),-0.5468 (0.0055),-12.8537 (0.1337),13.1799 (0.1129),4.6177 (0.0120),-0.6693 (0.0051),0.0676 (0.0012)
RLSM,44,4.6167 (0.0205),-0.6541 (0.0067),0.0746 (0.0005),-0.5400 (0.0034),-12.7787 (0.2212),13.0670 (0.1566),4.6407 (0.0178),-0.6699 (0.0036),0.0641 (0.0022)
FQI,44,4.5366 (0.0221),-0.5962 (0.0039),0.0695 (0.0005),-0.5216 (0.0066),-15.2857 (0.1248),14.3877 (0.0471),4.5476 (0.0137),-0.6766 (0.0054),0.0710 (0.0013)
RFQI,44,4.3469 (0.0708),-0.5463 (0.0073),0.0607 (0.0023),-0.3688 (0.0534),-19.9917 (1.0327),15.6835 (0.1199),4.1557 (0.0415),-0.5791 (0.0059),0.0687 (0.0019)
NLSM,44,4.5820 (0.0211),-0.6553 (0.0150),0.0742 (0.0008),-0.5268 (0.0182),-13.1123 (0.9381),12.9567 (0.5430),4.5996 (0.0531),-0.6713 (0.0048),0.0658 (0.0029)
DOS,44,4.6206 (0.0126),-0.6580 (0.0046),0.0747 (0.0003),-0.5350 (0.0050),-12.8331 (0.2236),13.0304 (0.1295),4.6536 (0.0113),-0.6695 (0.0042),0.0641 (0.0016)
\end{filecontents*}
\csvreader[head to column names, late after line=\\]{plots/greeks_comparison3.csv}{}%
{ \strike & \algo & \pricew & \pricewreg & \deltaw & \deltawreg & \gammaw & \gammawreg & \thetaw & \rhow & \vegaw}
\bottomrule
\end{tabular}
}
\caption{
Prices and Greeks computed for different strikes $K$ of a $1$-dimensional put option on  Black--Scholes.  For the binomial (B) algorithm, the spacing  of the FD method is set to $\varepsilon = 10^{-9}$, which is also used for the other algorithms for delta, theta, rho and vega. For the regression method, $\epsilon = 5$ and a polynomial basis up to degree $9$ are used.
}
\label{table: greeks}
\end{table*}

For comparability, we compute the Greeks for the same example as in \citep{letourneau2019simulated}. In particular, we consider a put option on $d=1$ stock following a Black--Scholes model with initial price $x_0=40$, strike $K \in \{ 36, 40, 44 \}$, rate $r=6\%$, volatility $\sigma = 20\%$, $N=10$ equidistant dates, maturity $T=1$, and $m=100'000$ paths. The models are run $10$ times and mean and standard deviations are reported in Table~\ref{table: greeks}. The price, delta and gamma are computed with both, the finite difference (respectively PDE) and the regression method.
As reference we use the binomial model with $N=50'000$ equidistant dates, for which only the finite difference (respectively PDE) method is used.
The hidden size was set to $10$ to account for the smaller input dimension and the payoff was not used as input except for DOS, where it improved the results considerably.
For RLSM  the activation function was changed to Softplus, since this  worked best, although all other tested activation functions did also yield good results.
%For NLSM and DOS the computation of theta (and therefore also gamma when using the PDE method), rho and vega was unstable\footnote{Changing the hidden size and the number of training epochs did not resolve this.}. Changing the forward to the central method did not resolve this issue. 
Overall, RLSM  and DOS with the regression method achieve the best results. Furthermore, we highlight, that the time advantage of RLSM  and RFQI  also comes  into play for the computation of Greeks, when increasing the dimension $d$.

For  RLSM   (with Softplus activation) we  additionally show  stability plots of the Greeks with respect to the spot price.  In particular we use the same setting as before of a put option on $d=1$ stock following a Black--Scholes model with rate $r=6\%$, $N=10$ dates and $m=100'000$ paths, however, we fix the strike $K=40$ and vary the spot price $x_0 \in [20, 60]$. Moreover, we vary the volatility $\sigma \in \{ 0.1, 0.2, 0.3 \}$ and the maturity $T \in \{ 0.5, 1, 2 \}$. For each combination, we run the algorithm $5$ times and plot the median of the results in Figure~\ref{fig:greeks stability plots}. 
Up to small numerical instabilities the resulting curves are smooth as known from theory. We observe the same qualitative behaviour of the Greeks as was shown in \citep[Section 5.2 - 5.6]{Bellefroid2022}.
  \begin{figure}[!h]
        \centering
            \includegraphics[width=\textwidth]{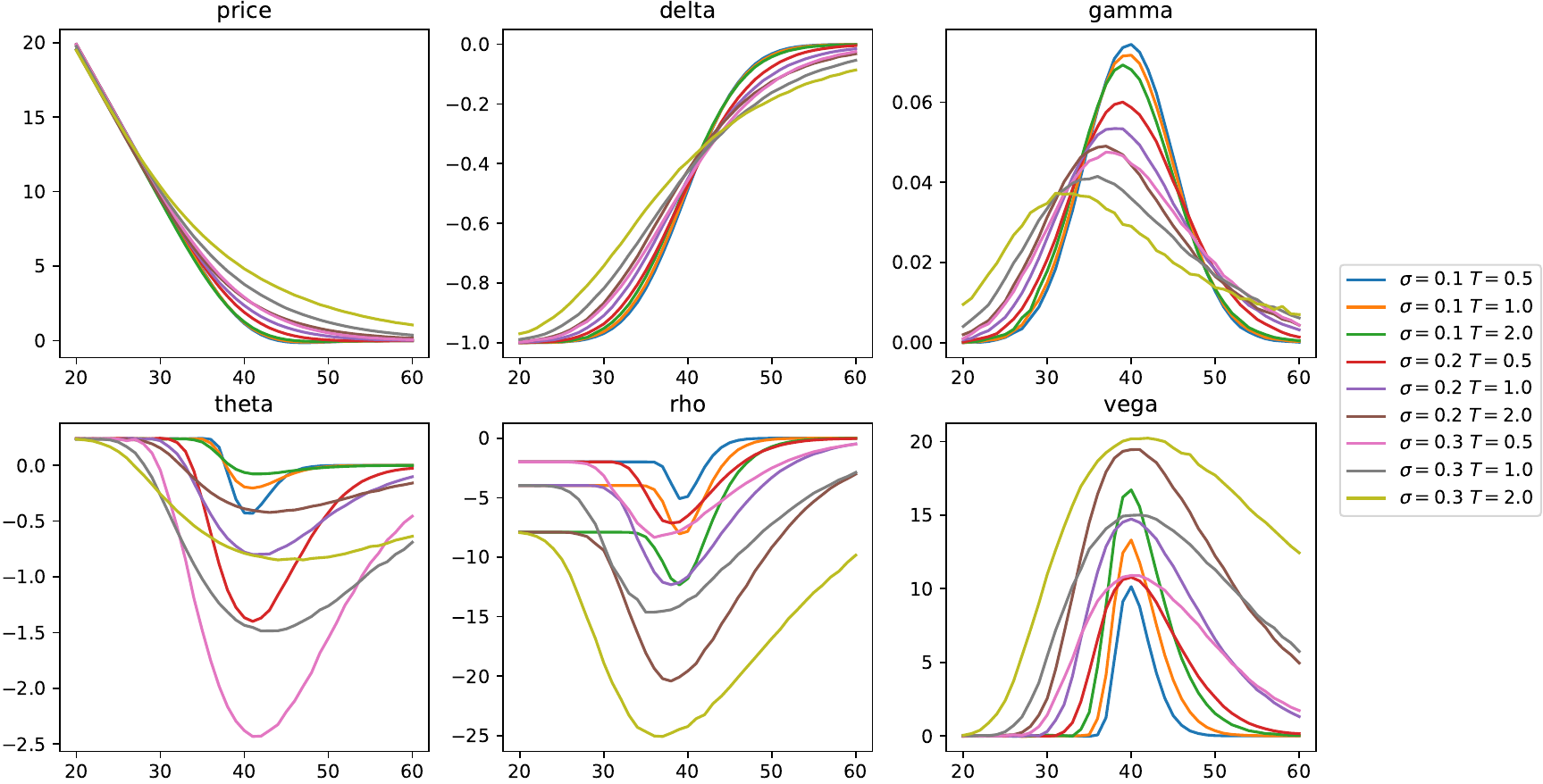}
%            \caption[Network2]%
%            {{\small Network 1}}    
        \caption{Median of the price and Greeks  computed with RLSM  plotted against the spot price $x_0$ for different volatilities $\sigma$ and maturities $T$. The price, delta and gamma are computed with the regression method with $\epsilon = 5$ and a polynomial basis up to degree $2$.}
        \label{fig:greeks stability plots}
    \end{figure}

\subsection{Discussion on the sensitivity to the randomness in the hidden layers}

We perform a test specifically designed to study the model's sensitivity to the randomness of the weights in the hidden layers.
In our previous tests in this paper we performed $10$ runs, where  a different set of paths and different weights of the hidden layer were chosen for each run. 
In order to test the sensitivity to the randomness of the weights, we  perform an experiment with $10$ runs, where only the set of hidden weights are different for each run, while the paths are the same. 

We compare RLSM  and NLSM in the setting of a 1-dimensional Black-Scholes call option with spot $x_0=100$ and strike $K=100$, where we use $100'000$ paths and $10$ exercises dates with $20$ hidden nodes and either $10$, $30$ or $50$ epochs of training for NLSM.

In order to have a fair comparison, we do not fix the initial weights of NLSM, as it would be equivalent to reusing the same random weights for RLSM  in each run, with the possibility of having a good or bad initialisation.
Hence, similar to RLSM's sensitivity to the randomness of the weights in the hidden layer, NLSM is sensitive to the randomness in the initialization of the weights (of the hidden layer). 
In order to reduce this sensitivity in the algorithms, one should always take the average of several runs with different sets of weights (and paths). This can be easily done in parallel in order to reduce the computation time.
The results of this sensitivity analysis are given in Table~\ref{table:1.}.
We see that the sensitivity of NLSM to the randomness of the initialization depends on the number of epochs of the training, becoming smaller with longer training.

In order to further reduce the sensitivity of RLSM  to the randomness of the hidden layer weights we propose a variant of it, which we call RLMSreinit. Instead of using the same random weights for each date, we use different ones, which has an averaging effect and therefore reduces the variance in multiple runs. 

\begin{table*}[!h]
\center
\resizebox{\textwidth}{!}{
\begin{tabular}{| c | c | c | c |  c |c |c| c | c |}
\toprule
algo &  $\#$epochs &  {price} & {delta} & {gamma} & theta & rho & vega & duration\\
\midrule
\begin{filecontents*}{tables/sensitivity_to_randomness.csv}
,algo,strike,greeks_method,hiddensize,nbepochs,pricew,deltaw,gammaw,thetaw,rhow,vegaw,comptimemedianw
7,NLSM,100,,20,10,8.8961 (0.0356),0.5845 (0.0018),0.0194 (0.0001),-4.8682 (0.0142),46.0934 (1.2538),39.4202 (0.0613),7.86s
8,NLSM,100,,20,30,8.9175 (0.0171),0.5836 (0.0008),0.0194 (0.0001),-4.8723 (0.0206),47.2922 (1.2718),39.3781 (0.0640),19.53s
9,NLSM,100,,20,50,8.9178 (0.0160),0.5835 (0.0006),0.0194 (0.0001),-4.8780 (0.0166),46.6717 (1.6730),39.3755 (0.1004),35.03s
0,RLSM,100,,20,,8.9439 (0.0179),0.5828 (0.0004),0.0195 (0.0001),-4.8952 (0.0153),48.0459 (0.7537),39.3425 (0.0951),1.37s
10,RLSMreinit,100,,20,,8.9425 (0.0081),0.5828 (0.0002),0.0195 (0.0000),-4.8887 (0.0078),47.7329 (0.3765),39.3477 (0.0573),1.52s
\end{filecontents*}
\csvreader[head to column names, late after line=\\]{tables/sensitivity_to_randomness.csv}{}%
{  \algo & \nbepochs & \pricew &  \deltaw  & \gammaw  & \thetaw & \rhow & \vegaw & \comptimemedianw}%
\bottomrule
\end{tabular}
}
\caption{
Prices and Greeks for NLSM (with different number of training epochs), RLSM  and RLSMreinit with standard deviations computed over 10 runs with different initializations on the same set of paths.
}
\label{table:1.}
\end{table*}

\section{Conclusion}
Based on a broad study of machine learning based approaches to approximate the solution of optimal stopping problems, we introduced two simple and powerful approaches, RLSM  and RFQI. As state-of-the-art algorithms, they are very simple to implement and have convergence guarantees. Moreover, similarly to the neural network methods, they are easily scalable to high dimensions and there is no need to choose basis functions by hand.  Furthermore, in our empirical study we saw that RLSM  and RFQI  are considerably faster than existing algorithms for high dimensional problems. In particular, up to $2400$ (and $4800$)  times faster than LSM (and FQI respectively) with basis functions of order 2; 5 to 16 times faster than NLSM and 20 to 66 times faster than DOS. 

In our Markovian experiments, RFQI  often achieves the best results and if not, usually is very close to  the best performing baseline method under consideration, reconfirming that reinforcement learning methods surpass backward induction methods. 

In our non-Markovian experiments on fractional Brownian Motion, our randomized recurrent neural network algorithm RRLSM  achieves similar results as the path-version of DOS, while requiring less training data and being much faster. However, this example also brought up the limitations of reinforcement learning based approaches, in particular of RFQI, which do not work well in those non-Markovian experiments. 

In our non-Markovian experiments on rough Heston, we concluded that there is no need of using a recurrent neural network, since  RLSM  has similar results as RRLSM. This is also the case with DOS and pathDOS.

Overall, the speed of our algorithms is very promising for applications in high dimensions and with many discretization times, where existing methods might become impractical and where our methods show very reliable  performance. 
To summarize, we suggest to use RFQI  for Markovian problems especially in high-dimensional settings and RLSM  for low-dimensional settings or when computing Greeks and upper bounds, RLSM  for  non-Markovian processes which do not have a strong path-dependence, as the stock price of rough Heston and finally RRLSM  for non-Markovian processes which have a strong path-dependence like fractional Brownian Motion.

%\section{Limitations}

% Acknowledgements should go at the end, before appendices and references

\if\jmlrstyle1
\acks{%
The authors would like to thank Sebastian Becker, Patrick Cheredito, Blanka Horvath, Arnulf Jentzen, Hartmut Maennel and Louis Paulot for helpful feedback and discussions. In addition, the authors would like to warmly thank the quant team of  Cr{\'e}dit Agricole CIB, and in particular Arthur Semin, Ryan Kurniawan and Wail El Allali for the great collaboration, which considerably improved the paper. Thanks to this collaboration, we provide the computation of the Greeks, we improved the sensitivity to the randomness of the hidden layers of RLSM and we improved the proof of convergence of RLSM. The authors would also like to thank the anonymous reviewers for their feedback leading to significantly improvements of the paper. Moreover, the authors would like to acknowledge support for this project from the Swiss National Science Foundation (SNF grant 179114).%
}
\else
\section{Acknowledge}
\fi

\if\twocolumns1
  \end{multicols}
\fi

\newpage
\appendix

\section{Convergence of the randomized least square Monte Carlo (RLSM)}
\label{sec:Convergence of the Longstaff-Schwartz version}
We first introduce some technical notation that will be helpful for the proofs. Then we describe the  steps from the theoretical idea of RLSM  to its implementable version that was presented in  Section~\ref{sec:algo 1}. These descriptions and proofs are based on  \citep{Tsitsiklis2001Regression, Clement2001AnAO}, in particular, our theoretical results are a direct consequence of these works and the universal approximation theorem of \cite{Zhang2012}. Nevertheless, we give a detailed description here for completeness. 

\subsection{Definitions}
We assume to have a sequence of infinitely many \emph{random} basis functions $\phi = (\phi_k)_{k \geq 1}$, where each $\phi_k$ is of the form
\begin{equation*}
\phi_k: \mathbb{R}^d \to \mathbb{R}, x \mapsto \phi_k(x)\coloneqq \sigma(\alpha_k^\top x + \beta_k),
\end{equation*}
with $\sigma$ a bounded activation function, $\alpha_k \in \mathbb{R}^d$ and $\beta_k \in \mathbb{R}$. The parameters $\alpha_k$ and $\beta_k$ have i.i.d. entries with a standard Gaussian distribution, hence the name \emph{random} basis functions. 
With $(\tilde \Omega, \tilde{\mathcal{F}}, \tilde{\mathbb{P}})$ we denote the probability space on which the random weights are defined. For each $K \in \mathbb{N}$ we define the operator  $\Phi_K$ acting on $\theta = (\theta_1, \dotsc, \theta_K) \in \mathbb{R}^K$ by
\begin{equation*}
(\Phi_K \theta) (x) \coloneqq \theta^\top \phi(x)  \coloneqq \sum_{k=1}^K \theta_k \phi_k(x) .
\end{equation*}
In particular, $\Phi_K$ is the operator producing a linear combination of the first $K$ random basis functions. We assume to have a Markovian, discrete time stochastic process $X = (X_0, \dotsc, X_N)$ defined on a filtered probability space $(\Omega, \mathcal{F}, (\mathcal{F}_n)_{n=0}^N, \mathbb{P})$. In particular, each $X_n$ is a $\mathcal{F}_n$-measurable random variable.
We assume that there exists an absolutely continuous measure $\mathbb{Q} \ll \mathbb{P}$, the pricing measure, and that the distribution of $X_n$ under $\mathbb{Q}$ is $\pi_n$.
For expectations with respect to these random variables under $\mathbb{Q}$, we write $\mathbb{E}[\cdot]$. For $0 \leq n \leq N$ we use the norm
\begin{equation*}
\lVert f \rVert_{\pi_n}^2 \coloneqq \mathbb{E}[|f(X_n)|_2^2] = \int_{\mathbb{R}} |f(x)|_2^2 d\pi_n(x),
\end{equation*}
where $|\cdot |_2$ is the Euclidean norm and $f$ a measurable function. We introduce the operators $E_n$ and $\Pi_n^K$ defined by 
\begin{align*}
&(E_nJ)(x) \coloneqq \mathbb{E}[J(X_{n+1}) | X_n =x ], \\
&(\Pi_n^K J) \coloneqq \arg \min_{\Phi_K \theta}\lVert J - \Phi_K \theta \rVert_{\pi_n},
\end{align*}
for $J \in L^2(\pi_n)$.
%When the time step is clear from the context, we use $E$ instead of $E_n$.
With $\hat E_n$ we denote the one-sample approximation of $E_n$, i.e.  $(\hat E_n J)(X_n) = J(X_{n+1})$, which is better understood in terms of a realization of $x = (x_0, \dotsc, x_N)$ of $X$ as $(\hat E_n J)(x_n) = J(x_{n+1})$.
Moreover, $\hat \Pi_n^K$ is the Monte Carlo approximation of $\Pi_n^K$, i.e. if $x_n^1, \dotsc, x_n^m$ are i.i.d.\ samples of $\pi_n$, then $(\hat\Pi_n^K J) \coloneqq \arg \min_{\Phi_K \theta} \frac{1}{m} \sum_{i=1}^m \left( J(x_n^i) - (\Phi_K \theta) (x_n^i) \right)^2$.
In the following, we write $\Pi_n$ and $\hat \Pi_n$ whenever $K$ is fixed.

The payoff at  any exercise time $n$ is given by $g(X_n)$ and we assume that they are square integrable, i.e. $\lVert g(X_n) \rVert_{\pi_n} < \infty$.

\subsection{Theoretical description of RLSM}\label{sec:Theoretical Description of OurAlgo}
We first introduce  the \emph{exact algorithm} to compute the continuation value and then give  definitions of the 2-step approximation of this exact algorithm.
%In the following we introduce abstract definitions for the different versions of the reservoir algorithm used to approximate the exact algorithm to compute the continuatiion value.
%We do this in two steps. 
The first step is to introduce projections on the subspace of functions spanned by $\Phi_K$, while assuming that (conditional) expectations can be computed exactly. We call this the \emph{idealized algorithm}. We remark that also the projection itself is based on minimizing an expectation.
The second step is to introduce Monte Carlo and one-sample approximations of the projections and (conditional) expectations using $m$ sample paths. This we call the \emph{implementable algorithm}, since it can actually be implemented.
Our goal is then to show that the price computed with those two approximation steps converges to the true price, when $K$ and $m$ increase to infinity.

\subsubsection{Exact algorithmic}
The \emph{continuation value} is the expected discounted payoff at the current time conditioned on a decision not to exercise the option now.
The \emph{exact algorithmic} definition of the continuation value is defined backwards step-wise as in \citep{Tsitsiklis2001Regression} as 
\begin{equation}\label{equ:exact cont value}
\begin{split}
\begin{cases}
Q_{N-1} &\coloneqq \alpha E_{N-1} g,\\
Q_n  &\coloneqq \alpha E_n \max(g, Q_{n+1}).
\end{cases}
\end{split}
\end{equation}

\subsubsection{Idealized algorithm}
Our \emph{idealized algorithm} to compute the continuation value, written similar as in \citep{Tsitsiklis2001Regression}, is defined for  fixed $K$ as
\begin{equation}\label{equ:idealised approx cont value}
\begin{split}
\begin{cases}
\tilde Q_{N-1}^K &\coloneqq \alpha E_{N-1} P_N^K,\\
\tilde Q_n^K  &\coloneqq \alpha E_n P_{n+1}^K,
\end{cases}
\end{split}
\end{equation}
where 
\begin{equation*}\label{equ:idealised decision}
\begin{split}
\begin{cases}
P_N^K & \coloneqq g, \\
P_n^K &\coloneqq g \mathbf{1}_{g \geq \alpha \Pi_{n}^K E_{n} P_{n+1}^K} + \alpha E_{n} P_{n+1}^K \mathbf{1}_{g < \alpha \Pi_{n}^K E_{n} P_{n+1}^K}.
\end{cases}
\end{split}
\end{equation*}
In particular, $P_n^K$ can be interpreted as the choice of the algorithm at time step $n$, to either execute and take the payoff or to continue with the expected discounted future payoff.
We drop the superscript $K$ whenever it is clear from the context which $K$ is meant.
We see from this equation, that the  difference from the idealized algorithm in \citep[described in (1) and before Theorem 1]{Tsitsiklis2001Regression} is, that we use the $\tilde Q_{n+1}$ instead of its linear approximation with the random basis functions $\Pi_n\tilde Q_{n+1}$, if we decide to continue. However, the decision to continue or to stop, is still based on the approximation $\Pi_n\tilde Q_{n+1}$ as it is also the case in the idealized algorithm \citep{Tsitsiklis2001Regression}. If the linear approximation is exact, both algorithms produce the same output, but if it is not exact, our algorithm uses a better approximation of the continuation value.

\subsubsection{Implementable algorithm}
Finally, we define our \emph{implementable algorithm} to compute the continuation value, which is an approximation of the idealized algorithm using the  approximations $\hat E_n$ and $\hat \Pi_n^K$ as
\begin{equation}\label{equ:implementable approx cont value}
\begin{split}
\begin{cases}
\hat{\tilde Q}_{N-1}^K &\coloneqq \alpha \hat E_{N-1} \hat P_N^K, \\%= \alpha \hat E_{N-1} g,\\
\hat{\tilde Q}_n^K  &\coloneqq \alpha \hat E_n \hat P_{n+1}^K,\\% =\alpha \hat E_n \hat L_n^K \hat{\tilde Q}_{n+1}^K,
\end{cases}
\end{split}
\end{equation}
where 
\begin{equation*}
\begin{split}
\begin{cases}
\hat P_N^K &\coloneqq g, \\
\hat P_n^K &\coloneqq g \mathbf{1}_{g \geq \alpha \hat\Pi_{n}^K \hat E_{n} \hat P_{n+1}^K} + \alpha \hat E_{n} \hat P_{n+1}^K \mathbf{1}_{g < \alpha \hat \Pi_{n}^K \hat E_{n} \hat P_{n+1}^K}.
\end{cases}
\end{split}
\end{equation*}
Also here we drop the superscript $K$ whenever it is clear from the context which $K$ is meant.

\subsection{Preliminary results}\label{sec:Preliminary results}
The following result is similar to \cite[Theorem 3]{Zhang2012} and states, that the error of the approximation of any integrable function by randomized neural networks  converges $\tilde{\mathbb{P}}$-a.s. to $0$ as the number of hidden nodes goes to infinity, where $\tilde{\mathbb{P}}$ is the probability measure associated with the random weights. 
While \cite[Theorem 3]{Zhang2012} shows universal approximation in $L^p$-norm with respect to the Lebesgue integral on a compact subset, we show it with respect to a probability measure on the entire space. We note that our result also holds when replacing the probability measure with a finite measure. In particular, our result implies the result of \cite{Zhang2012}, by using the finite measure that coincides with the Lebesgue measure on the respective compact set and vanishes outside. For completeness, we give an independent proof of our result here.

\begin{theorem}\label{thm: convergence of projection}
Let $0 \leq n \leq N-1$ and $J$ be an square integrable function, i.e. $\lVert J \rVert_{\pi_n} <\infty$, then
\begin{equation*}
\lVert \Pi_n^K J - J \rVert_{\pi_n} \xrightarrow[K \to \infty]{\tilde{\mathbb{P}}\text{-a.s.}} 0.
\end{equation*}
\end{theorem}

\begin{lemma}\label{lem:existence open neighbourhood}
Let $\mathcal{X}, \mathcal{Y}$ be a normed spaces and $\mu$ a probability measure on $\mathcal{X}$ with its Borel $\sigma$-Algebra.
Let $J: \mathcal{X} \times \mathcal{Y} \to \mathbb{R}$ be a bounded function such that for each $C > 0$ and each $x \in \mathcal{X}$ with $\lVert x \rVert < C$ the function $y \mapsto J(x,y)$ is Lipschitz continuous with Lipschitz constant $L_C$ (depending only on $C$ but not on $x$). Then for any $\epsilon > 0$ and $y \in \mathcal{Y}$ there exists an open neighbourhood $O(y, \epsilon) \subset \mathcal{Y}$ such that $y \in O(y, \epsilon)$ and for every $\tilde y \in O(y, \epsilon)$ we have
\begin{equation*}
\int_\mathcal{X} \lvert J(x,y) - J(x, \tilde{y}) \rvert^2 d\mu(x) < \epsilon.
\end{equation*}
\end{lemma}
\begin{proof}
Since $J$ is bounded, there exists $M$ such that $\lvert J \rvert <M$. Since $\mu$ is finite, there exists some $C$ such that $\mu(\{ x \in \mathcal{X} | \lVert x \rVert \geq  C \}) < \frac{\epsilon}{8 M^2}$. Hence, for any $\tilde{y} \in \mathcal{Y}$
\begin{equation*}
\int_\mathcal{X} \lvert J(x,y) - J(x, \tilde{y}) \rvert^2 \mathbf{1}_{\lVert x \rVert \geq  C} \, d\mu(x)
< \int_\mathcal{X} (2 M)^2 \mathbf{1}_{\lVert x \rVert \geq C} \, d\mu(x)  < \epsilon/2.
\end{equation*}
Let us choose $O(y, \epsilon) \coloneqq B\left(y, \sqrt{\frac{\epsilon}{2 L_C^2}}\right)$, the open ball with radius $\sqrt{\frac{\epsilon}{2 L_C^2}}$ and center $y$. Then for any $x$ with $\lVert x \rVert < C$ and $\tilde{y} \in O(y, \epsilon)$ we have $\lvert J(x, y) - J(x, \tilde{y}) \rvert < L_C \lVert y - \tilde{y} \rVert < \sqrt{\epsilon/2}$.
Therefore, 
\begin{equation*}
\int_\mathcal{X} \lvert J(x,y) - J(x, \tilde{y}) \rvert^2 \mathbf{1}_{\lVert x \rVert < C} \, d\mu(x) 
< \int_\mathcal{X} \epsilon /2  \, d\mu(x)  < \epsilon/2.
\end{equation*}
Together, this yields the result.
\end{proof}

We first prove the following weaker version of the statement of Theorem \ref{thm: convergence of projection}.

\begin{lemma}\label{lem: for thm convergence of projection}
Let $1 \leq n \leq M$ and $J$ be an integrable function, i.e. $\lVert J \rVert_{\pi_n} <\infty$, then
\begin{equation*}
\lVert \Pi_n^K J - J \rVert_{\pi_n} \xrightarrow[K \to \infty]{\tilde{\mathbb{P}}} 0.
\end{equation*}
\end{lemma}
\begin{proof}
We fix $\varepsilon > 0$. We have to show that 
\begin{equation*}
\lim_{K \to \infty} \tilde{\mathbb{P}}\left[\lVert \Pi_n^K J - J \rVert_{\pi_n} > \varepsilon \right] = 0.
\end{equation*}
By the universal approximation theorem \cite[Theorem 1]{hornik1991approximation}, there exists a 1-hidden layer neural network $\hat J$ with $n_1$ hidden neurons such that $\lVert \hat J - J \rVert_{\pi_n} < \varepsilon/2$.
Without loss of generality we assume that the bias of the last layer is $0$, which can be established by introducing another hidden neuron which is constant as function of its input.
We notice that also $\Phi_K \theta$ is a 1-layer neural network with $K$ hidden nodes (and the same activation function as $\hat J$), where the weights of the input layer are i.i.d. sampled of a normal distribution and then fixed and the weights of the output layer are $\theta$.
Let $\theta^\star \in \mathbb{R}^{n_1}$ be the weights of the output layer of $\hat{J}$.
For each of the hidden nodes $1 \leq \nu \leq n_1$ of $\hat J$ we denote the mapping from the input to this hidden node by $\hat{J}^h_\nu$. Let $W_\nu$ be the weights defining $\hat{J}^h_\nu$, and denote by $\Psi$ the operator mapping the weights to the corresponding neural network layer, such that $\Psi W_\nu = \hat{J}^h_\nu$. 
We know from \cite{LipDNN} that $\hat{J}^h_\nu$ is Lipschitz continuous w.r.t. the weights for a bounded input $\lVert x \rVert \leq N$. Moreover, since the activation function is bounded, so is $\hat{J}^h_\nu$. Therefore, by Lemma \ref{lem:existence open neighbourhood} there exists an open neighbourhood  $\mathcal{W}_\nu$ of $W_\nu$ such that for all $W \in \mathcal{W}_\nu$ we have 
\begin{equation*}
\lVert \Psi W - \Psi W_\nu \rVert_{\pi_n} < \tfrac{\varepsilon}{2 \sqrt{n_1} \lvert \theta^\star \rvert_2}.
\end{equation*} 
For any non-empty open set, the probability that a standard Gaussian random variable lies in this open set is positive. Let $V_k$ be the weights of the $k$-th random map $\phi_k$, i.e. $\phi_k = \Psi V_k$ and note that $V_k$ is a vector of i.i.d. standard Gaussian random variables.
Since $\mathcal{W}_\nu$ is open, we therefore have that $\tilde{\mathbb{P}}[V_k \in \mathcal{W}_\nu] > 0$.
By independence of the weights we have that with probability 1 each $\hat J ^h_\nu$ is approximated well by some $\phi_k$ when $K \to \infty$. Indeed, let $K = n_1 \tilde{K}$, then we have
\begin{equation*}
\begin{split}
\tilde{\mathbb{P}}  & \left[ \forall 1 \leq \nu \leq n_1 \; \exists 1 \leq k \leq K : V_k \in \mathcal{W}_\nu \right]  \\
	& \geq  \tilde{\mathbb{P}}\left[ \forall 1 \leq \nu \leq n_1 \; \exists (\nu -1)\tilde{K} < k \leq \nu \tilde{K} : V_k \in \mathcal{W}_\nu \right] \\
	& = \prod_{\nu = 1}^{n_1} \tilde{\mathbb{P}}\left[\exists (\nu -1)\tilde{K} < k \leq \nu \tilde{K} : V_k \in \mathcal{W}_\nu \right] \\
	& = \prod_{\nu = 1}^{n_1} \left( 1 - \tilde{\mathbb{P}}\left[\forall (\nu -1)\tilde{K} < k \leq \nu \tilde{K} : V_k \notin \mathcal{W}_\nu \right] \right) \\
	& = \prod_{\nu = 1}^{n_1} \left( 1 - \tilde{\mathbb{P}}\left[V_1 \notin \mathcal{W}_\nu \right]^{\tilde{K}} \right) \xrightarrow{\tilde{K} \to \infty} 1,
\end{split}
\end{equation*}
where we used in line 3 and 5 independence of the weights and for the limit that $\tilde{\mathbb{P}}\left[V_1 \notin \mathcal{W}_\nu \right] < 1$.
We define $\tilde{\theta}^\star \in \mathbb{R}^K$ to have the $k$-th coordinate equal to $\theta^\star_\nu$ if $k = k(\nu) \coloneqq \arg \min_{j} \lVert \phi_j - \hat{J}^h_\nu \rVert_{\pi_n}$ or $0$ otherwise. Here we assume without loss of generality that all $k(\nu)$ are different (if they are not, the weights are summed up). Then we have 
\begin{equation*}
\begin{split}
\lVert \Pi_n^K J - J \rVert_{\pi_n} 
	 \leq \lVert \Phi_K \tilde{\theta}^\star - J \rVert_{\pi_n} 
	& \leq \lVert \Phi_K \tilde{\theta}^\star - \hat J \rVert_{\pi_n} + \lVert \hat{J} - J \rVert_{\pi_n} \\
	& \leq  \lVert \Phi_K \tilde{\theta}^\star - \hat J \rVert_{\pi_n} + \varepsilon/2
\end{split}
\end{equation*}
and therefore
\begin{equation*}
\begin{split}
\tilde{\mathbb{P}} \left[ \lVert \Pi_n^K J - J \rVert_{\pi_n} > \varepsilon \right]  
	 & \leq \tilde{\mathbb{P}}\left[ \lVert \Phi_K \tilde{\theta}^\star - \hat J \rVert_{\pi_n} > \varepsilon/2 \right] \\
	& = \tilde{\mathbb{P}}\left[ \lVert (\Psi V_k)_{k=1}^K \tilde{\theta}^\star - (\Psi W_\nu)_{\nu=1}^{n_1} \theta^\star \rVert_{\pi_n} > \varepsilon/2 \right].
\end{split}
\end{equation*}
Now we notice, that $\tilde{\theta}^\star_k$ is $0$ unless $k = k(\nu)$ for some $1 \leq \nu \leq n_1$. Hence,
\begin{equation*}
\begin{split}
\tilde{\mathbb{P}}  \left[ \lVert \Pi_n^K J - J \rVert_{\pi_n} > \varepsilon \right]  
	& \leq \tilde{\mathbb{P}}\left[ \lVert (\Psi V_{k(\nu)})_{\nu=1}^{n_1} {\theta}^\star - (\Psi W_\nu)_{\nu=1}^{n_1} \theta^\star \rVert_{\pi_n} > \varepsilon/2 \right] \\
	& \leq \tilde{\mathbb{P}}\left[ \lVert (\Psi V_{k(\nu)})_{\nu=1}^{n_1}  - (\Psi W_\nu)_{\nu=1}^{n_1} \rVert_{\pi_n} > \tfrac{\varepsilon}{2 \lvert \theta^\star  \rvert_2} \right] \\
	& \leq \tilde{\mathbb{P}}\Big[ \exists 1 \leq \nu \leq n_1:  \lVert \Psi V_{k(\nu)}  - \Psi W_\nu \rVert_{\pi_n} > \tfrac{\varepsilon}{2 \sqrt{n_1} \lVert \theta^\star  \rVert} \Big] \\
	& = 1 - \tilde{\mathbb{P}}\left[ \forall 1 \leq \nu \leq n_1 \exists 1 \leq k \leq K : V_k \in \mathcal{W}_\nu \right] \; \xrightarrow{K \to \infty} 0.
\end{split}
\end{equation*}
For the second inequality we used the Cauchy-Schwarz inequality and that $\lVert \theta^\star \rVert_{\pi_n} = \lvert \theta^\star \rvert_2 $. In the last equality we used that $k(\nu)$ is chosen such that the distance between $\Psi V_{k(\nu)}$ and $\Psi W_\nu$ is minimized.
\end{proof}

\begin{proof}[Proof of Theorem \ref{thm: convergence of projection}.]
By Lemma \ref{lem: for thm convergence of projection} we know that $\lVert \Pi_n^K J - J \rVert_{\pi_n} \xrightarrow[K \to \infty]{\tilde{\mathbb{P}}} 0$ which implies that there exists a subsequence $(K_m)_{m\geq 1}$ s.t. $\lVert \Pi_n^{K_m} J - J \rVert_{\pi_n} \xrightarrow[m \to \infty]{\tilde{\mathbb{P}}\text{-a.s.}} 0$. Let $\hat{\Omega} \subset \tilde{\Omega}$ with $\tilde{\mathbb{P}}(\hat{\Omega})=1$ be the set on which this convergence holds and let $\omega \in \hat{\Omega}$. Hence, for each $\epsilon > 0$ there exists $m_\epsilon$ such that for $m \geq m_\epsilon$ we have $\lVert \Pi_n^{K_m} J - J \rVert_{\pi_n}(\omega) \leq \epsilon.$ Now it is enough to remark that the projection can only get better when more random basis functions are used, since the space on which is projected gets larger, implying that for $K \leq \tilde K$, 
\begin{equation*}
\lVert \Pi_n^{K} J - J \rVert_{\pi_n}(\omega) \geq \lVert \Pi_n^{\tilde K} J - J \rVert_{\pi_n}(\omega).
\end{equation*}
Therefore, also the original sequence converges at this $\omega$, since given $\epsilon > 0$ for all $K \geq K_{m_\epsilon}$ we have 
\begin{equation*}
\lVert \Pi_n^{K} J - J \rVert_{\pi_n}(\omega) \leq \lVert \Pi_n^{K_{m_\epsilon}} J - J \rVert_{\pi_n}(\omega) \leq \epsilon.
\end{equation*}
\end{proof}

%\begin{rem}
Theorem \ref{thm: convergence of projection} holds equivalently if neural networks with more than 1 hidden layer are used. The proof is a straight forward extension of the proof given above.
%\end{rem}

\subsection{Convergence results}\label{sec:Convergence result 1}

The price of the Bermudan  approximation of the American option can be expressed with the exact algorithm as
\begin{equation*}
U_0:= \max\left(g(X_0), Q_0(X_0) \right),
\end{equation*}
the price computed with the idealized algorithm is 
\begin{equation*}
U_0^K \coloneqq \max\left(g(X_0), \tilde{Q}_0^K(X_0) \right)
\end{equation*}
and the price computed with the implementable algorithm is
\begin{equation*}
U_0^{K,m} \coloneqq \max\left(g(X_0), \frac{1}{m} \sum_{i=1}^m \hat{\tilde{Q}}_0^K(x_0, x_1^i, \dotsc, x_N^i) \right).
\end{equation*}

We provide two different convergence results with two different assumptions. The first result is based on \citep{Clement2001AnAO} which needs a technical assumption that might not be satisfied in general. The second result is based on \citep{Zanger2020},  which replaces this assumption by a stronger integrability assumption on the payoff.

\subsubsection{Convergence results based on \texorpdfstring{\citet{Clement2001AnAO}}{Cl\'ement et al.}}
\label{sec:Convergence of the Longstaff-Schwartz version 1}

Combining the following two results, convergence of $U_0^{K,m_K}$ to $U_0$ as $K \to \infty$ can be established by choosing a suitable sequence $(m_K)_{K\geq 1}$, under the assumption that $g(X_n)$ is square integrable for all $0 \leq n \leq N $.

\begin{theorem}\label{thm: conv LS 1}
The idealized price $U_0^K$ converges to the correct price $U_0$ $\tilde{\mathbb{P}}$-a.s.\ as $K \to \infty$.
\end{theorem}

\begin{theorem}\label{thm: conv LS 2}
We assume that $\mathbb{Q} [ \alpha \Pi_n^K E_n P_{n+1}^K(X_n) = g(X_n) ] = 0$ for all $0 \leq n \leq N-1$.
Then the implementable price $U_0^{K,m}$ converges almost surely to the idealized price $U_0^K$  as $m \to \infty$.
\end{theorem}

The proofs are a direct consequence of \cite{Clement2001AnAO}.

\begin{proof}[Proof of Theorems \ref{thm: conv LS 1} and \ref{thm: conv LS 2}]
The proofs are implied by the results presented in  \cite[Section 3]{Clement2001AnAO}. We only need to establish that their assumption $A_1$ is satisfied. The assumption $A_2$ is actually not needed, as explained below.

Assumption $A_1$ is that $(\phi_k(X_n))_{k \geq 1}$ is total in $L^2(\sigma(X_n))$ for every $1 \leq n \leq N-1$, which is used to show that $\lVert \Pi_n^K Q_n - Q_n \rVert_{\pi_n}$ converges to $0$. We replace this assumption by our Theorem \ref{thm: convergence of projection}, which therefore yields $\tilde{\mathbb{P}}$-almost sure convergence in the result.

Assumption $A_2$ is that for every $1 \leq n \leq N$ and every $K>0$, if $\sum_{k=0}^K \lambda_k \phi_k(X_n) = 0$ almost surely, then all $\lambda_k=0$. This assumption is actually only needed for the projection weights to be uniquely defined, such that they can be expressed by the closed-form ordinary least squares formula.
Otherwise, if this assumption is not satisfied, there exist several weight vectors $\theta$, which all define the same projection $\Phi_K \theta$ minimizing the projection objective.
By Gram--Schmidt, we can generate an orthonormal basis $(\tilde \phi_k)_{1 \leq k \leq \tilde K(K)}$ of the linear subspace of $L^2$ that is spanned by $(\phi_k)_{1 \leq k \leq K}$, with $\tilde K(K) \leq K$.
By its definition, $(\tilde \phi_k)_{1 \leq k \leq \tilde K(K)}$ satisfies assumption $A_2$ and therefore, the results of \cite[Section 3]{Clement2001AnAO} can be applied. 
Finally, we note that the projections are the same, no matter whether $(\tilde \phi_k)_{1 \leq k \leq \tilde K}$ or $(\phi_k)_{1 \leq k \leq K}$ are used to describe the space that is spanned. 
We are interested in the convergence of the price. Considering the definition \eqref{equ:implementable approx cont value}, we see that the price depends only on the projection but not on the used weights. Therefore, we can  conclude that the same statements hold with our originally defined random basis functions $(\phi_k)_{1 \leq k \leq K}$.
\end{proof}

%\begin{remark}
The technical assumption that  $\mathbb{Q} [ \alpha \Pi_n^K E_n P_{n+1}^K(X_n) = g(X_n) ] = 0$ for all $0 \leq n \leq N-1$ of the result of \cite{Clement2001AnAO} that shows up in Theorem~\ref{thm: conv LS 2}  is not always satisfied. In particular, it is easy to construct examples of finite probability spaces, where this is not the case. Indeed, consider  the easiest possible case of  probability space which is a singelton, with a (deterministic) constant stock price without discounting, then $\mathbb{Q} [ \Pi_n^K E_n P_{n+1}^K(X_n) = g(X_n) ] = 1$.
Therefore, in the next section, we  provide a different proof based on the work of \cite{Zanger2020}, which replaces this assumption by a slightly stronger integrability assumption on the payoff process.
%\end{remark}

\subsubsection{Convergence results based on \texorpdfstring{\cite{Zanger2020}}{Zanger}}
\label{sec:Convergence of the Longstaff-Schwartz version 2}

After the work of \cite{Clement2001AnAO}, improved theoretical guarantees to the original Least Squares Monte Carlo algorithm (LSM)  have been proposed, such as \citep{Stentoft2004, Egloff2005monte, Gobet2005regression}. An important improvement of the  convergence results is done in \citep{Zanger2009convergence, Zanger2013QuantitativeError,Zanger2018convergence, Zanger2020}. In particular,  \cite{Zanger2009convergence} raised the issue that \cite{Clement2001AnAO} has additional restrictions on the law of the underlying Markov process such as the assumption in Theorem~\ref{thm: conv LS 2} mentioned above.
\cite{Zanger2009convergence}  proposed a generalized LSM algorithm and provides a proof of convergence in probability \citep[Theorem 5.1]{Zanger2009convergence}. In this theorem, the condition of \cite{Clement2001AnAO} is not needed, but instead  the payoff needs to be  bounded almost surely \citep[Definition 5.1 and 5.2]{Zanger2009convergence}. \cite{Zanger2013QuantitativeError} provides error estimates (convergence rates), even when the underlying process and payoff process are not necessarily in $L^{\infty}$. Later, \cite{Zanger2018convergence} provides a convergence result \citep[Corollary 5.5]{Zanger2018convergence}  without the assumption in Theorem~\ref{thm: conv LS 2} of \cite{Clement2001AnAO}, but with a bounded payoff process. However, this time,  almost sure convergence is shown instead of convergence in probability. Finally, in the last paper \cite{Zanger2020},  the assumption of having a bounded payoff process is replaced by a condition on its  moments \cite[Corollary 1]{Zanger2020}. We use this last result to prove our second convergence theorem.
To state this result we  define the truncation operator $\mathcal{T}_\lambda$ for truncation level $\lambda > 0$ acting on any real-valued function $f$ by 
\begin{equation*}
\mathcal{T}_\lambda f (x) = \begin{cases}
f(x), & \text{if } |f(x)| \leq \lambda, \\
\lambda \operatorname{sign}(f(x)), & \text{otherwise}.
\end{cases}
\end{equation*}

\begin{theorem}
Assume that there exists some $2 < p \leq \infty$ such that $$M_p \coloneqq \max_{1 \leq n \leq N}\lVert g(X_n) \rVert_{L^p}^p < \infty$$
and that all payoffs are non-negative.
Moreover, assume that we use the truncated versions of the payoffs $g(X_n)$ in Algorithm~\ref{algo:1} as well as the truncated versions of the randomized neural networks, with truncation level $1 \leq \lambda < \infty$.
Then 
$$ \mathbb{E} \left[ \left| U_0^{K,m} - U_0 \right| \right] \xrightarrow[K,m \to \infty]{\tilde{ \mathbb{P}}-a.s. } 0,$$
when choosing $\lambda = m^{1/8}$.
\end{theorem}

The proof is a direct consequence of \cite[Corollary 1]{Zanger2020}.
\begin{proof}
Let us fix the number of paths $m$ and the number of random basis functions $K$. Then  \cite[Corollary 1]{Zanger2020} implies that
\begin{multline*}
\mathbb{E} \left[ \left| U_0^{K,m} - U_0 \right| \right] \leq 6^N\left(\frac{C\lambda^2 \left( \sqrt{\nu c_0} \log^{\frac{1}{2}}(m) + \log^{\frac{1}{2}}(C_0)\right)}{\sqrt{m}} \right. \\ 
+ \left. 4\sqrt{\varepsilon} + \max_{n=1, \dots, N-1} \left( \inf_{f\in \mathcal{B}_n^{K,\lambda}} \| f - Q_n \|_{\pi_n}\right) + \left( \frac{8M_p\lambda^{(2-p)}}{p-2} \right)^{1/2} \right),
\end{multline*}
where $C_0 = C(c_0\nu+1)^4(C\lambda^4)^{2\nu(1+c_0)}$, $c_0=2(N+1) \log_2(e(N+1))$ and $C$ is a numerical constant with $1\leq C <\infty$, and $\varepsilon\geq 0$ as defined in \citep[ Equation 13]{Zanger2020}.
Here, $\nu$ is the Vapnik–Chervonenkis (VC) dimension of the set of randomized neural networks, which is  finite  according to \cite[Remark 8]{Zanger2020}. 
For each exercise time $1 \leq n \leq N-1$ the set $\mathcal{B}_n^{K,\lambda}$ is defined to be the set of all $\lambda$-truncated randomized neural networks using the first $K$ random basis functions (i.e. any truncated version of a linear combinations of the basis functions $(\phi_k)_{1\leq k \leq K}$). In particular
\begin{equation*}
\mathcal{B}_n^{K,\lambda} = \{ \mathcal{T}_{\lambda} f | f \in \operatorname{span}\{\phi_1, \dotsc, \phi_K\}\},
\end{equation*}
where $\mathcal{T}_\lambda$ is the operator truncating a function at $\lambda$.
Note that for any function $f$, we have that
\begin{equation*}
\| (\mathcal{T}_\lambda f - Q_n) 1_{\{ |Q_n| < \lambda \} } \|_{\pi_n} \leq \| (f - Q_n)1_{\{ |Q_n| < \lambda \} }  \|_{\pi_n}\,.
\end{equation*}
Therefore, 
\begin{multline*}
\inf_{f\in \mathcal{B}_n^{K,\lambda}} \| (f - Q_n) 1_{\{ |Q_n| < \lambda \} } \|_{\pi_n} 
\leq \inf_{f\in \operatorname{span}\{\phi_1, \dotsc, \phi_K\}}\| (f - Q_n)1_{\{ |Q_n| < \lambda \} }  \|_{\pi_n}\\
\leq \inf_{f\in \operatorname{span}\{\phi_1, \dotsc, \phi_K\}}\| f - Q_n  \|_{\pi_n}
= \| \Pi_n^K Q_n - Q_n  \|_{\pi_n}\,.
\end{multline*}
Hence, we can now bound the approximation error with truncated randomized neural networks by
\begin{multline*}
\inf_{f\in \mathcal{B}_n^{K,\lambda}} \| f - Q_n \|_{\pi_n} 
	\leq \inf_{f\in \mathcal{B}_n^{K,\lambda}} \left( \| (f - Q_n) 1_{\{ |Q_n| < \lambda \} } \|_{\pi_n} + \| (f - Q_n) 1_{\{ |Q_n| \geq \lambda \} } \|_{\pi_n} \right)\\
	\leq \inf_{f\in \mathcal{B}_n^{K,\lambda}} \| (f - Q_n) 1_{\{ |Q_n| < \lambda \} } \|_{\pi_n} +  \sup_{f\in \mathcal{B}_n^{K,\lambda}} \| (f - Q_n) 1_{\{ |Q_n| \geq \lambda \} } \|_{\pi_n} \\
	\leq \| \Pi_n^K Q_n - Q_n  \|_{\pi_n} + 2 \| Q_n 1_{\{ |Q_n| \geq \lambda \} } \|_{\pi_n} ,
\end{multline*}
where in the last inequality we used  that functions in $\mathcal{B}_n^{K,\lambda}$ are truncated at $\lambda$ implying that  they are bounded by $|Q_n|$ on the set $\{ |Q_n| \geq \lambda \}$.
Moreover, we can choose $\varepsilon = 1/m$, replace $\lambda = m^{1/8}$ and simplify all expressions by using one common constant $\tilde C$ to rewrite
\begin{multline*}
\mathbb{E} \left[ \left| U_0^{K,m} - U_0 \right| \right] \leq \tilde C  \left(
\frac{\log^{\frac{1}{2}}(m)}{m^{1/4}} + \frac{1}{\sqrt{m}} \right. \\ 
+\left.  \max_{n=1, \dots, N-1}  \left(  \| \Pi_n^K Q_n - Q_n  \|_{\pi_n} + 2 \| Q_n 1_{\{ |Q_n| \geq m^{1/8} \} } \|_{\pi_n} \right) + m^{\frac{2-p}{16}}\right).
\end{multline*}
Now it suffices to note that the terms $\| \Pi_n^K Q_n - Q_n  \|_{\pi_n}$ converge to $0$ as $K \to \infty$ by Theorem~\ref{thm: convergence of projection}, the terms $\| Q_n 1_{\{ |Q_n| \geq m^{1/8} \} } \|_{\pi_n}$ converge to $0$ as $m \to \infty$ by dominated convergence and the remaining terms trivially converge to $0$ as $m \to \infty$.
\end{proof}

\section{Convergence of the randomized fitted Q-iteration (RFQI)}
\label{sec:Convergence of reservoir optimal stopping via reinforcement learning}
Similarly as in Appendix~\ref{sec:Convergence of the Longstaff-Schwartz version}, we first introduce some additional technical notation needed for the proofs. Then, we describe the  steps from the theoretical idea of RFQI  to its implementable version that was presented in  Section~\ref{sec:optimal stopping via reinforcement learning}. 
In contrast to Appendix~\ref{sec:Convergence of the Longstaff-Schwartz version}, the algorithms described here are applied simultaneously for all times.
Again, the proof is a direct consequence of \citep{Tsitsiklis2001Regression} and Theorem~\ref{thm: convergence of projection}, but is given in detail for completeness.

\subsection{Definitions}\label{sec:Definitions 2}
In Section 6, \cite{Tsitsiklis2001Regression} introduced a reinforcement learning version of their optimal stopping algorithm, where a stopping function is learned that generalizes over time. In particular, instead of learning a different function for each time step, a single function that gets the time as input is learned with an iterative scheme.  
In accordance with this, the random basis functions are redefined such that they also take time as input
\begin{equation*}
\begin{split}
\phi_k: &\mathbb{R}^d \times \{ 0 , \dotsc, N-1 \} \to \mathbb{R}, \\
&(x,n) \mapsto \phi_k(x,n)\coloneqq \sigma(\alpha_k^\top (x,n)^\top + \beta_k),
\end{split}
\end{equation*}
with $\alpha_k \in \mathbb{R}^{d+1}$ and $\beta_k \in \mathbb{R}$. For $0 \leq n \leq N-1$ let $\Phi_{K,n}$ be defined similarly to before as
\begin{equation*}
(\Phi_{K,n} \theta)(x) \coloneqq \theta^\top \phi(x,n) \coloneqq \sum_{k=1}^K \theta_k \phi_k(x,n),
\end{equation*}
for $\theta \in \mathbb{R}^K$ and $x \in \mathbb{R}^d$. Moreover, let $\Phi_k \coloneqq (\Phi_{K,0}, \dotsc, \Phi_{K,N-1})$, such that 
\begin{equation*}
\Phi_k \theta \coloneqq (\Phi_{K,0} \theta, \dotsc, \Phi_{K,N-1} \theta).
\end{equation*} 
In the following, we consider the product space $(L^2)^N \coloneqq L^2(\pi_0)\times \dotsb \times L^2(\pi_{N-1})$, which is the space on which the functions for all time steps can be defined concurrently. For $J = (J_0, \dotsc, J_{N-1}) \in (L^2)^N$ we define the norm
\begin{equation*}
\lVert J \rVert_\pi \coloneqq \frac{1}{N} \sum_{n=0}^{N-1} \lVert J_n \rVert_{\pi_n},
\end{equation*}
where $\lVert \cdot \rVert_{\pi_n}$ is as defined in Appendix~\ref{sec:Convergence of the Longstaff-Schwartz version}. Let us define the projection operator $\Pi^K$ as
\begin{equation*}
(\Pi^K J) \coloneqq \arg \min_{\Phi_K \theta } \lVert \Phi_K \theta - J \rVert_{\pi},
\end{equation*}
for $J = (J_0, \dotsc, J_{N-1}) \in (L^2)^N$. Finally, we define the operator
\begin{equation}
\label{equ:operator H}
H :  (L^2)^N \to (L^2)^N, \quad
\begin{pmatrix}
J_0 \\ \vdots \\ J_{N-2} \\ J_{N-1}
\end{pmatrix}
\mapsto 
\begin{pmatrix}
\alpha E_0 \max(g, J_1) \\ \vdots \\ \alpha E_{N-2} \max(g, J_{N-1}) \\ \alpha E_{N-1} g
\end{pmatrix},
\end{equation}
where $E_n$ and $g$ are as defined previously.

\subsection{Theoretical description of the algorithm}\label{sec:Abstract description of the reservoir algorithm 2}
Based on the definitions in Appendix~\ref{sec:Theoretical Description of OurAlgo}, we first introduce the \emph{exact algorithm} and then give the two-step approximation with the \emph{idealized} and \emph{implementable algorithm}.

\subsubsection{Exact algorithm}
Let $Q_n$ be as defined in \eqref{equ:exact cont value}, then $Q \coloneqq (Q_0, \dotsc, Q_{N-1})$ satisfies $Q = HQ$ by definition. In particular, $Q$ is a fixed point of $H$.
It was shown in \cite[Section 6]{Tsitsiklis2001Regression} that $H$ is a contraction with respect to the norm $\lVert \cdot \rVert_\pi$ with contraction factor $\alpha$.
Hence, the Banach fixed point theorem implies that there exists a unique fixed point, which therefore has to be $Q$, and that for any starting element $J^0 \in (L^2)^N$, $J^i$ converges to $Q$ in $(L^2)^N$ as $i \to \infty$, where $J^{i+1} \coloneqq H J^i$. This yields a way to find the \emph{exact algorithm} $Q$ iteratively.

\subsubsection{Idealized algorithm}
The combined operator $\Pi^K H$ is a contraction on the space $\Pi^K (L^2)^N$, since the projection operator is a non-expansion as outlined in \cite[Section 6]{Tsitsiklis2001Regression}. 
The \emph{idealized algorithm} is then defined as the unique fixed point $\tilde{Q}^K$ of $\Pi^K H$, which can again be found by iteratively applying this operator to an arbitrary starting point.
Since any element in $\Pi^K (L^2)^N$ is given as $\Phi_K \theta$ for some weight vector $\theta \in \mathbb{R}^K$, this iteration can equivalently be given as iteration on the  weight vectors. 
To do this, let us assume without loss of generality that $(\phi_k)_{1\leq k \leq K}$ are linearly independent (if not, see the strategy in Proof of Theorem \ref{thm: conv LS 1} and \ref{thm: conv LS 2}).
Then, given some starting weight vector $\theta_K^{0}$, the iterative application of $\Pi^K H$ defines the weight vectors 
\begin{multline*}
\theta_K^{i+1} \coloneqq \alpha \left( \mathbb{E}\left[\sum_{n=0}^{N-1} \phi_{1:K}^\top (X_n, n)\phi_{1:K}(X_n, n) \right]  \right)^{-1} \\
\cdot \mathbb{E}\left[ \sum_{n=0}^{N-1} \phi_{1:K}^\top(X_n, n)  
\cdot  \max\left(g(X_{n+1}), (\Phi_{K,n+1} \theta_K^{i})(X_{n+1})\right) \right],
\end{multline*}
where $\phi_{1:K} = (\phi_{1}, \dotsc, \phi_{K})$. This closed-form solution is exactly the ordinary least squares (OLS) formula and this result was shown in \cite[Section 6]{Tsitsiklis2001Regression}.

\subsubsection{Implementable algorithm}
An \emph{implementable version} of this iteration is defined by the Monte Carlo approximation of the weight vectors. In particular, we assume that $m$ realizations $(x_0^j, \dotsc, x_N^j)_{1\leq j \leq m}$ of $X$ are sampled and fixed for all iterations. Then for $\hat{\theta}_{K,m}^{0} = \theta_K^{0}$ we iteratively define
\begin{multline*}
\hat{\theta}^{i+1}_{K,m} \coloneqq \alpha \left( \sum_{j=1}^m \sum_{n=0}^{N-1} \phi_{1:K}^\top (x_n^j, n)\phi_{1:K}(x_n^j, n)  \right)^{-1} \\
\cdot \sum_{j=1}^m \sum_{n=0}^{N-1} \phi_{1:K}^\top(x_n^j, n) 
\cdot \max\left(g(x_{n+1}^j), (\Phi_{K,n+1} \hat\theta^{i}_{K,m})(x_{n+1}^j)\right) ,
\end{multline*}
which in turn defines $\hat{Q}^{K,m,i} \coloneqq \Phi_K \hat\theta^{i}_{K,m}$.
As explained in \cite[Section 6]{Tsitsiklis2001Regression}, this implementable iteration can equivalently be described as iteratively applying the operator $\widehat{\Pi^K H}$. Here $\widehat{\Pi^K H}$ is identical to $\Pi^K H$, but with the measures $\pi_n$ replaced by the empirical measures $\hat{\pi}_n$ arising from the sampled trajectories $(x_0^j, \dotsc, x_N^j)_{1 \leq j \leq m}$. Hence,  $\widehat{\Pi^K H}$ is also a contraction and Banach's fixed point theorem implies convergence to the unique fixed point 
\begin{equation*}
\hat{Q}^{K,m,i} \xrightarrow{i \to \infty} \hat{Q}^{K,m} =: \Phi_K \hat\theta^{\star}_{K,m}.
\end{equation*}
%which is the \textbf{implementable algorithm}.
We note that this also implies that $\hat\theta^{i}_{K,m} \xrightarrow{i \to \infty} \hat\theta^{\star}_{K,m}$.

\subsection{Convergence result}
In the following, we show that prices of Bermudan options computed with the two approximation steps of the exact algorithm converge to the correct price, as $K, m \to \infty$. 
The prices are defined similarly as in Appendix~\ref{sec:Convergence result 1}. 
Hence, it is enough to show that $\hat{Q}^{K,m_i,i}$ converges to $\tilde{Q}^K$ as $ i\to\infty$ and that  $\tilde{Q}^K$ converges to $Q$ as $ K \to\infty$.

\begin{theorem}
$\tilde{Q}^K$ converges $\tilde{\mathbb{P}}$-a.s. to $Q$ as $ K \to\infty$, i.e.
\begin{equation*}
\lVert \tilde{Q}^K - Q  \rVert_{\pi} \xrightarrow[K \to \infty]{\tilde{\mathbb{P}}-a.s.} 0.
\end{equation*}
\end{theorem}

\begin{proof}
First, let us recall \cite[Theorem 3]{Tsitsiklis2001Regression}, which states that for $0 <\kappa < 1$  the contraction factor of $\Pi^K H$, we have
\begin{equation*}
\lVert \tilde{Q}^K - Q  \rVert_{\pi} \leq \frac{1}{\sqrt{1-\kappa^2}} \lVert \Pi^K Q - Q \rVert_\pi.
\end{equation*}
Now remark that since $\Pi^K$ is a non-expansion and $H$ a contraction with factor $\alpha$, we have $\kappa \leq \alpha < 1$. Therefore, for every $K$ we have
\begin{equation}\label{equ:proof thm conv RL 1}
\lVert \tilde{Q}^K - Q  \rVert_{\pi} \leq \frac{1}{\sqrt{1-\alpha^2}} \lVert \Pi^K Q - Q \rVert_\pi.
\end{equation}
Finally, we remark that Theorem \ref{thm: convergence of projection} holds equivalently for the norm $\lVert \cdot \rVert_\pi$, since the universal approximation theorem can equivalently be applied to the functions with the combined input $(x,n)$. Hence, the right hand side of \eqref{equ:proof thm conv RL 1} converges to $0$ $\tilde{\mathbb{P}}$-a.s. as $K \to \infty$. 
\end{proof}

We recall that the weight vectors $\hat{\theta}_{K,m}^i$ are random variables since they depend on the $m$ sampled trajectories of $X$. 

\begin{lemma}\label{lem:2}
For any fixed $i \in \mathbb{N}$ we have that  $\hat\theta^{i}_{K,m}$ converges  to $\theta^{i}_{K}$ $\mathbb{Q}$-a.s. as $m \to \infty$.
\end{lemma}

\begin{proof}
The proof follows the proof of \cite[Theorem 2]{Tsitsiklis2001Regression}.
We introduce the intermediate weight as
\begin{multline*}
\tilde{\theta}^{i}_{K,m} \coloneqq \alpha \left( \sum_{j=1}^m \sum_{n=0}^{N-1} \phi_{1:K}^\top (x_n^j, n)\phi_{1:K}(x_n^j, n)  \right)^{-1} \\
\cdot \sum_{j=1}^m \sum_{n=0}^{N-1} \phi_{1:K}^\top(x_n^j, n) 
\cdot \max\left(g(x_{n+1}^j), (\Phi_{K,n+1} \theta^{i-1}_{K})(x_{n+1}^j)\right).
\end{multline*}
Then it is clear that $\tilde{\theta}^{i}_{K,m}$ converges to ${\theta}^{i}_{K}$ $\mathbb{Q}$-a.s. as $m \to \infty$, by the strong law of large numbers. Hence, 
$\delta_i(m) \coloneqq \lvert \tilde{\theta}^{i}_{K,m} - {\theta}^{i}_{K} \rvert_2$
converges to $0$ $\mathbb{Q}$-a.s.
Moreover, for suitably chosen random variables $A_i(m)$ that remain bounded as $m \to \infty$, we have 
\begin{equation*}
\hat\theta^{i}_{K,m} - \tilde{\theta}^{i}_{K,m} = A_i(m) \lvert \hat{\theta}^{i-1}_{K,m} - {\theta}^{i-1}_{K} \rvert_2.
\end{equation*}
Therefore we have by the triangle inequality
\begin{equation*}
\lvert \hat{\theta}^{i}_{K,m} - {\theta}^{i}_{K} \rvert_2 \leq \delta_i(m) + A_i(m) \lvert \hat{\theta}^{i-1}_{K,m} - {\theta}^{i-1}_{K} \rvert_2.
\end{equation*}
Since (by our choice) we start with the same weight vector $\hat{\theta}_{K,m}^0 = \theta_K^0$, we can conclude by induction that 
\begin{equation*}
\lvert \hat{\theta}^{i}_{K,m} - {\theta}^{i}_{K} \rvert_2 \xrightarrow[m \to \infty]{\mathbb{Q}-a.s.} 0.
\end{equation*}
However, we remark that this proof only works as long as $i$ is fixed, but not in the limit $i \to \infty$, because the inductive steps would lead to an infinite sum. 
\end{proof}

\begin{theorem}
Let $K \in \mathbb{N}$ be fixed. Then there exists a random sequence $(m_i)_{i\geq 0}$ such that $\hat{Q}^{K, m_i, i}$ converges $\mathbb{Q}$-a.s. to $\tilde{Q}^K$ as $i \to \infty$, i.e.
\begin{equation*}
\lVert \hat{Q}^{K, m_i, i} -  \tilde{Q}^K \rVert_\pi \xrightarrow[i \to \infty]{\mathbb{Q}-a.s.} 0.
\end{equation*}
\end{theorem}

\begin{proof}
Let us define $\theta_K^{\star} \in \mathbb{R}^K$ to be the weight vector of the unique fixed point $\tilde{Q}^K$ of $\Pi^K H$, i.e. $\tilde{Q}^K = \Phi_K \theta_K^{\star}$. From Banach's fixed point theorem we know that $\lvert \theta_K^{i} - \theta_K^{\star}  \rvert_2 \to 0$ as $i \to \infty$.\\
With Lemma \ref{lem:2} we know that for every $i \in \mathbb{N}$ there exists $\Omega_i \subset \Omega$ with $\mathbb{Q}(\Omega_i) = 1$ such that $\hat\theta^{i}_{K,m}(\omega)$ converges to $\theta^{i}_{K}$ for all $\omega \in \Omega_i$. 
Let $\Omega_\infty \coloneqq \cap_{i=1}^\infty \Omega_i$ be the set on which this convergence holds for all $i \in \mathbb{N}$, then $\mathbb{Q}(\Omega_\infty) = 1$. Fix $\omega \in \Omega_\infty$.
Now let us choose $m_0 = 0$ and for every $i > 0$, $m_i > m_{i-1}$ such that $\lvert \hat\theta^{i}_{K,m_i}(\omega) -  \theta^{i}_{K} \rvert_2 \leq 1/i$.
Therefore, we obtain that
\begin{equation*}
\begin{split}
\lvert \hat\theta^{i}_{K,m_i}(\omega) - \theta_K^{\star} \rvert_2 
	\leq \lvert \hat\theta^{i}_{K,m_i}(\omega) -  \theta^{i}_{K} \rvert_2 + \lvert  \theta^{i}_{K} - \theta_K^{\star}  \rvert_2 
	\leq \frac{1}{i} + \lvert  \theta^{i}_{K} - \theta_K^{\star}  \rvert_2,
\end{split}
\end{equation*}
which converges to $0$ when $i$ tends to infinity.
\end{proof}

\section{Convergence of the  randomized recurrent least squares Monte Carlo (RRLSM)}
\label{sec:Convergence of the recurrent Longstaff-Schwartz version}
In this section, we extend the results of Appendix~\ref{sec:Convergence of the Longstaff-Schwartz version} to the non-Markovian setting, where we assume that the path up to the current time is a Markov process. 
In particular, given a discrete time stochastic process $X = (X_0, \dotsc, X_N) $ as before, we assume that its extension $Z = (Z_0, \dotsc, Z_N)$ with $Z_n = (X_n, X_{n-1}, \dotsc, X_0, 0, \dotsc, 0 )$ taking values in $\mathbb{R}^{N+1 \times d}$ for all $0 \leq n \leq N$ is a Markov process.
Hence, all results of Appendix~\ref{sec:Convergence of the Longstaff-Schwartz version} hold similarly up to replacing $X$ by $Z$ and they also hold for payoff functions that depend on the entire path of $X$ up to the current time. In particular, this immediately implies that RLSM  with the path input $Z$ approximates the correct price of the Bermudan option arbitrarily well as $K \to \infty$.
Therefore, it is only left to show that an equivalent result to Theorem~\ref{thm: convergence of projection} holds for our randomized recurrent neural network \eqref{equ:Randomized recurrent NN}, which takes $X$ as input instead of $Z$, but makes use of a latent variable in which information about the past is stored. 

Fix some $1 \leq n \leq N-1$ and let $\pi_n$ now be the distribution of $Z_n$ under $\mathbb{Q}$. Moreover, let the basis functions $\phi^n = (\phi^n_k)_{ k \geq 1 }$ be now given by the $n$-th latent variable $h_n$ of  \eqref{equ:Randomized recurrent NN}. 
In particular, we define $\phi^n_k$ as the function mapping $z_n = (x_n, x_{n-1}, \dotsc, x_0, 0, \dotsc, 0) $ to the $k$-th coordinate of the recursively defined vector
\begin{equation}\label{equ:hn}
h_{n}  =     \sigma(A_x x_n +A_h h_{n-1} + b),
\end{equation}
where $h_{-1}=0$.
By abuse of notation, for growing $k$ we let the matrices grow by adding new rows of random elements to $b, A_x$ and $A_h$ and filling up the new columns of previous rows of $A_h$ with zeros. Like this, $\phi^n_k$ is well defined for all $k \geq 1$.
The operator $\Pi_n^K$ is defined similarly as before, but with this new set of basis functions, defined on the set of $\pi_n$-integrable functions $J$.
Then we have to show that the following result is true, so that the assumptions for Theorem \ref{thm: conv LS 1} and \ref{thm: conv LS 2}  are satisfied. The remainder of their proof works as before.

\begin{proposition}\label{prop: convergence of projection RNN case}
If the activation function $\sigma$ is invertible then for all $0 \leq n \leq N-1$, 
\begin{equation}\label{equ:convergence RNN}
\lVert \Pi_n^K Q_n - Q_n \rVert_{\pi_n} \xrightarrow[K \to \infty]{\tilde{\mathbb{P}}\text{-a.s.}} 0.
\end{equation}
\end{proposition}

Before we start with the proof, we remark that standard results for the approximation of dynamical systems with RNNs \citep{Schafer2006RNNUniversal} and reservoir computing systems \citep{GononOrtega2020} do not apply here, since the dynamical system to approximate $Q = (Q_0, \dotsc , Q_{N-1})$ is not time-invariant (in the language of \cite{GononOrtega2020}).

\begin{proof}
Firstly, we note that it is enough to show that for any $\epsilon >0$ there exists some size $K \in \mathbb{N}$ and weight matrices $b, A_x, A_h$ such that the corresponding neural network approximation $\tilde \Pi^n_K Q_n$ satisfies $\lVert \tilde \Pi_n^K Q_n - Q_n \rVert_{\pi_n} < \epsilon$ for all $0 \leq n \leq N-1$. Indeed, if this is true, the convergence \eqref{equ:convergence RNN} follows by the same arguments as in Lemma~\ref{lem: for thm convergence of projection} and Theorem~\ref{thm: convergence of projection}. \\
Secondly, we note that it is enough to show the  statement above for any fixed $n$ separately, i.e. that for each $0 \leq n \leq N-1$ and $\epsilon > 0$ there exist $K^n \in \mathbb{N}$ and weight matrices $b^n, A_x^n, A_h^n$ such that the corresponding neural network approximation $\tilde \Pi^n_K Q_n$ satisfies $\lVert \tilde \Pi_n^K Q_n - Q_n \rVert_{\pi_n} < \epsilon$. Indeed, if this is true, the stronger statement follows immediately by setting 
\begin{equation*}
A_x = \begin{pmatrix}
A_x^0 \\
\vdots \\
A_x^{N-1}
\end{pmatrix},
\quad A_h = \begin{pmatrix}
A_h^0 \\
& \ddots \\
& & A_h^{N-1}
\end{pmatrix}
\text{ and }
b = \begin{pmatrix}
b^0 \\
\vdots \\
b^{N-1}
\end{pmatrix}.
\end{equation*}
Hence, let us fix some $\epsilon > 0$ and $0 \leq n \leq N-1$ and let us assume that $d=1$ for simplicity of notation, while the extension to $d>1$ is immediate. We know from the universal approximation theorem \cite[Theorem 1]{hornik1991approximation} that there exists some neural network $f$ such that $\lVert  f - Q_n  \rVert_{\pi_n} < \epsilon$. The difference between the approximation $\tilde \Pi_K^n$ and $f$ is that $\tilde \Pi_K^n$ gets a recurrent input, while $f$ gets the entire path as input. However, since $n$ is fixed and finite, we can simply accumulate the same path information in $h_n$ by setting $\hat b=0$, $\hat A_x = (1, 0, \dotsc, 0)^\top \in \mathbb{R}^{n}$ and $$\hat A_h= \begin{pmatrix}
0 & \hdots & & &  0 \\
1 & 0 & \hdots & & 0 \\
0 & \ddots  & & & \vdots \\
\vdots & & & &  \\
0 & \hdots & 0 & 1 & 0 \\
\end{pmatrix} \in \mathbb{R}^{n \times n}.
$$
Indeed, with this choice we have $\hat h_{n-1} = ( \sigma(x_{n-1}) , \sigma(\sigma(x_{n-2})), \dotsc, \sigma^{(n)}(x_0) )^\top$ according to \eqref{equ:hn}.
It remains to show that the input $z_n$ to $f$ can be replaced by $(x_n, \hat h_{n-1})$.
For this, let us define the function 
$$\varphi:  (x_n , \dotsc, x_0, 0, \dotsc, 0) \mapsto (x_n, \sigma(x_{n-1}) ,  \dotsc, \sigma^{(n)}(x_0)).$$
Under the assumption that $\sigma$ is invertible also $\varphi$ is  and  there exists a function $\tilde Q_n$  such that 
$\tilde Q_n \circ \varphi  = Q_n$.
Since $Q_n$ is integrable with respect to $\pi_n$, the change of variables formula implies that $\tilde Q_n$ is  integrable with respect to $\varphi^{-1} \circ \pi_n$ and $\mathbb{E}^{(\varphi^{-1} \circ \pi_n)}[\tilde Q_n ]  = \mathbb{E}^{\pi_n}[ \tilde Q_n \circ \varphi] = \mathbb{E}^{\pi_n}[ Q_n]$. 
Therefore,  there exists a neural network $\tilde f = \tilde \beta^\top \, \sigma(\tilde A \cdot + \tilde b)$ such that 
$$\lVert   \tilde f \circ \varphi -  Q_n  \rVert_{\pi_n} = \lVert  ( \tilde f - \tilde Q_n ) \circ \varphi \rVert_{\pi_n} = \lVert  \tilde f - \tilde Q_n  \rVert_{\varphi^{-1} \circ \pi_n} < \epsilon.$$ 
By extending $\hat b, \hat A_x, \hat A_h$ to 
$$
b = \begin{pmatrix}
\hat b \\
\tilde{b}
\end{pmatrix} ,
\quad
A_x = \begin{pmatrix}
\hat A_x \\
\tilde A_1
\end{pmatrix},
\quad
A_h =  \begin{pmatrix}
\hat A_h  & 0 \\
\tilde A_{2:n+1} & 0
\end{pmatrix}
,
$$
where $\tilde A  = (\tilde A_1, \tilde A_{2:n+1})$, we get
$$h_n = \begin{pmatrix}
\hat h_n \\
\tilde{h}_n
\end{pmatrix} = \begin{pmatrix}
\sigma(\hat A_x x_n + \hat A_h \hat h_{n-1} + \hat b) \\
\sigma (\tilde A \varphi(z_n) + \tilde b ) 
\end{pmatrix}, $$
where $\tilde \beta^\top \tilde{h}_n = \tilde f(\varphi(z_n)) $.
Therefore, we can conclude the proof, since the corresponding approximation $\tilde \Pi_K^n$ satisfies $\lVert  \tilde \Pi_K^n Q_n  -  Q_n  \rVert_{\pi_n} \leq  \lVert   \tilde f \circ \varphi -  Q_n  \rVert_{\pi_n} \leq \epsilon$.
\end{proof}

\begin{remark}
The idea of the proof is to use the recurrent structure only to recover the path-wise input $z_n$ for which the standard  feed-forward neural network approximation results can be used. This is clearly less efficient than using the path-wise input directly.
However, in practice, the recurrent neural network approach is usually  more efficient than the path-wise approach, finding  better ways to  store and process the past information than the one given in the proof. This is in line with our empirical findings.
\end{remark}

\begin{landscape}
\subsection{Stopping of a fractional Brownian motion -- table}\label{sec:Stopping of Fractional Brownian Motion Table}
The  results shown in the plots of Section \ref{sec:The Non-Markovian Case -- Optimally Stopping Fractional Brownian Motions} are given in Table \ref{hurst_table_all}.

\begin{table*}[!h]
\center
\scalebox{0.7}{
\begin{tabular}{|c|r r r r r r r r|r r r r r r r r|}
\toprule
{} & \multicolumn{8}{c |}{price} & \multicolumn{8}{c |}{duration} \\
H & DOS & pathDOS & RLSM & RRLSM & FQI & RFQI & RRFQI & pathRFQI & DOS & pathDOS & RLSM & RRLSM & FQI & RFQI & RRFQI & pathRFQI\\
\midrule
0.01  &  0.85 (0.02) &  1.48 (0.01) &  0.84 (0.01) &  1.45 (0.01) &  0.79 (0.01) &  0.78 (0.02) &  0.85 (0.07) &  1.09 (0.08) &    1m15s &   2m59s &   0s &    1s &   9s &   5s &   18s &      18s \\
0.05  &  0.67 (0.02) &  1.24 (0.01) &  0.65 (0.01) &  1.24 (0.01) &  0.68 (0.01) &  0.67 (0.02) &  0.71 (0.04) &  0.99 (0.07) &    1m15s &   3m 1s &   0s &    1s &   9s &   4s &   19s &      19s \\
0.1   &  0.50 (0.02) &  0.99 (0.01) &  0.49 (0.01) &  1.02 (0.01) &  0.57 (0.01) &  0.55 (0.01) &  0.56 (0.05) &  0.83 (0.03) &    1m12s &   2m58s &   0s &    1s &  10s &   4s &   19s &      20s \\
0.15  &  0.37 (0.02) &  0.77 (0.01) &  0.38 (0.01) &  0.82 (0.01) &  0.47 (0.02) &  0.45 (0.02) &  0.47 (0.05) &  0.65 (0.03) &    1m13s &   2m59s &   0s &    1s &   9s &   4s &   19s &      18s \\
0.2   &  0.28 (0.01) &  0.60 (0.01) &  0.31 (0.01) &  0.64 (0.01) &  0.38 (0.01) &  0.35 (0.09) &  0.31 (0.02) &  0.53 (0.02) &    1m15s &   2m58s &   1s &    1s &   9s &   4s &   19s &      17s \\
0.25  &  0.23 (0.01) &  0.44 (0.01) &  0.25 (0.01) &  0.49 (0.01) &  0.29 (0.01) &  0.26 (0.05) &  0.26 (0.04) &  0.39 (0.02) &    1m14s &   2m58s &   1s &    1s &   9s &   4s &   18s &      19s \\
0.3   &  0.18 (0.01) &  0.30 (0.01) &  0.20 (0.01) &  0.36 (0.01) &  0.21 (0.01) &  0.17 (0.01) &  0.15 (0.01) &  0.27 (0.01) &    1m13s &   2m57s &   1s &    1s &   9s &   4s &   18s &      18s \\
0.35  &  0.13 (0.01) &  0.19 (0.01) &  0.15 (0.01) &  0.25 (0.01) &  0.14 (0.01) &  0.13 (0.02) &  0.12 (0.03) &  0.17 (0.01) &    1m15s &   2m57s &   1s &    1s &   9s &   4s &   18s &      19s \\
0.4   &  0.08 (0.01) &  0.10 (0.01) &  0.10 (0.01) &  0.14 (0.01) &  0.09 (0.01) &  0.06 (0.01) &  0.06 (0.01) &  0.10 (0.02) &    1m15s &   2m59s &   1s &    1s &   9s &   4s &   18s &      19s \\
0.45  &  0.04 (0.01) &  0.03 (0.01) &  0.05 (0.01) &  0.06 (0.01) &  0.04 (0.01) &  0.02 (0.01) &  0.03 (0.01) &  0.05 (0.01) &    1m14s &   2m58s &   0s &    1s &   9s &   4s &   18s &      18s \\
0.5   &  0.00 (0.00) &  0.01 (0.01) &  0.00 (0.00) &  0.00 (0.00) &  0.00 (0.00) &  0.00 (0.01) &  0.01 (0.01) &  0.00 (0.01) &    1m14s &   2m57s &   0s &    1s &   9s &   4s &   18s &      18s \\
0.55  &  0.03 (0.01) &  0.02 (0.01) &  0.03 (0.01) &  0.05 (0.01) &  0.00 (0.00) &  0.00 (0.00) &  0.00 (0.00) &  0.00 (0.00) &    1m16s &   3m 0s &   1s &    1s &   9s &   4s &   17s &      18s \\
0.6   &  0.07 (0.00) &  0.09 (0.01) &  0.08 (0.01) &  0.10 (0.01) &  0.00 (0.01) &  0.00 (0.01) &  0.00 (0.00) &  0.00 (0.00) &    1m12s &   2m56s &   1s &    1s &   9s &   4s &   17s &      18s \\
0.65  &  0.10 (0.01) &  0.14 (0.01) &  0.12 (0.01) &  0.16 (0.01) &  0.00 (0.00) &  0.01 (0.01) &  0.00 (0.00) &  0.00 (0.00) &    1m13s &   2m59s &   1s &    1s &   9s &   4s &   17s &      18s \\
0.7   &  0.14 (0.01) &  0.19 (0.01) &  0.16 (0.01) &  0.20 (0.01) &  0.00 (0.00) &  0.00 (0.00) &  0.00 (0.00) &  0.00 (0.00) &    1m13s &   2m57s &   1s &    1s &   9s &   4s &   18s &      18s \\
0.75  &  0.18 (0.01) &  0.23 (0.01) &  0.19 (0.00) &  0.23 (0.01) &  0.00 (0.00) &  0.00 (0.01) &  0.00 (0.00) &  0.00 (0.00) &    1m15s &   2m55s &   1s &    1s &   9s &   4s &   18s &      18s \\
0.8   &  0.22 (0.01) &  0.26 (0.01) &  0.23 (0.01) &  0.26 (0.01) &  0.00 (0.01) &  0.00 (0.01) &  0.00 (0.00) &  0.00 (0.00) &    1m15s &   2m58s &   1s &    1s &   9s &   4s &   17s &      18s \\
0.85  &  0.26 (0.00) &  0.29 (0.01) &  0.27 (0.01) &  0.29 (0.01) &  0.00 (0.01) &  0.00 (0.01) &  0.00 (0.00) &  0.00 (0.00) &    1m16s &   2m55s &   1s &    1s &   9s &   4s &   18s &      18s \\
0.9   &  0.30 (0.01) &  0.33 (0.01) &  0.30 (0.00) &  0.32 (0.00) &  0.00 (0.01) &  0.00 (0.00) &  0.00 (0.00) &  0.00 (0.00) &    1m14s &   2m55s &   1s &    1s &   9s &   4s &   18s &      18s \\
0.95  &  0.34 (0.01) &  0.35 (0.00) &  0.34 (0.01) &  0.35 (0.00) &  0.00 (0.00) &  0.00 (0.00) &  0.00 (0.00) &  0.00 (0.00) &    1m 9s &   2m55s &   1s &    1s &   9s &   4s &   18s &      18s \\
0.999 &  0.38 (0.01) &  0.39 (0.01) &  0.38 (0.01) &  0.38 (0.00) &  0.00 (0.00) &  0.01 (0.01) &  0.00 (0.00) &  0.00 (0.00) &    1m18s &   2m45s &   1s &    1s &   9s &   4s &   18s &      18s \\
\bottomrule
\end{tabular}

}
\caption{
Results of stopping a fractional Brownian Motion for different Hurst parameters.
}
\label{hurst_table_all}
\end{table*}

%\caption{Results of stopping fractional Brownian Motion for different Hurst parameters.}
\end{landscape}

\subsection{Non-Markovian stock models -- additional tables}\label{sec:Non-Markovian stock models -- Additional Tables}
Additional results for the non-Markovian setting of a Heston model without the  variance as input are given in Tables~\ref{table_Dim_Heston_MaxCallr0_gt1}-\ref{table_Dim_Heston_MaxCall_div_gt1}.

\begin{table*}[!h]
\center
\resizebox{\textwidth}{!}{
\begin{tabular}{|c|r r r r r r r|r r r r r r r|}
\toprule
{} & \multicolumn{7}{c |}{price} & \multicolumn{7}{c |}{duration} \\
$d$ & LSM & DOS & NLSM & RLSM & FQI & RFQI & EOP & LSM & DOS & NLSM & RLSM & FQI & RFQI & EOP\\
\midrule
5         &   8.34 (0.07) &   8.29 (0.09) &   8.17 (0.06) &   8.31 (0.07) &   8.23 (0.04) &   8.34 (0.08) &   8.23 (0.04) &       11s &        7s &       3s &       0s &        3s &       0s &       0s \\
10        &  11.83 (0.07) &  11.81 (0.09) &  11.39 (0.16) &  11.83 (0.07) &  11.77 (0.04) &  11.82 (0.05) &  11.79 (0.05) &       29s &        6s &       3s &       0s &        6s &       0s &       0s \\
50        &  19.60 (0.07) &  20.04 (0.04) &  18.14 (0.37) &  19.32 (0.05) &  20.05 (0.06) &  20.08 (0.06) &  20.06 (0.03) &     8m50s &        7s &       3s &       0s &     6m36s &       1s &       0s \\
100       &  20.51 (0.09) &  23.57 (0.07) &  21.29 (0.46) &  22.87 (0.04) &  23.56 (0.07) &  23.67 (0.05) &  23.67 (0.05) &    40m44s &        9s &       3s &       0s &  1h21m35s &       1s &       0s \\
500       &             - &  31.62 (0.06) &  28.38 (0.55) &  31.33 (0.04) &             - &  32.09 (0.06) &  32.14 (0.02) &         - &       44s &       8s &       1s &         - &       1s &       0s \\
1000      &             - &  34.99 (0.08) &  33.03 (0.50) &  35.06 (0.04) &             - &  35.83 (0.05) &  35.84 (0.03) &         - &     1m16s &      15s &       2s &         - &       1s &       0s \\
2000      &             - &  37.77 (0.07) &  36.77 (0.32) &  38.83 (0.06) &             - &  39.64 (0.07) &  39.61 (0.04) &         - &     2m17s &      25s &       4s &         - &       2s &       0s \\
\bottomrule
\end{tabular}

}
\caption{Max call option on Heston for different numbers of stocks $d$.
}
\label{table_Dim_Heston_MaxCallr0_gt1}
\end{table*}

%\caption{Max call option on Heston for different numbers of stocks $d$. }

\begin{table*}[!h]
\center
\resizebox{\textwidth}{!}{
\begin{tabular}{|c|r r r r r r|r r r r r r|}
\toprule
{} & \multicolumn{6}{c |}{price} & \multicolumn{6}{c |}{duration} \\
$d$ & LSM & DOS & NLSM & RLSM & FQI & RFQI & LSM & DOS & NLSM & RLSM & FQI & RFQI\\
\midrule
5         &  12.29 (0.07) &  12.26 (0.06) &  12.12 (0.08) &  12.25 (0.07) &  12.38 (0.08) &  12.34 (0.07) &       12s &        6s &       3s &       0s &        2s &       0s \\
10        &  16.55 (0.06) &  16.54 (0.10) &  16.03 (0.19) &  16.50 (0.06) &  16.63 (0.09) &  16.64 (0.06) &       30s &        6s &       3s &       0s &       10s &       0s \\
50        &  25.24 (0.07) &  25.66 (0.07) &  23.67 (0.35) &  24.87 (0.04) &  25.71 (0.07) &  25.68 (0.04) &     8m42s &        8s &       3s &       0s &     7m34s &       1s \\
100       &  26.84 (0.09) &  29.22 (0.07) &  26.47 (0.62) &  28.45 (0.03) &  29.26 (0.06) &  29.32 (0.07) &    42m26s &       12s &       4s &       0s &  1h24m 4s &       1s \\
500       &             - &  36.47 (0.05) &  33.80 (0.65) &  36.26 (0.05) &             - &  36.93 (0.04) &         - &       56s &      13s &       1s &         - &       1s \\
1000      &             - &  39.25 (0.04) &  37.01 (0.34) &  39.33 (0.02) &             - &  39.93 (0.04) &         - &     1m49s &      23s &       2s &         - &       2s \\
2000      &             - &  41.45 (0.03) &  39.92 (0.26) &  42.25 (0.05) &             - &  42.78 (0.04) &         - &     3m58s &      43s &       5s &         - &       2s \\
\bottomrule
\end{tabular}

}
\caption{Min put option on Heston  for different numbers of stocks $d$ and varying initial stock price $x_0$. Here $r=2\%$ is used as interest rate.
}
\label{table_$x_0$s_Dim_Heston_MinPut_gt1}
\end{table*}

%\caption{Min put option on Heston  for different numbers of stocks $d$ and varying initial stock price $x_0$. Here $r=2\%$ is used as interest rate. }

\begin{table*}[!h]
\center
\resizebox{\textwidth}{!}{
\begin{tabular}{|c|r r r r r r|r r r r r r|}
\toprule
{} & \multicolumn{6}{c |}{price} & \multicolumn{6}{c |}{duration} \\
$d$ & LSM & DOS & NLSM & RLSM & FQI & RFQI & LSM & DOS & NLSM & RLSM & FQI & RFQI\\
\midrule
5         &   4.82 (0.03) &   4.78 (0.04) &   4.68 (0.04) &   4.75 (0.04) &   4.29 (0.12) &   4.57 (0.06) &       12s &        5s &       3s &       0s &        2s &       0s \\
10        &   7.20 (0.06) &   7.16 (0.04) &   6.92 (0.06) &   7.13 (0.05) &   6.60 (0.14) &   6.76 (0.16) &       29s &        6s &       3s &       0s &        8s &       0s \\
50        &  13.48 (0.05) &  13.98 (0.03) &  12.44 (0.18) &  13.69 (0.04) &  13.79 (0.03) &  13.72 (0.07) &     8m34s &        8s &       3s &       0s &     7m 7s &       1s \\
100       &  14.63 (0.07) &  17.13 (0.06) &  15.19 (0.32) &  16.83 (0.04) &  16.97 (0.07) &  16.99 (0.04) &    39m49s &       12s &       6s &       0s &  1h23m 4s &       1s \\
500       &             - &  24.31 (0.08) &  21.83 (0.63) &  24.37 (0.04) &             - &  24.69 (0.05) &         - &       54s &      12s &       1s &         - &       1s \\
1000      &             - &  27.42 (0.07) &  25.64 (0.55) &  27.73 (0.03) &             - &  28.08 (0.06) &         - &     1m39s &      23s &       2s &         - &       2s \\
2000      &             - &  30.10 (0.08) &  29.27 (0.36) &  31.09 (0.04) &             - &  31.50 (0.06) &         - &     3m47s &      43s &       5s &         - &       2s \\
\bottomrule
\end{tabular}

}
\caption{Max call  option on Heston  for different numbers of stocks $d$. Here $r=5\%$ is used as interest rate and $\delta = 10\%$ as dividend rate.
}
\label{table_Dim_Heston_MaxCall_div_gt1}
\end{table*}

%\caption{Max call  option on Heston  for different numbers of stocks $d$. Here $r=5\%$ is used as interest rate and $\delta = 10\%$ as dividend rate.}

\bibliography{references}

\end{document}